\documentclass[11pt]{article}

\makeatletter{}

\usepackage[colorlinks=true]{hyperref}
\usepackage[dvipsnames]{xcolor}
\usepackage[hang,flushmargin]{footmisc}

\usepackage{overpic}
\usepackage{subcaption}
\usepackage{amsfonts}
\usepackage{amsmath}
\usepackage{amsthm}
\usepackage{mathtools}
\usepackage{bm}

\usepackage{booktabs}    
\usepackage{array}
\usepackage{makecell}
\usepackage{colortbl}
\usepackage[show]{simple_notes}

\usepackage{natbib}

\usepackage[font=small,margin=0.0in,labelfont={sc},labelsep={colon}]{caption}

\usepackage{tikz}
\usepackage{microtype}
\usepackage{enumitem}

\definecolor{dark-gray}{gray}{0.3}
\definecolor{dkgray}{rgb}{.4,.4,.4}
\definecolor{dkblue}{rgb}{0,0,.5}
\definecolor{medblue}{rgb}{0,0,.75}
\definecolor{rust}{rgb}{0.5,0.1,0.1}

\usepackage[colorlinks=true]{hyperref}

\hypersetup{urlcolor=rust}
\hypersetup{citecolor=dkblue}
\hypersetup{linkcolor=dkblue}

\usepackage{setspace}

\usepackage{graphicx}
\usepackage{booktabs,longtable,tabu} 
\usepackage{multirow} 
\usepackage{float}

\usepackage[full]{textcomp}

\RequirePackage{times}

\RequirePackage{fancyhdr}
\RequirePackage{color}
\RequirePackage{algorithm}
\RequirePackage{algorithmic}
\RequirePackage{natbib}
\RequirePackage{eso-pic} 
\RequirePackage{forloop}

\usepackage{bm}

\numberwithin{equation}{section} 

\floatstyle{plaintop}
\newfloat{recipe}{thp}{lor}
\floatname{recipe}{Recipe}
\numberwithin{recipe}{section}

\usepackage{microtype}
\usepackage{graphicx}
\usepackage{booktabs} 

\usepackage{hyperref}

\usepackage{overpic}
\usepackage{subcaption}
\usepackage{amsfonts}
\usepackage{amsmath}
\usepackage{amsthm}
\usepackage{mathtools}
\usepackage{bm}

\usepackage{booktabs}    
\usepackage{array}
\usepackage{makecell}
\usepackage{colortbl}

\usepackage{amsmath}
\usepackage{amssymb}
\usepackage{amsthm}
\usepackage{upgreek}
\usepackage{mathrsfs}
\usepackage[show]{simple_notes}
\theoremstyle{plain}
\newtheorem{theorem}{Theorem}
\newtheorem{fact}{Fact}

\newtheorem{lemma}{Lemma}

\theoremstyle{definition}

\theoremstyle{remark}

\makeatletter
\newsavebox{\@brx}
\newcommand{\llangle}[1][]{\savebox{\@brx}{\(\m@th{#1\langle}\)}%
	\mathopen{\copy\@brx\kern-0.6\wd\@brx\usebox{\@brx}}}
\newcommand{\rrangle}[1][]{\savebox{\@brx}{\(\m@th{#1\rangle}\)}%
	\mathclose{\copy\@brx\kern-0.6\wd\@brx\usebox{\@brx}}}
\makeatother

\newcommand{\R}{\mathbb{R}}

\renewcommand{\hat}[1]{\widehat{#1}}

\newcommand{\D}{\mathcal{D}}

\renewcommand{\P}{\mathbb{P}}
\newcommand{\E}{\mathbb{E}}

\newcommand{\e}{\epsilon}

\newcommand{\tr}{\operatorname{tr}}

\newcommand{\var}{\operatorname{var}}

\newcommand{\ttop}{^{\top}}

\newcommand{\op}{_{\text{op}}}

\newcommand{\tsum}{\textstyle\sum}
\newcommand{\ts}{\textstyle}

\newcommand{\mnorm}[1]{\left\vert\kern-1.5pt\left\vert\kern-1.5pt\left\vert #1\right\vert\kern-1.5pt\right\vert\kern-1.5pt\right\vert}

\definecolor{astral}{RGB}{0,164,239}
\usepackage{fancyhdr}

\newcommand\blfootnote[1]{%
  \begingroup
  \renewcommand\thefootnote{}\footnote{#1}%
  \addtocounter{footnote}{-1}%
  \endgroup
}
\begin{document}

\begin{center}

{\bf{\LARGE{Error Estimation for Sketched SVD via the Bootstrap}}}

\vspace*{.3in}

{\large{
\begin{tabular}{cc}
Miles E.~Lopes$^1$, N.~Benjamin Erichson$^2$, Michael W.~Mahoney$^2$\\[0.4cm]

\end{tabular}
}}
\end{center}

\begin{abstract}
	In order to compute fast approximations to the singular value decompositions (SVD) of very large matrices, randomized sketching algorithms have become a leading approach.
	However, a key practical difficulty of sketching an SVD is that the user does not know how far the sketched singular vectors/values are from the exact ones.
	Indeed, the user may be forced to rely on analytical worst-case error bounds, which do not account for the unique structure of a given problem. As a result, the lack of tools for error estimation often leads to much more computation than is really necessary.
	To overcome these challenges, this paper develops a fully data-driven bootstrap method that numerically estimates the  \emph{actual} error of sketched singular vectors/values. In particular, this allows the user to inspect the quality of a rough initial sketched SVD, and then adaptively predict how much extra work is needed to reach a given error tolerance.
	Furthermore, the method is computationally inexpensive, because it operates only on sketched objects, and it requires \emph{no passes} over the full matrix being factored. Lastly, the method is supported by theoretical guarantees and a very encouraging set of experimental results.
\end{abstract}
\blfootnote{\noindent 1. UC Davis Department of Statistics.\\   2. UC Berkeley Department of Statistics and International Computer Science Institute.}


\section{Introduction}
During the past fifteen years, randomized sketching algorithms have emerged as a powerful framework for computing approximate solutions to large-scale matrix problems in machine learning, data analysis, and scientific computing. Accordingly, given that the singular value decomposition (SVD) is among the most essential matrix computations in these domains, it has been a major focal point in the literature on randomized numerical linear algebra (RandNLA)
\citep[e.g.,][among many others]{frieze2004fast,drineas2006fast2,rokhlin2010,clarkson2009numerical,halko2011finding,halkopca2011,woodruff2014sketching,musco:2015,MD16_chapter,tropp2019streaming}.
Broadly speaking, this line of work has led to a variety of randomized SVD algorithms that can offer higher speed and scalability than classical deterministic algorithms --- 
provided that the user is willing to tolerate some approximation error.
For this reason, the performance of a sketched SVD hinges on an appropriate tradeoff between computational cost and approximation error. However, one of the key unresolved challenges for users is that the actual error is \emph{unknown}, and as a result, it is hard to control the tradeoff efficiently.

In practice, this issue has typically been handled in one of two ways, each with their own limitations.  The simplest option is to rely on informal rules of thumb for deciding how much computation to spend on a sketched SVD (e.g.,~in terms of the ``sketch size''); but such rules give no warning when they fail, and this creates significant uncertainty in downstream computations. Alternatively, a more cautious option is to use analytical worst-case error bounds; but these present other challenges. First, these bounds often involve constants that are unspecified or dependent on unknown parameters. Second, even when explicit constants are available, worst-case bounds are necessarily pessimistic, and they do not account for the unique structure of a given problem. 

Based on these concerns, the RandNLA literature has shown rising interest in \emph{a  posteriori error estimation}, which seeks to improve upon worst-case analysis by numerically quantifying error with data-driven methods~\citep[e.g.,][]{liberty2007randomized,woolfe2008fast,halko2011finding,martinsson2016randomized,sorensen2016deim,duersch2017randomized,Lopes:ICML:2018,yu2018efficient,lopes2019bootstrap_matrix_mult,tropp2019streaming}. 
(Note that hereafter we will use the simpler phrase ``error estimation''.) Likewise, error estimation has the potential to make computations \emph{data-adaptive}, so that ``just enough'' work is done to achieve a specific error tolerance for a specific input.
Nevertheless, error estimation methods are still scarce for many sketching algorithms, and in particular, there has not yet been a method that can directly estimate the errors of the singular vectors/values in a sketched SVD. Therefore, the primary aim of the current paper is to develop a method for solving this problem. (More specific contributions associated with the method will be outlined in Section~\ref{sec:contrib}.)

\subsection{Preliminaries on SVD and Sketching}
Before we can explain the problem of error estimation in precise terms, it is necessary to briefly review a few aspects of classical SVD algorithms and their sketched versions.

\paragraph{Classical SVD.} \ Let $A\in\R^{n\times d}$ be a very large deterministic input matrix with $n\geq d$. (In the case where $d\geq n$, all of our work can be applied to the transpose of $A$ instead.) The SVD of $A$ is a factorization of the~form
\begin{equation*}
A = U \Sigma V^\top,
\end{equation*}
where the matrix $U \in \mathbb{R}^{n\times d}$ has orthonormal columns $u_1,\dots,u_d\in\R^n$ called left singular vectors, the matrix $V \in \mathbb{R}^{d\times d}$ has orthonormal columns $v_1,\dots,v_d\in\R^d$ called right singular vectors, and the non-negative matrix $\Sigma=\text{diag}(\sigma_1,\dots,\sigma_d)\in\R^{d\times d}$ contains the singular values $\sigma_1\geq \cdots\geq \sigma_d$. It will also be convenient to refer to the ``partial SVD'', which  returns $(u_1,\sigma_1,v_1),\dots,(u_k,\sigma_k,v_k)$ for some $k\in\{1,\dots,d\}$, and includes the ordinary SVD as a special case when $k=d$.
In large-scale settings, it is often unaffordable to use classical (deterministic) algorithms to compute the partial SVD to machine precision. 
With respect to floating point operations, the $\mathcal{O}(ndk)$ cost of this computation can be prohibitive, but an even more severe obstacle arises with respect to communication costs. Namely, in the common situation when $A$ is too large to be stored in fast memory, classical algorithms are often infeasible because they require many passes over the entire matrix $A$~\citep[cf.][]{golub2012matrix}.

\paragraph{Sketched SVD.}\ As a way of improving scalability, sketching algorithms proceed by mapping $A$ to a much shorter matrix $\tilde A\in \R^{t\times d}$ with $k\ll t\ll n$, referred to as a ``sketch'' of $A$. More specifically, the matrix $\tilde A$ is constructed as
\begin{equation*}
\textstyle
\tilde A =SA,
\end{equation*}
where $S\in\R^{t\times n}$ is a random ``sketching matrix'' that is generated by the user. In essence, the matrix $S$ is generated so that $\tilde A$ captures enough information to approximately reconstruct $(u_1,v_1,\sigma_1),\dots,(u_k,v_k,\sigma_k)$, and a myriad of choices for $S$ have been proposed in the literature \citep[cf.][for overviews]{mahoney2011randomized,woodruff2014sketching,kannan2017}. Most commonly, these choices ensure that the rows of $S$ are i.i.d.~vectors in $\R^n$, and that $\E[S\ttop S]=I_n$. For instance, two of the most well-known choices are \emph{Gaussian random projections (RP)}, where the rows of $S$ are~drawn from the Gaussian distribution $N(0,\ts\frac{1}{t}I_n)$, and  \emph{row-sampling matrices (RS)}, where the rows of $S$ are drawn from the set of re-scaled standard basis vectors $\{\ts\frac{1}{\sqrt{t p_1}}\,e_1,\dots,\ts\frac{1}{\sqrt{t p_n}}\,e_n\}\subset\R^n$ with sampling probabilities $p_1,\dots,p_n$.

Once the sketch $\tilde A$ has been obtained, the partial SVD of $A$ can then be approximated with a variety of approaches that entail different costs and benefits. (We refer to the previously cited papers for further background.)
Among these possibilities, our work will focus on the well-established \emph{sketch-and-solve} approach, which has the merit of being highly ``pass efficient'' (in the sense of~\citet{drineas2006fast2}) and inexpensive with respect to communication. In addition, it will follow as a consequence of our work that this approach has an extra advantage of being very amenable to error estimation. Furthermore, error estimation for sketch-and-solve is relevant to other approaches as well, because if the estimated error is large, then this can guide the user to consider a different approach that may deliver higher accuracy at higher cost.

To summarize how the sketch-and-solve approach works, it applies a classical partial SVD algorithm to the small matrix $\tilde A$ in order to compute its leading $k$ singular values $\tilde\sigma_1,\dots,\tilde \sigma_k$ and right singular vectors $\tilde v_1,\dots,\tilde v_k\in\R^d$. Next, 
another set of vectors $\breve u_j:=A\tilde v_j$ are computed for $1\leq j\leq k$ and  then normalized to yield $\tilde u_j:=\breve u_j/\|\breve u_j\|_2$.\footnote{In the unlikely case $A\tilde v_j=0$, we put $\tilde u_j:=0$, and going forward, we will use this same convention when normalizing vectors.} Then, the sequence $(\tilde u_1,\tilde \sigma_1,\tilde v_1),\dots,(\tilde u_k,\tilde \sigma_k,\tilde v_k)$ is returned as an approximation to $(u_1,\sigma_1,v_1),\dots,(u_k,\sigma_k,v_k)$. Altogether, the number of floating point operations involved is $\mathcal{O}(tdk+C_{\text{sketch}})$, where $C_{\text{sketch}}$ is the cost to obtain $\tilde A$. In particular, for some popular types of row-sampling sketching matrices, this latter cost is $C_{\text{sketch}}=\mathcal{O}(nd)$, and hence only linear in the size of the input $A$. Furthermore, in terms of communication, this approach generally \emph{only requires 1 or 2 passes over} $A$.

\subsection{The Error Estimation Problem}\label{sec:problem} 
After the sketched quantities $(\tilde u_1,\tilde \sigma_1,\tilde v_1),\dots,(\tilde u_k,\tilde \sigma_k,\tilde v_k)$ been computed, the user would (in principle) like to be able to compare them with the exact quantities $(u_1,\sigma_1, v_1),\dots,(u_k,\sigma_k, v_k)$ in terms of various error measures. To unify our discussion, let $\rho(w,w')$ denote a generic non-negative measure of error for comparing two unit vectors $w$ and $w'$ of the same dimension. Also, since it is of interest to have uniform control of error over a general set of indices $\mathcal{J}\subset\{1,\dots,k\}$, we will
consider the following random error variables
\begin{equation*}
\small
\tilde \e_{_U}\!(t):= \max_{j\in\mathcal{J}}\rho(\tilde u_j, u_j) \ \ \ \ \text{ and } \ \ \ \ \tilde \e_{_V}\!(t):= \max_{j\in\mathcal{J}}\rho(\tilde v_j, v_j),
\end{equation*}
as well as $\tilde \e_{_{\Sigma}}\!(t):= \!\max_{j\in\mathcal{J}}|\tilde\sigma_j-\sigma_j|$. Although these random variables have been written as functions of $t$ to indicate that they depend on the choice of $t$ through the sketched quantities, it is important to note that $\tilde \e_{_U}\!(t)$, $\tilde \e_{_{\Sigma}}\!(t)$, and $\tilde \e_{_V}\!(t)$ are \emph{never observed} by the user. 

\paragraph{Problem formulation.} \ Our goal is to estimate the \emph{tightest possible} upper bounds on $\tilde\e_{_U}\!(t)$, $\tilde \e_{_{\Sigma}}\!(t),$ and $\tilde\e_{_V}\!(t)$ that hold with probability at least $1-\alpha$, for some desired choice of $\alpha\in (0,1)$. In statistical terminology, such bounds are called the $(1-\alpha)$-quantiles of the error variables, and will be denoted as $q_{_U}\!(t)$, $q_{_{\Sigma}}\!(t)$, and $q_{_V}\!(t)$. More explicitly, $q_{_U}\!(t)$ is an unknown deterministic parameter defined as
$$q_{_U}\!(t):=\inf\Big\{q\in[0,\infty) \, \Big|\, \P\big(\tilde \e_{_U}\!(t) \leq q\big) \geq   1-\alpha\Big\},$$
and similarly for $q_{_{\Sigma}}\!(t)$ and $q_{_{V}}(t)$.

With the above notation in place, we propose to develop a fully data-driven method that will produce numerical estimates $\hat q_{_U}\!(t)$, $\hat q_{_{\Sigma}}\!(t)$, and $\hat q_{_V}\!(t)$.
Specifically, the proposed method is intended to satisfy two main criteria: (1) The estimates should be accurate substitutes for the true quantiles, in the sense that the event 
\begin{equation}\label{eqn:goodevent}
\tilde \e_{_U}\!(t) \ \leq \ \hat q_{_U}\!(t)
\end{equation}
occurs with probability nearly equal to $1-\alpha$, and likewise for $\hat q_{_{\Sigma}}\!(t)$ and $\hat q_{_{V}}\!(t)$; (2) The method should be computationally affordable  --- so that the extra step of error estimation does not interfere with the overall benefit of sketching. Accordingly, our work in Sections~\ref{sec:computational_considerations},~\ref{sec:theory}, and~\ref{sec:results} will show that these criteria are achieved.

\begin{figure}[t]
	\centering
	\DeclareGraphicsExtensions{.pdf}
	\begin{overpic}[width=0.55\textwidth]{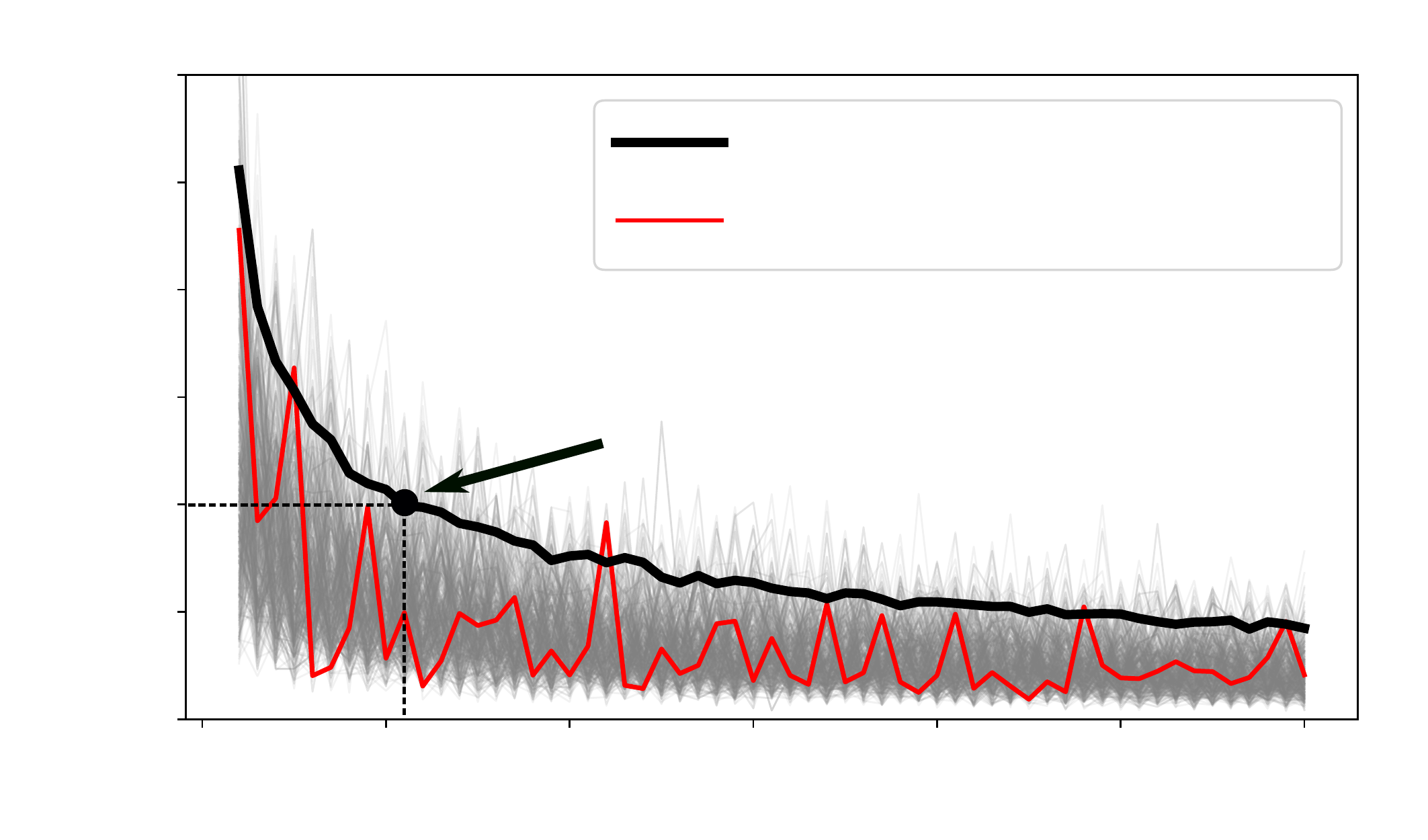} 
		\put(-1,24){\rotatebox{90}{\footnotesize error \ $\tilde \e_{_U}\!(t)$ }}
		\put(45,-1){\color{black}{\footnotesize sketch size $t$}} 	
		\put(32,33){\color{black}{\footnotesize \parbox{1.9in}{The error \ $\tilde \e_{_U}\!(t)$ is smaller than $0.02$ with probability 0.95 when $t=550$.}}}	
		
		\put(52,47.5){\color{black}{\footnotesize true \, 0.95-quantile}} 	
		\put(52,42.5){\color{black}{\footnotesize random evolution of $\tilde \e_{_U}\!(t)$}} 	
					
		\put(13.5,5){\color{black}{\footnotesize 0}} 	
		\put(24,5){\color{black}{\footnotesize 500}} 	
		\put(36,5){\color{black}{\footnotesize 1000}} 						
		\put(49,5){\color{black}{\footnotesize 1500}} 						
		\put(62,5){\color{black}{\footnotesize 2000}} 						
		\put(75,5){\color{black}{\footnotesize 2500}} 						
		\put(88,5){\color{black}{\footnotesize 3000}}

		\put(6,7.5){\color{black}{\footnotesize 0.00}} 	
		\put(6,22.5){\color{black}{\footnotesize 0.02}} 	
		\put(6,37.5){\color{black}{\footnotesize 0.04}} 						
		\put(6,52){\color{black}{\footnotesize 0.06}} 						
					
	\end{overpic}
	\caption{Visual interpretation of the quantile $q_{_U}\!(t)$.} 
	\label{fig:motivation}
	\vspace{-0.3cm}
\end{figure}
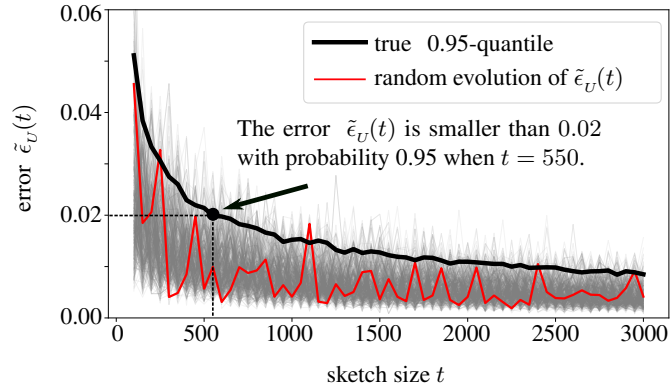

To give a more visual interpretation of the unknown quantile $q_{_U}\!(t)$, we show in Figure~\ref{fig:motivation} how it is related to the fluctuations of the error variable $\tilde \e_{_U}\!(t)$. If we imagine a hypothetical experiment where an oracle tells the user how $\tilde \e_{_U}\!(t)$ evolves as rows are incrementally added to a random sketching matrix $S$ (up to $t=3000$ rows), then the red curve displays this evolution. Similarly, the gray curves display the corresponding evolution over many independent repetitions of the same experiment. At any fixed $t$, the black curve represents the 0.95-quantile $q_{_U}\!(t)$, which lies above $\tilde \e_{_U}\!(t)$  in 95\% of the experiments.
Lastly, it should be emphasized that the user is not able to see any of these curves in practice.

In this way, if the unknown curve $q_{_U}\!(\cdot )$ were accessible, it could tell the user if a given initial sketch size $t_0$ is sufficient, and it could also allow the user to predict what larger sketch size $t_1>t_0$ might be needed to achieve a smaller error. With this motivation in mind, our proposed method will allow the user to obtain an accurate approximation to the curve $q_{_U}\!(\cdot)$. Furthermore, the method will produce the approximate curve after only a \emph{single run} of the sketching algorithm at a \emph{single initial sketch size $t_0$}. Indeed, it is somewhat surprising that this is possible, considering that $q_{_U}\!(\cdot)$ theoretically describes the error over many independent runs.

\subsection{Main Contributions and Related Work}\label{sec:contrib}

From a practical standpoint, the most significant contribution of our work is that it provides the first way to directly estimate the errors of singular vectors/values in a sketched SVD. By comparison, the most closely related methods only provide indirect error estimates, since they are designed to compute norm bounds with respect to $A$ and a low-rank approximant\footnote{Note that the top singular vectors/values of two matrices can be close when a norm distance between the matrices is large.}~\citep[e.g.,~][]{liberty2007randomized,woolfe2008fast,halko2011finding,sorensen2016deim,yu2018efficient,tropp2019streaming}. In particular, these types of approaches have mostly been limited to either the Frobenius or spectral norms, and generally produce upper bounds on the norms that are conservative \citep[cf.][p.1342]{yu2018efficient}. On the other hand, our approach is very flexible with respect to the choice of error measure, and it does not suffer from conservativeness because it targets the quantiles of $\tilde \e_{_U}\!(t)$, $\tilde \e_{_{\Sigma}}\!(t)$, and $\tilde \e_{_V}\!(t)$. Another crucial distinction is that the cited approaches generally require 1 or more extra passes over $A$, whereas our approach requires \emph{no passes}.

\paragraph{Other related work.} \ In recent years, bootstrap methods for error estimation have been considered in a number of related settings. In the statistics literature, the papers~\citep{el2019non,naumov2019bootstrap} have analyzed the bootstrap as a way to estimate the errors of sample eigenvalues and sample eigenvectors in the context of large covariance matrices, which is a topic originating from the classical results in~\cite{beran1985}. However, due to their focus on covariance matrices, these works are not directly applicable to analyzing both the left and right singular vectors of a sketched SVD.
Also, the theoretical setups in these works do not cover the important case of row-sampling sketches that is handled by our approach. More generally,  bootstrapping and other statistical approaches have recently been applied to quantify the errors of randomized algorithms for least-squares~\citep{ahfock2017statistical,Lopes:ICML:2018,dobriban2019}, matrix multiplication~\citep{lopes2019bootstrap_matrix_mult,lopes2019bootstrapping}, and gradient descent~\citep{fang2018,li2018,su2018,fang2019}.

\section{Method for Error Estimation}
\label{sec:method}

\paragraph{Intuition.} \ If it were possible to generate many independent samples of the error variables $\tilde \e_{_U}\!(t)$, $\tilde \e_{_{\Sigma}}\!(t)$, and $\tilde \e_{_V}\!(t)$, it would be straightforward to construct estimates of $q_{_U}\!(t)$, $q_{_{\Sigma}}\!(t)$ and $q_{_V}\!(t)$. For instance, if an oracle provided 100 independent samples, say $\tilde \e_{_V,1}(t),\dots,\tilde \e_{V,100}(t)$, of the error variable $\tilde \e_{_V}\!(t)$, then the 95th percentile of those 100 numbers would generally be a good estimate of $q_{_V}\!(t)$ when $\alpha$ is chosen as 0.05. However, in practice, generating these samples is infeasible, because it would require the user to re-run the sketching algorithm many times, and then find the (unknown) sketching error for each run. In spite of this difficulty, it turns out that it \emph{is possible} to efficiently generate \emph{approximate} samples of the error variables, and this is the essence of the bootstrap approach.

\paragraph{Generating approximate samples.} \
In order to generate approximate samples of $\tilde \e_{_U}\!(t)$, $\tilde \e_{_{\Sigma}}\!(t)$, and $\tilde \e_{_V}\!(t)$, the key idea is to create randomly ``perturbed versions'' of the sketched singular vectors/values. These randomly perturbed versions are denoted as $(\tilde u_1^*,\tilde \sigma_1^*,\tilde v_1^*),\dots,(\tilde u_k^*,\tilde\sigma_k^*,\tilde v_k^*)$, and will be designed to satisfy the following property. Namely, for each $j$, the fluctuations of $(\tilde u_j^*,\tilde \sigma_j^*,\tilde v_j^*)$ around $(\tilde u_j,\tilde \sigma_j,\tilde v_j)$ should be statistically similar to the fluctuations of  $(\tilde u_j,\tilde \sigma_j,\tilde v_j)$ around $(u_j,\sigma_j, v_j)$. In other words, this idea can be understood in terms of a correspondence between a ``bootstrap world'' and the ``real world'', such that the quantities $(\tilde u_j,\tilde \sigma_j,\tilde v_j)$ and $(\tilde u_j^*,\tilde \sigma_j^*,\tilde v_j^*)$ respectively play the roles of exact and sketched solutions in the bootstrap world.

The only remaining ingredient to address is the random mechanism for computing the perturbed quantities $(\tilde u_1^*,\tilde \sigma_1^*,\tilde v_1^*)$, $\dots,$ $(\tilde u_k^*,\tilde \sigma_k^*,\tilde v_k^*)$. In short this is done by forming a matrix $\tilde A^*\in\R^{t\times d}$ whose rows are sampled with replacement from the rows of $\tilde A$, and then doing computations with $\tilde A^*$ that are analogous to the ones in the original sketching algorithm. These details are given in Algorithm~\ref{alg:bootstrap}. 

As a matter of notation for expressing the outputs of Algorithm~\ref{alg:bootstrap}, it is necessary to define the empirical $(1-\alpha)$-quantile of a list of real numbers $x_1,\dots,x_B$. This quantity is written as $\text{quantile}[x_1,\dots,x_B;1-\alpha]$, and is defined as $\inf\{q\in \R \, |\, F_B(q)\geq 1-\alpha\}$, where we write $F_B(q):=\ts\frac{1}{B}\sum_{b=1}^B 1\{x_b\leq q\}$ for the empirical distribution function associated with $x_1,\dots,x_B$.

\setcounter{algorithm}{0}
\begin{algorithm}[t] 
{
\small
	\caption{(Bootstrap estimation of sketching error).}\label{alg:bootstrap}
	\normalfont
	\vspace{-0.2cm}
	\noindent \flushleft {\bf Input}: The sketch $\tilde{A}\in\R^{t\times d}$, the sketched sequence $(\tilde\sigma_1,\tilde v_1),\dots,(\tilde\sigma_k,\tilde v_k)$, and the number of samples $B$.\\[+0.1cm]
	
	$\bullet$ Compute the vectors $\tilde A \tilde v_1,\dots,\tilde A\tilde v_k$ and let $\breve u_1,\dots,\breve u_k$ denote their normalized versions with respect to the $\ell_2$-norm.\\[0.2cm]
	$\bullet $ {\bf For } $b=1,\dots,B$\; {\bf do \ in \ parallel}
	\begin{enumerate}
		\item Form a matrix $\tilde{A}^* \in \mathbb{R}^{t\times d}$ whose rows are obtained by sampling $t$ rows with replacement from $\tilde A$.
		\item Compute the top $k$ singular values and right singular vectors of $\tilde A^*$, denoted as $\tilde \sigma_1^*,\dots,\tilde\sigma_k^*$ and $\tilde v_1^*,\dots,\tilde v_k^*$. Then, compute the bootstrap samples
		\vspace{-0.5cm}
		\begin{align}
		& \tilde\e_{_{\Sigma,b}}^*(t):=\max_{j\in\mathcal{J}} |\tilde\sigma_j^* - \tilde \sigma_j|\label{eqn:Sigmasamples}\\[0.2cm]
		& \tilde \e_{_{V,b}}^*(t):=\max_{j\in\mathcal{J}} \rho( \tilde v_j^*,\tilde v_j).\label{eqn:Vsamples}
		\end{align}
		\vspace{-0.5cm}
		\item Compute the vectors $\tilde A\tilde v_1^*,\dots,\tilde A\tilde v_k^*$ and let $\breve u_1^*,\dots,\breve u_k^*$ denote their normalized versions with respect to the  $\ell_2$-norm. Then, compute the bootstrap sample

		\vspace{-0.2cm}
		\begin{align}
		& \tilde \e_{_{U,b}}^*(t):=\max_{j\in\mathcal{J}}\rho(\breve u_j^*, \breve u_j)\label{eqn:Usamples}.
		\end{align}
	\end{enumerate}
	\vspace{-0.3cm}
	{\bf Return:} The estimates $\hat{q}_{_U}\!(t)$, $\hat{q}_{_{\Sigma}}\!(t)$, and $\hat{q}_{_V}\!(t)$. They are defined as $\hat{q}_{_{U}}\!(t):=\text{quantile}[\tilde \e_{_{U,1}}^*\!(t),\dots,\tilde \e_{_{U,B}}^*\!(t);1-\alpha]$, and similarly for $\hat{q}_{_{\Sigma}}\!(t)$, and $\hat{q}_{_V}\!(t)$ using the samples generated in~\eqref{eqn:Sigmasamples} and~\eqref{eqn:Vsamples}. \\[0.1cm]

}
\end{algorithm}
\paragraph{Remarks.} \  To clarify a couple of small items, we do not use a subscript $b$ on the right sides of equations~\eqref{eqn:Sigmasamples},~\eqref{eqn:Vsamples}, and~\eqref{eqn:Usamples} because only the left sides need to be stored. With regard to the number of bootstrap samples $B$, our experiments will show that the modest choice $B=30$ works well in our settings of~interest.

\section{Computational Considerations}\label{sec:computational_considerations}
Given that sketching algorithms from RandNLA are intended to improve the efficiency of computations, it is important to explain why the extra step of error estimation does not interfere with this goal. Below, we describe some special aspects of Algorithm~\ref{alg:bootstrap} that make error estimation affordable.

\paragraph{Pass efficiency and scalability.} \ Because the inputs to Algorithm~\ref{alg:bootstrap} consist entirely of sketched objects, it follows that error estimation \emph{requires no access} to the full matrix $A$ (i.e.,~zero passes). Furthermore, in common situations where $n\gg d$, this also means that the processing cost of Algorithm~\ref{alg:bootstrap} will be \emph{independent of the large dimension $n$} (because the sizes of the sketched objects are independent of $n$). Indeed, these two features of Algorithm~\ref{alg:bootstrap} are quite favorable in comparison to a sketched SVD, which typically requires at least 1 or 2 passes over $A$, and typically has a processing cost that is linear in $n$. 

\paragraph{Parallelism and cloud/serverless computing.} \ The loop in Algorithm~\ref{alg:bootstrap} can be executed in an embarrasingly parallel manner, since each iteration $b=1,\dots,B$ is independent of the others. In addition, the computations at each iteration only have a small $\mathcal{O}(td)$ memory requirement, which is well-suited to modern distributed computing environments, such as cloud/serverless computing~\citep[cf.][]{kleiner2014,jonas2019cloud}. Likewise, it is natural to consider distributing the $B$ iterations across $\mathcal{O}(B)$ machines, and in this case, the processing cost of Algorithm~\ref{alg:bootstrap} is only $\mathcal{O}(tdk)$ on a per-machine basis. In fact, our experiments in Section~\ref{sec:results} show that when $A$ is on the order of 100GB, it is possible to obtain high quality error estimates  \emph{in a matter of seconds} when Algorithm~\ref{alg:bootstrap} is distributed across $B$ machines.

\paragraph{Extrapolation.} \ One more valuable feature of Algorithm~\ref{alg:bootstrap} is that it can be substantially accelerated via an \emph{extrapolation rule}. At a high level, this refers to a two-step process of (1) computing a ``rough'' sketched SVD based on an initial sketch size $t_0$, and then (2) using Algorithm~\ref{alg:bootstrap} to forecast what larger sketch size $t_1 > t_0$ is sufficient to achieve a desired error tolerance. At a more technical level, the extrapolation rule may be derived from the fact that the error variables $\tilde \e_{_U}\!(t)$, $\tilde \e_{_{\Sigma}}\!(t)$, and $\tilde \e_{_V}\!(t)$ tend to have fluctuations on the order of $1/\sqrt{t}$ (due to the central limit theorem).

Based on this anticipated scaling behavior, the error $\tilde \e_{_U}(t_0)$ at an initial sketch size $t_0$ should be larger than the error $\tilde \e_{_U}(t_1)$ at a sketch size $t_1>t_0$ by a factor of about $\sqrt{t_1/t_0}$. Hence, this suggests that if we use Algorithm~\ref{alg:bootstrap} to obtain an error estimate $\hat q_{_U}(t_0)$ from the initial sketched SVD, then we can re-scale this estimate by a factor of $\sqrt{t_0/t_1}$ to get a ``free'' estimate of $q_{_U}(t_1)$. In other words, we may define the extrapolated error estimate
	\begin{equation}\label{eq:extra_rule}
	\hat q^{ \text{\,\,ext\,}}_{_U}\!(t_1) \,:= \,\ts\frac{\sqrt{t_0}}{\sqrt{t_1}}\,\hat q_{_U}\!(t_0)
	\end{equation}
	for any choice of $t_1$ greater than $t_0$, and likewise for $\hat q^{ \text{\,\,ext\,}}_{_{\Sigma}}\!(t_1)$ and $\hat q^{ \text{\,\,ext\,}}_{_V}\!(t_1)$.
	
	The crucial point to notice about the extrapolation rule~\eqref{eq:extra_rule} is that running Algorithm~\ref{alg:bootstrap} based on a sketch of size $t_0$ is much cheaper than a sketch of size $t_1$ (by a factor of $t_1/t_0$ per iteration). Moreover, it turns out that this rule provides accurate estimates even when $t_1$ is larger than $t_0$ by \emph{an order of magnitude}, and this will be demonstrated empirically in Section~\ref{sec:results}. Altogether, this allows the user to allocate computational resources in a way that is \emph{adaptive} to the input at hand.

\section{Theory}\label{sec:theory}

In this section, we present our main theoretical result (Theorem~\ref{thm:main}), which shows that all three quantile estimates $\hat{q}_{_U}\!(t)$, $\hat q_{_{\Sigma}}\!(t)$, and $\hat q_{_V}\!(t)$ produced by Algorithm~\ref{alg:bootstrap} are asymptotically valid substitutes for the unknown quantiles $q_{_U}\!(t)$, $q_{_{\Sigma}}\!(t)$, and $q_{_V}\!(t)$. Furthermore, the result is applicable to either of the cases where $n \gg d$ or $ d\gg n$.
For brevity, we will deal only with the former case, because the latter case can be handled by considering the transpose of $A$.

\paragraph{Theoretical setup.} \ Our result is formulated in terms of a sequence of deterministic matrices $A_n\in\R^{n\times d}$ indexed by $n=1,2,\dots$, such that $d$ remains fixed as $n\to\infty$. Likewise, the number $k\in\{1,\dots,d\}$ and the set of indices $\mathcal{J}\subset\{1,\dots,k\}$ remain fixed as well. In addition, for each $n$, there is an associated random sketching matrix $S_n\in\R^{t_n\times n}$ and a number of bootstrap samples $B_n$ such that $t_n\to\infty$ and $B_n\to\infty$ as $n\to\infty$. Here, it is important to note that we make no restriction on the sizes of $t_n$ and $B_n$ relative to $n$, and hence we allow $t_n/n\to 0$ and \smash{$B_n/n\to 0$}. Lastly, in order to lighten notation in Theorem~\ref{thm:main}, we will suppress dependence on $n$ for the outputs of Algorithm 1, as well the exact singular vectors/values $(u_j,v_j,\sigma_j)$ of $A_n$ and their sketched versions $(\tilde u_j,\tilde v_j,\tilde \sigma_j)$.

With regard to the choice of error measure $\rho$ for the sketched singular vectors, we will focus on the ``sine distance'' $\rho_{\sin}$, defined for any Euclidean unit vectors $w$ and $w'$ of the same dimension as
\begin{equation}\label{eqn:sindist}
\rho_{\sin}(w,w') :=\sqrt{1-(w\ttop w')^2}.
\end{equation}
This is a standard error measure in the analysis of SVD, because it is invariant to sign changes of $w$ and $w'$, and hence automatically handles the sign ambiguity of singular vectors~\citep[cf.][]{Stewart:Sun,laug}. Its name derives from the fact that it can be interpreted as the sine of the acute angle between the one-dimensional subspaces spanned by $w$ and $w'$. 

Next, we state some assumptions for analyzing different types of sketching matrices. When $S_n$ is a Gaussian random projection, we make the following assumption.\\[-0.4cm]

\paragraph{Assumption RP.} \ \emph{There is a positive definite matrix $\mathsf{G}_{\infty}$ in $\R^{d\times d}$ such that $\ts\frac{1}{n}A_n\ttop A_n\to \mathsf{G}_{\infty}$ as $n\to\infty$, and the eigenvalues of $\mathsf{G}_{\infty}$ each have multiplicity 1.}\\

In the case when $S_n$ is a row-sampling matrix, we will use an assumption that augments Assumption RP with a few conditions. To state these conditions,  let $(p_{1},\dots,p_{n})$ denote the row-sampling probabilities for $S_n$, and let $a_l\in\R^d$ denote the $l$th row of $A_n$. In addition, let $\tilde r_n\in\R^d$ denote the first row of the re-scaled sketch $\ts\frac{\sqrt t}{\sqrt n}S_nA_n$, and let $\mathsf{v}_1,\mathsf{v}_2\in\R^d$ denote the top two eigenvectors of $\mathsf{G}_{\infty}$. 

\paragraph{Assumption RS.} \ \emph{The following conditions hold in addition to Assumption RP. For any fixed matrix $C\in\R^{d\times d}$, the sequence $\var(\tilde r_n\ttop C \tilde r_n)$ converges to a finite limit $\ell(C)$, possibly zero, as $n\to\infty$. Furthermore, if $C$ is chosen as $C=\mathsf{v}_1\mathsf{v}_1\ttop$ or $C=\mathsf{v}_{1}\mathsf{v}_{2}\ttop$, then the limit $\ell(C)$ is positive. Lastly, the growth condition $\max_{1\leq l\leq n}\|\ts\frac{1}{\sqrt{np_l}} a_{l}\|_2=o(t_n^{1/8})$ holds as $n\to\infty$.}

\paragraph{Remarks.} \ To provide some explanation for Assumptions RP and RS, the first mostly plays the role of a ``stability'' condition, which ensures that various functions of $A_n$ have well-behaved limits as $n\to\infty$. In particular, the $\ts\frac{1}{n}$ prefactor of the matrix $\ts\frac{1}{n}A_n\ttop A_n$ is natural because it allows the matrix to be written as the average $\ts\frac{1}{n}\sum_{l=1}^n a_la_l\ttop$. Also, the requirement that the eigenvalues of $\mathsf{G}_{\infty}$ have multiplicity 1 is used so that tools from matrix calculus can be applied to the matrix $\ts\frac{1}{n}A_n\ttop A_n$ within a neighborhood of $\mathsf{G}_{\infty}$. (Without a requirement of this type, the functions that send a matrix to its  eigenvectors/values become non-differentiable~\citep[Ch.~9.8]{Magnus:2019}.) Next, the conditions in Assumption RS are needed to rule out certain extreme types of matrices $A_n$ that interfere with techniques related to the central limit theorem. Lastly, in the supplementary material, we provide detailed examples of matrices $A_n$ that satisfy both assumptions.

In a nutshell, our main result shows that for large problems, Algorithm~1 provides estimates $\hat q_{_U}\!(t)$, $\hat q_{_{\Sigma}}\!(t)$, and $\hat q_{_{V}}(t)$ that nearly achieve the ideal coverage probability of $1-\alpha$, as in~\eqref{eqn:goodevent}. In addition, it is worth noting that the probability $\P$ in Theorem~\ref{thm:main} accounts for all sources of randomness (both from the sketching matrix and bootstrap sampling).

\begin{theorem}\label{thm:main}
Suppose that Assumption RP holds when $S_n$ is a Gaussian random projection, or that Assumption RS holds when $S_n$ is a row-sampling matrix. Also, let $\hat q_{_U}(t_n)$, $\hat q_{_{\Sigma}}(t_n)$, and $\hat q_{_V}(t_n)$ denote the outputs of Algorithm~\ref{alg:bootstrap}. Then, for any fixed set $\mathcal{J}\subset\{1,\dots,k\}$ with $1\in\mathcal{J}$, and any $\alpha\in (0,1)$, the following three limits hold as $n\to\infty$,
	\vspace{-0.1cm}
	\begin{align}
	&\P\Big(\max_{j\in\mathcal{J}} \rho_{\sin}(\tilde u_j,u_j) \, \leq \, \hat{q}_{_U}\!(t_n)\Big) \, \xrightarrow{ \ \ }\, 1-\alpha,\label{eqn:limU}\\[0.2cm]
	& \P\Big(\max_{j\in\mathcal{J}} |\tilde \sigma_j-\sigma_j|\, \leq \, \hat{q}_{_\Sigma}\!(t_n)\Big) \, \xrightarrow{ \ \ } \, 1-\alpha,\label{eqn:limSigma}\\[0.2cm]
		& \P\Big(\max_{j\in\mathcal{J}} \rho_{\sin}(\tilde v_j,v_j) \, \leq \, \hat{q}_{_V}\!(t_n)\Big) \, \xrightarrow{ \ \ }\, 1-\alpha.\label{eqn:limV}
			\end{align}

\end{theorem}
\paragraph{Remarks.} \ The proof is deferred to the appendices due its length. The main theoretical challenge is to establish central limit theorems for each of the random variables $\max_{j\in\mathcal{J}} \rho_{\sin}(\tilde u_j,u_j)$, $\max_{j\in\mathcal{J}} |\tilde \sigma_j-\sigma_j|$, and $\max_{j\in\mathcal{J}} \rho_{\sin}(\tilde v_j,v_j)$, as well as their bootstrap analogues. In  carrying this out, some of the essential technical ingredients are explicit formulas for matrix differentials (Jacobians) associated the functions that send a matrix to its eigenvectors/values~\citep[Ch.~9.8]{Magnus:2019}. More specifically, these formulas play an important role in determining the asymptotic variance in the central limit theorems just mentioned.  Another notable technical point is that the analysis handles the left singular vectors (in $\R^n$) and the right singular vectors (in $\R^d$) in a streamlined way --- even though the left singular vectors have a \emph{diverging dimension} as $n\to\infty$.

Lastly, to understand how Theorem~\ref{thm:main} fits into the broader context of the literature on sketched SVD, it should be emphasized that our analysis is based on \emph{distributional approximation}, whereas most other theoretical work has been based on large-deviation results. The key distinction is that distributional approximation allows us to show that the coverage probabilities of $\hat{q}_{_U}\!(t_n)$, $\hat{q}_{_{\Sigma}}\!(t_n)$ and $\hat{q}_{_V}\!(t_n)$ approach the ideal value of $1-\alpha$, whereas large-deviation results are generally only able to quantify such probabilities up to constants that are typically unspecified or conservative. In this way, our theoretical guarantees show that the bootstrap method allows the user to have fine-grained control over the coverage probability through the choice of $\alpha$.

\section{Experiments}
\label{sec:results}

In this section, we present a collection of synthetic and natural examples that demonstrate the practical performance of Algorithm~\ref{alg:bootstrap}. In particular, we show that the extrapolation rule (\ref{eq:extra_rule}) accurately predicts error as a function of the sketch size $t$. For simplicity, all the synthetic examples in Section~\ref{sec:results_synthetic} deal with the approximation of the leading triple $(u_1,\sigma_1,v_1)$, so that the error variables $\tilde \e_{_U}\!(t)$, $\tilde \e_{_{\Sigma}}\!(t)$, and $\tilde \e_{_V}\!(t)$ correspond to the index set $\mathcal{J}=\{1\}$. Other choices of the index set $\mathcal{J}$ are considered for real data in Section~\ref{sec:exapp}, as well as for synthetic data in Appendix~\ref{app:otherJ}. Also, the sine distance~\eqref{eqn:sindist} will be used in all examples as the measure of error for the singular~vectors.

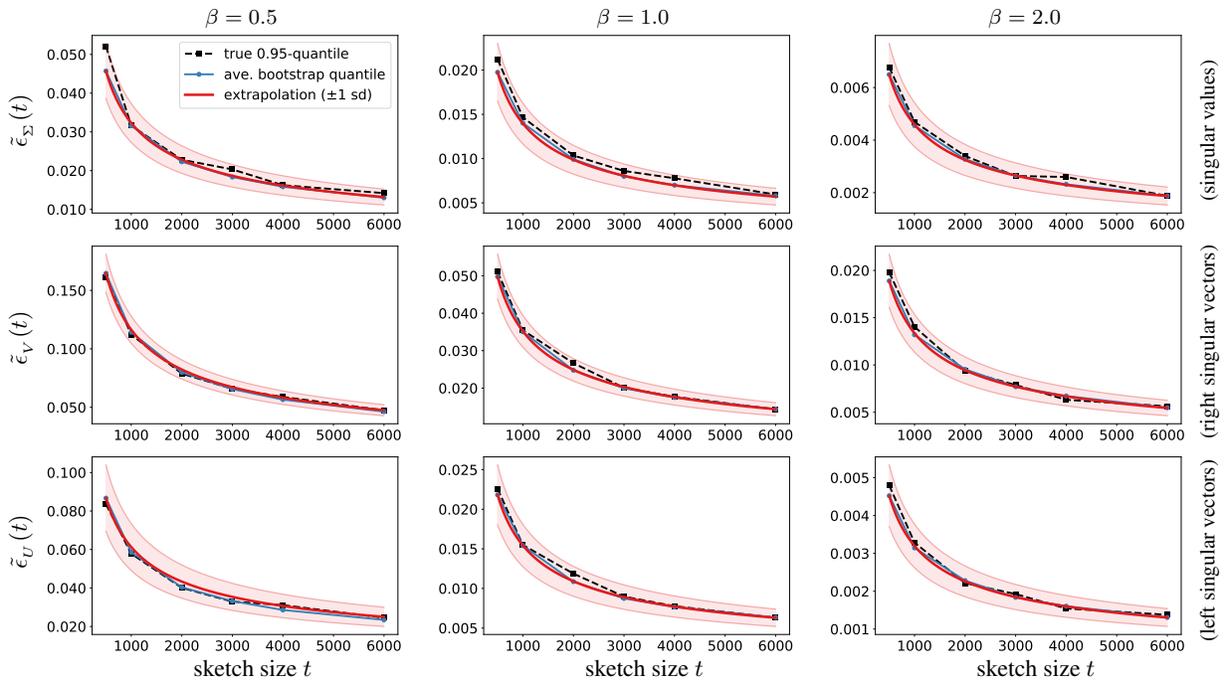
\begin{figure*}[!b]
	
	\centering
	\begin{subfigure}{1\textwidth}	
		\centering
		\DeclareGraphicsExtensions{.pdf}
		\begin{overpic}[width=0.31\textwidth]{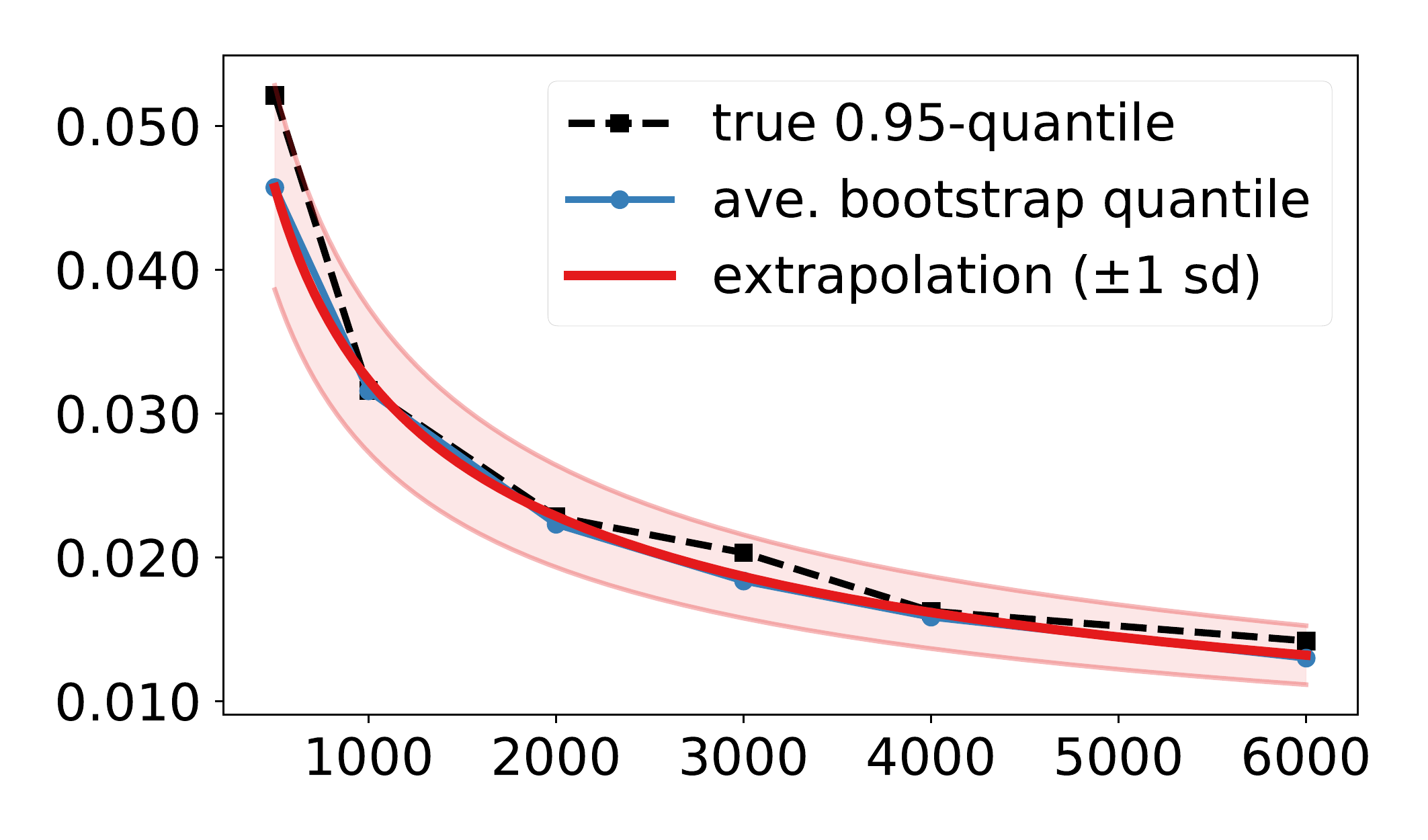} 
			\put(-6,26){\rotatebox{90}{\footnotesize $\tilde\e_{_{\Sigma}}(t)$}}
			\put(45,58){\color{black}{\scriptsize $\beta=0.5$}} 			
		\end{overpic}\hspace*{-0.2cm}
		~
		\begin{overpic}[width=0.31\textwidth]{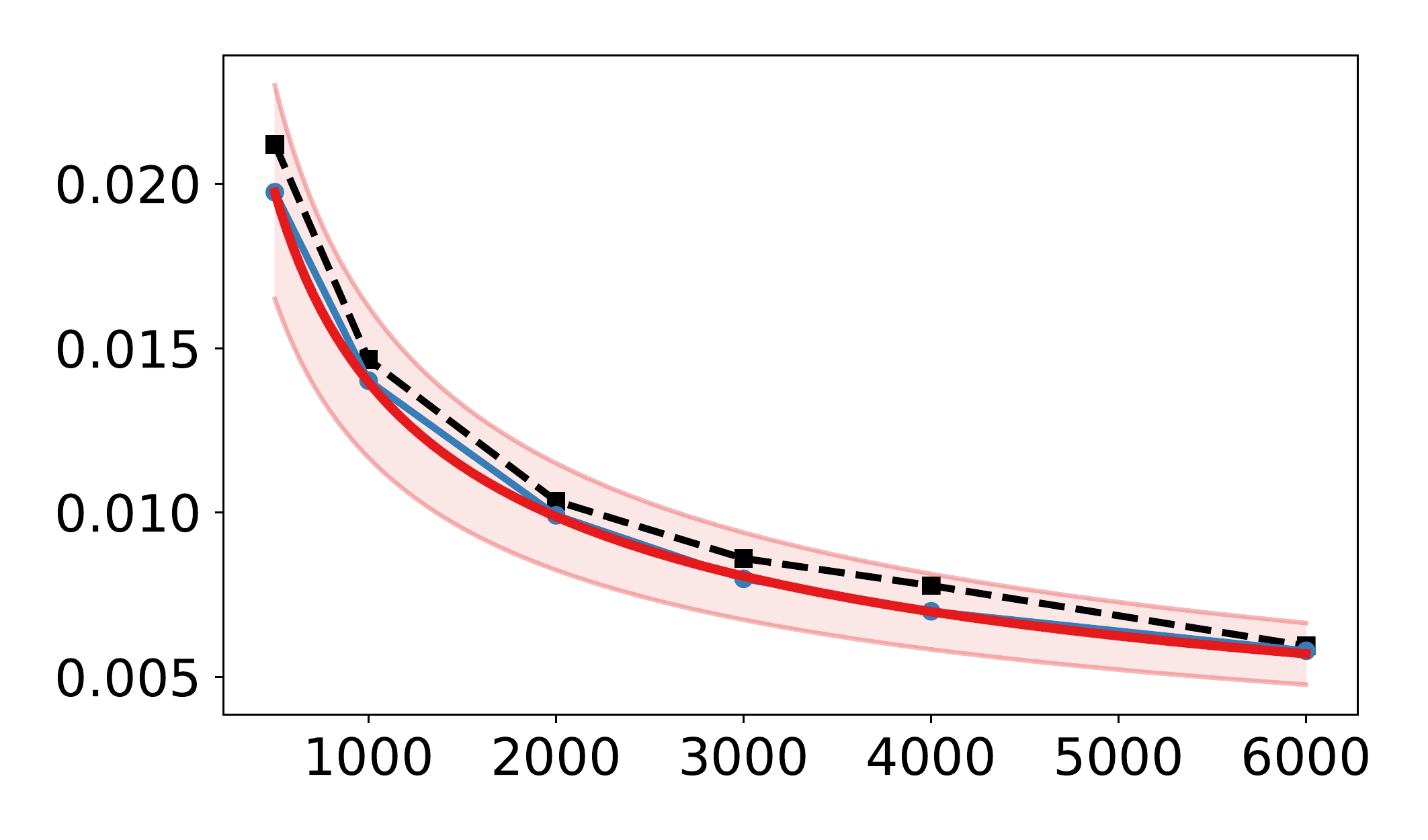} 
			\put(45,58){\color{black}{\scriptsize $\beta=1.0$}} 			 			
		\end{overpic}\hspace*{-0.2cm}
		~
		\begin{overpic}[width=0.31\textwidth]{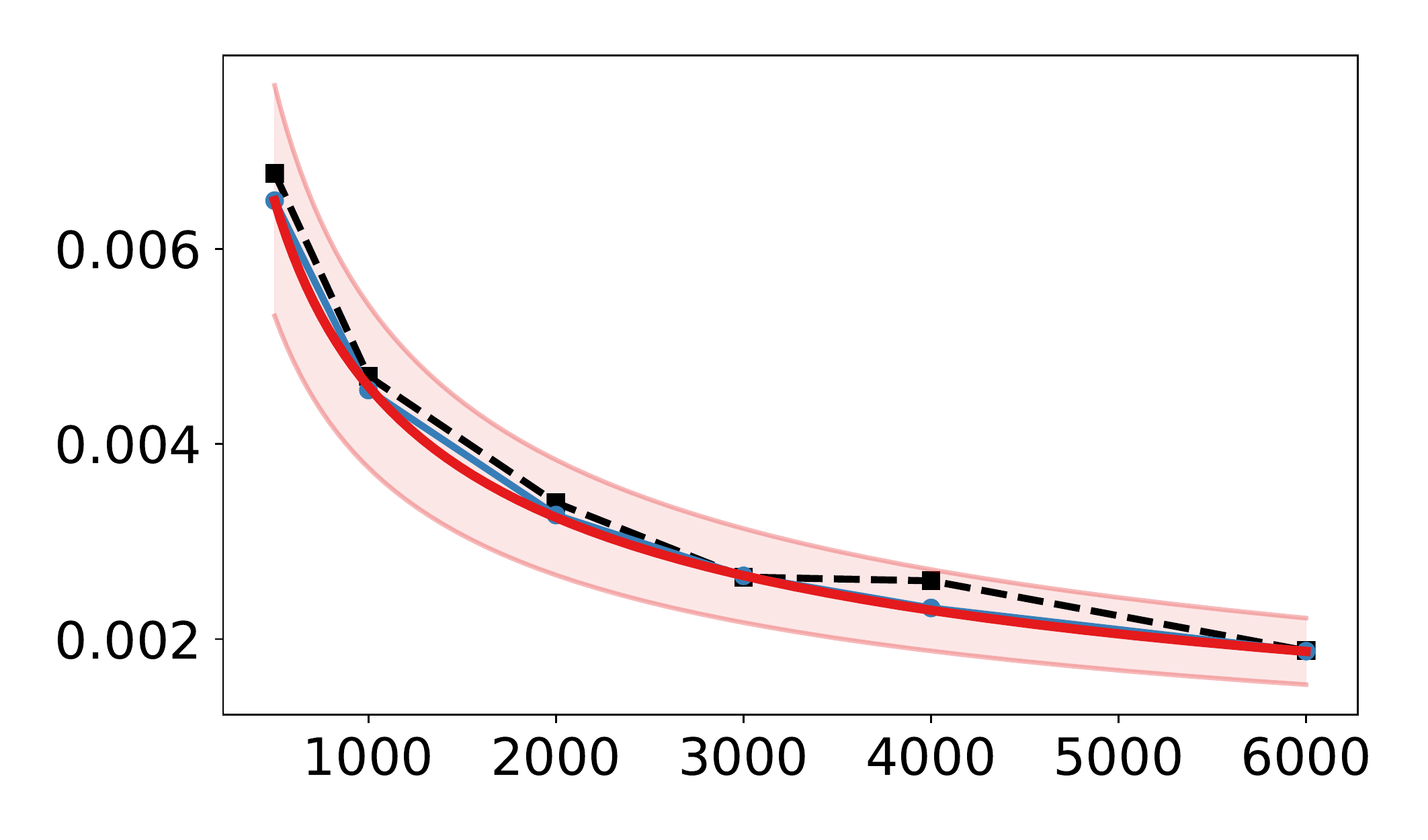} 
			\put(45,58){\color{black}{\scriptsize $\beta=2.0$}} 
			\put(100,12){\rotatebox{90}{\scriptsize (singular values)}}
		\end{overpic}
	\end{subfigure}\vspace{-0.2cm}	
	
	\begin{subfigure}{1\textwidth}	
		\centering
		\DeclareGraphicsExtensions{.pdf}
		\begin{overpic}[width=0.31\textwidth]{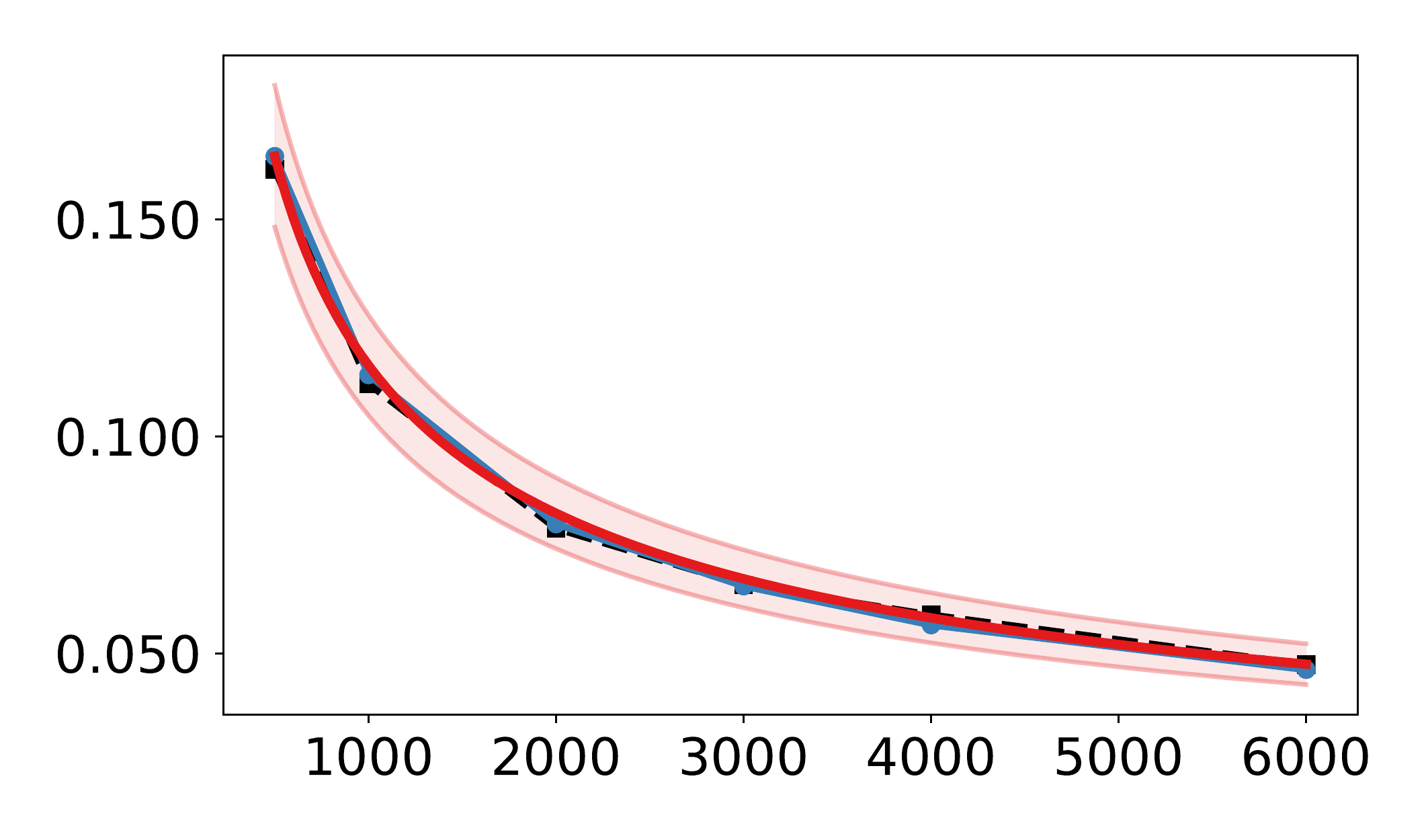} 
			\put(-6,24){\rotatebox{90}{\footnotesize $\tilde\e_{_{V}}(t)$}}
		\end{overpic}\hspace*{-0.2cm}
		~
		\begin{overpic}[width=0.31\textwidth]{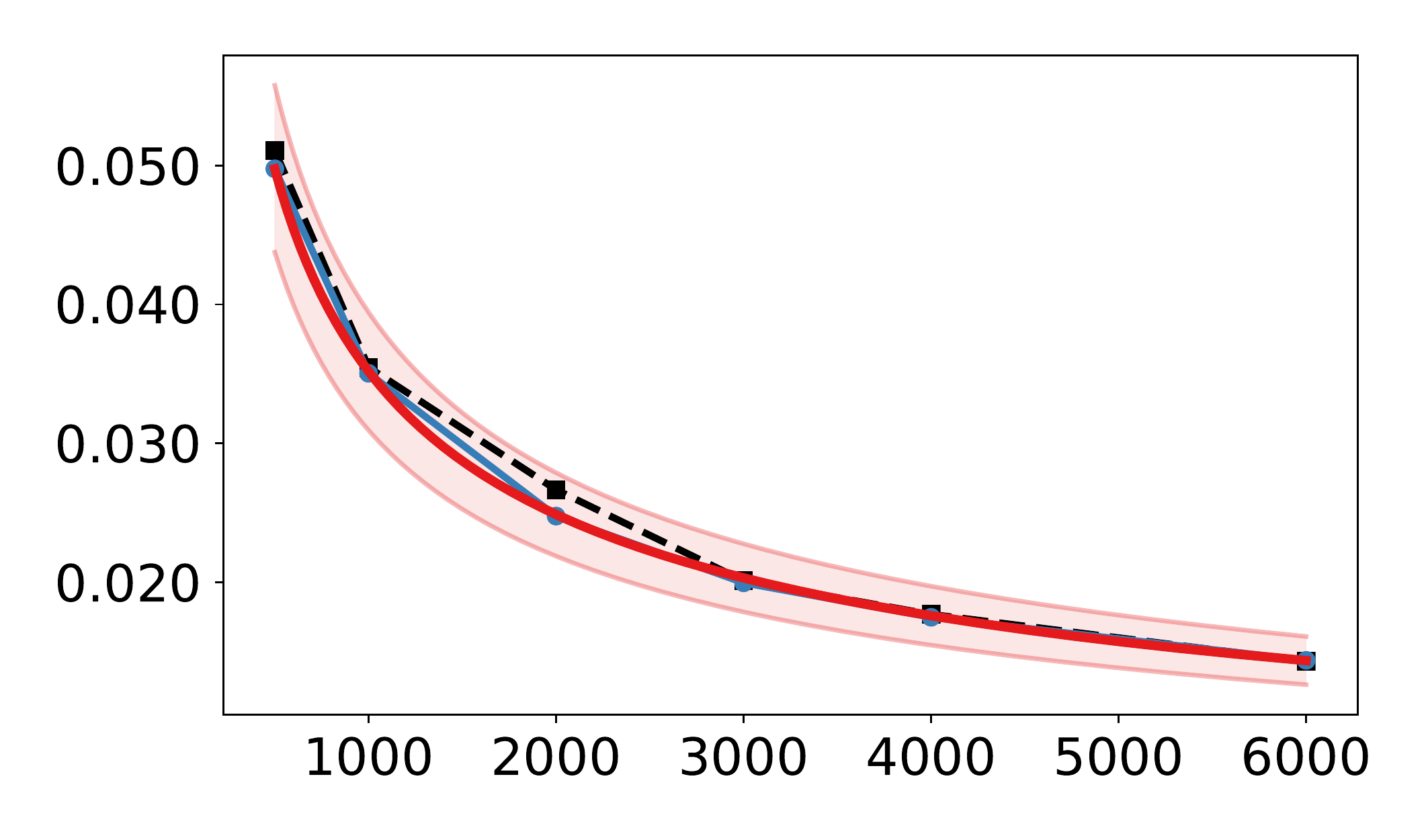} 
		\end{overpic}\hspace*{-0.2cm}
		~
		\begin{overpic}[width=0.31\textwidth]{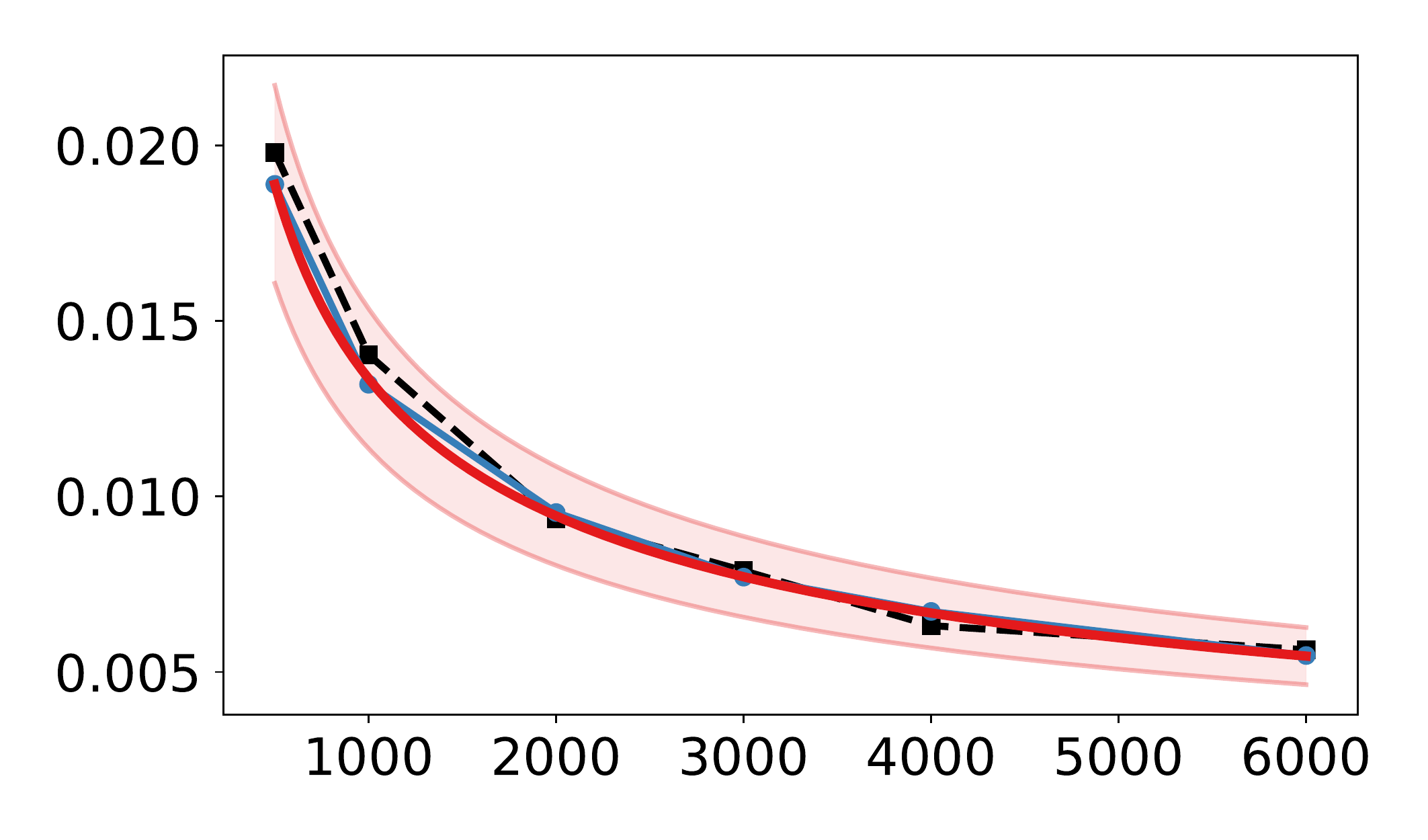} 
			\put(100,5){\rotatebox{90}{\scriptsize (right singular vectors)}}			
		\end{overpic}
	\end{subfigure}\vspace{-0.2cm}	

	\begin{subfigure}{1\textwidth}	
	\centering
	\DeclareGraphicsExtensions{.pdf}
	\begin{overpic}[width=0.31\textwidth]{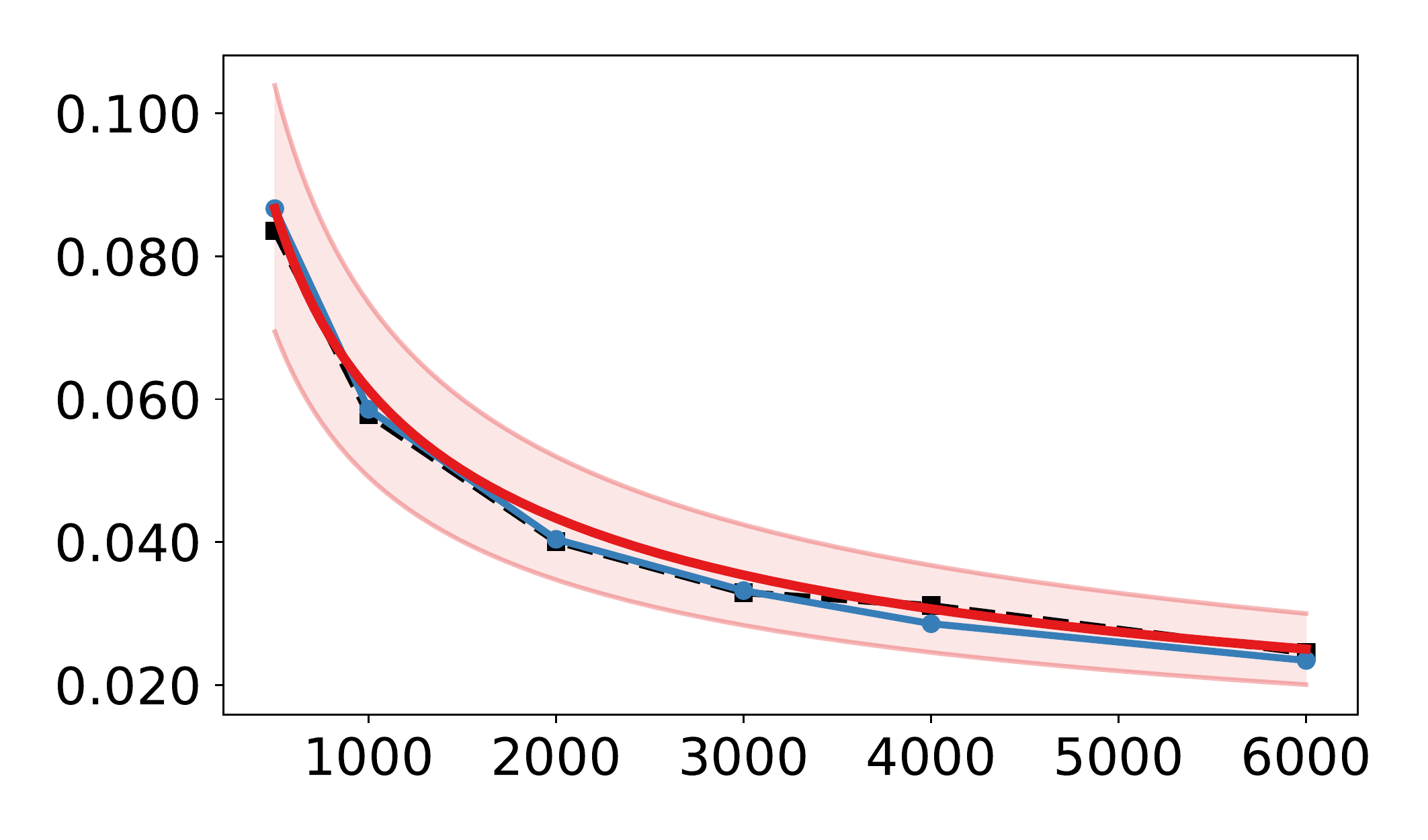} 
		\put(42,-2){\color{black}{\footnotesize sketch size $t$}}   
		\put(-6,26){\rotatebox{90}{\footnotesize $\tilde\e_{_{U}}(t)$}}
	\end{overpic}\hspace*{-0.2cm}
	~
	\begin{overpic}[width=0.31\textwidth]{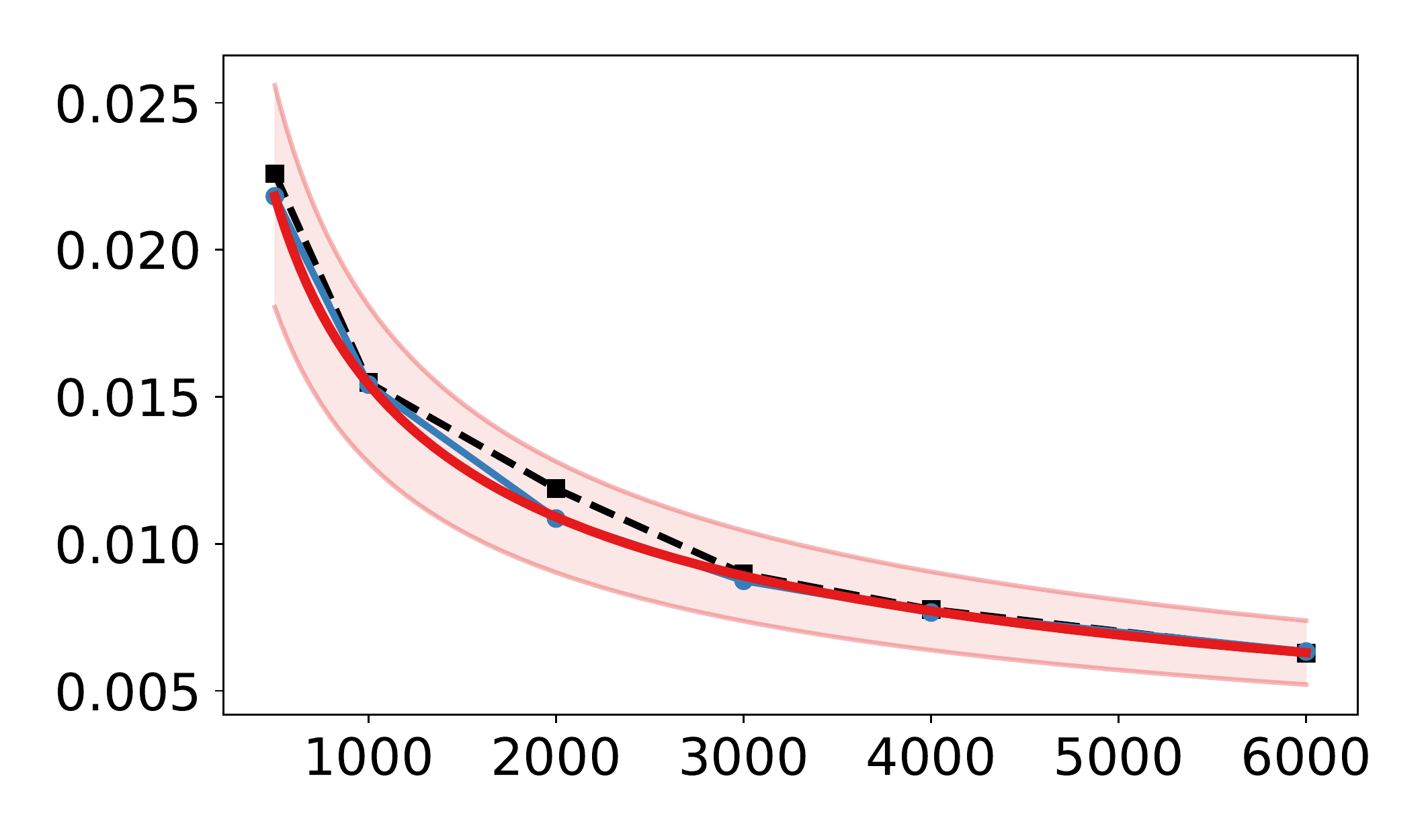} 
		\put(42,-2){\color{black}{\footnotesize sketch size $t$}}   
	\end{overpic}\hspace*{-0.2cm}
	~
	\begin{overpic}[width=0.31\textwidth]{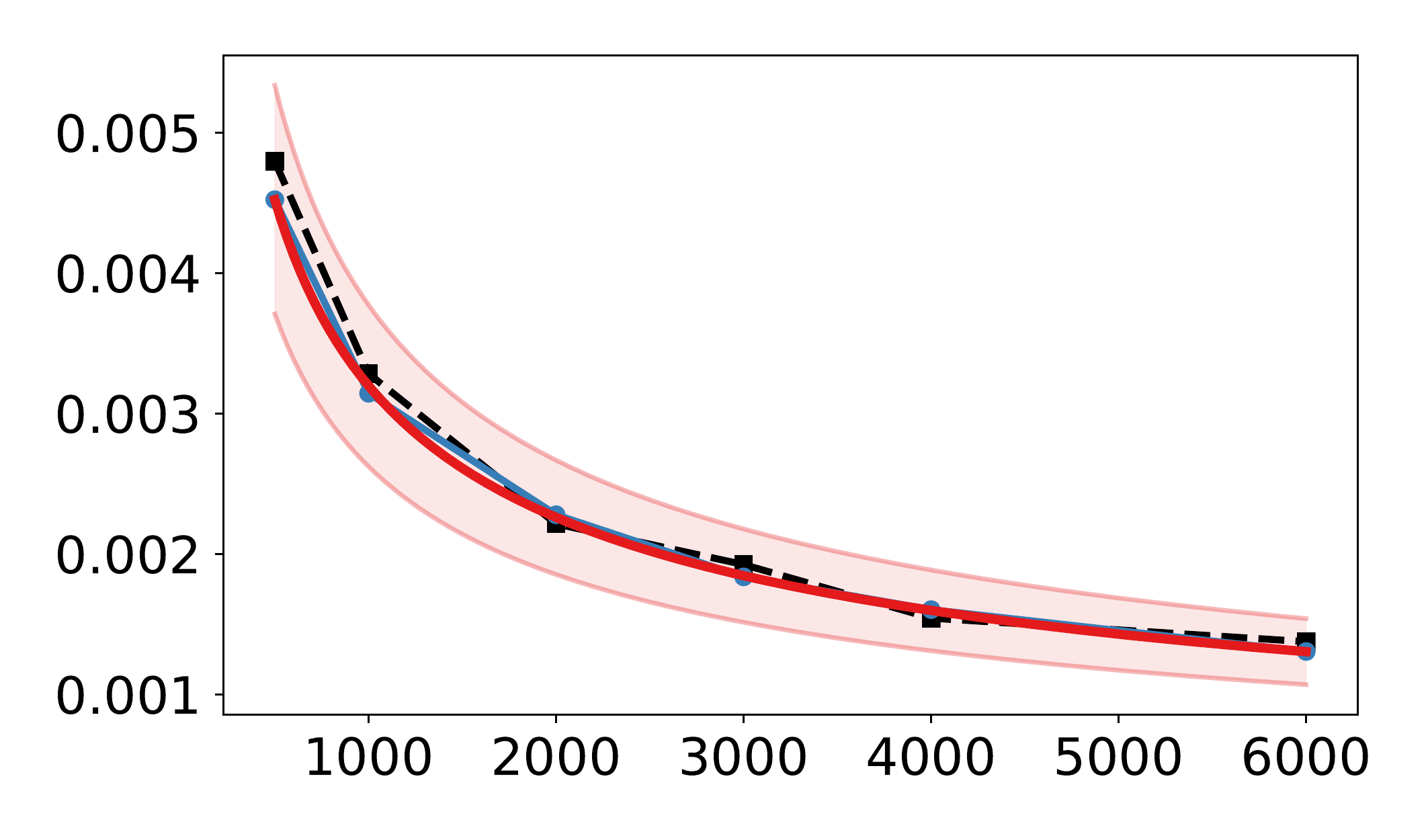} 
		\put(42,-2){\color{black}{\footnotesize sketch size $t$}}   
		\put(100,7){\rotatebox{90}{\scriptsize (left singular vectors)}}			
	\end{overpic}
	\end{subfigure}
	
	\vspace{+.2cm}	
	
	\caption{We consider artificial matrices of dimension $(n,d)=(10^5,3\times 10^3)$ that have singular value decay profiles of the form $\sigma_j=j^{-\beta}$ for $j\in\{1,\dots,d\}$ with $\beta\in\{0.5, 1.0, 2.0\}$. The error variables correspond to the index set $\mathcal{J}=\{1\}$, and the simulations involve $500$ trials and $30$ bootstraps per trial. The rows correspond to the error quantiles for the singular values (top), right singular vectors (middle), and left singular vectors (bottom).}

	\label{fig:results_svd_sketching_poly}
\end{figure*}

\subsection{Synthetic examples}\label{sec:results_synthetic}

First, we consider tall synthetic matrices with $(n,d)=(10^5,3000)$ that are characterized by low effective rank and varying degrees of singular value decay.

\paragraph{Parameter settings.} \ 
The matrix $A$ was specified in terms of the three factors $U$, $\Sigma$, and $V$ of its SVD.
The factors $U$ and $V$ were generated at random from the uniform (Haar) distributions on the sets of orthonormal matrices of sizes $n\times d$ and $d\times d$ respectively.
The singular values of $A$ were chosen as $\Sigma = \textup{diag}(1^{-\beta},2^{-\beta},\dots,d^{-\beta})$ for three choices of the decay parameter $\beta\in\{0.5,1.0,2.0\}$.

\paragraph{Design of experiments.} \ For each choice of the sketch size in a grid $t\in\{500,\dots,6000\}$, we generated $500$ independent sketching matrices $S\in \mathbb{R}^{t\times n}$, which yielded $500$ realizations of $\tilde{A} \in \mathbb{R}^{t\times d}$. 
(Here, we used ``squared-length sampling''~\citep[cf.][]{frieze2004fast} to construct the sketch $\tilde A$ in each trial, since it is one of the most popular options for row sampling.)
Next, each realization of $\tilde A$ yielded sketching errors for the leading triple $(u_1,\sigma_1,v_1)$ corresponding to the index set $\mathcal{J}=\{1\}$, which were then used to compute the corresponding error variables $\e_{_U}\!(t)$, $ \e_{_{\Sigma}}\!(t)$, and $ \e_{_V}\!(t)$.
In turn, we treated the empirical 95th percentiles of these 500 realizations as ground truth for the ideal quantiles $q_{_U}\!(t)$, $q_{_{\Sigma}}\!(t)$, and $q_{_V}\!(t)$, plotted with black dashed lines in Figure~\ref{fig:results_svd_sketching_poly}.

In a similar manner, Algorithm~\ref{alg:bootstrap} was applied to the sketched SVD resulting from each of the 500 matrices $\tilde A$ at each sketch size $t\in\{500,\dots,6000\}$, using a choice of $B=30$ in every instance. In total, this produced 500 realizations of $\hat{q}_{_U}\!(t)$, $\hat{q}_{_{\Sigma}}\!(t)$ and $\hat{q}_{_V}\!(t)$ at each $t$.
The respective averages of these 500 estimates at each $t$ are plotted with solid blue lines in Figure~\ref{fig:results_svd_sketching_poly}.
To study the performance of the extrapolation rule~\eqref{eq:extra_rule}, we applied it to each of the 500 quantile estimates produced at $t_0=500$, which resulted in 500 realizations of each of the curves $\hat q^{ \text{\,\,ext\,}}_{_U}\!(\cdot)$, $\hat q^{ \text{\,\,ext\,}}_{_{\Sigma}}\!(\cdot)$, and $\hat q^{ \text{\,\,ext\,}}_{_V}\!(\cdot)$.
The respective averages of each type of curve are plotted with solid red lines in Figure~\ref{fig:results_svd_sketching_poly}, and the light red envelopes represent $\pm 1$ standard deviation around the average.

\paragraph{Results for synthetic examples.} \ The results show that the bootstrap quantile estimates, as well as their extrapolated versions, are excellent approximations to the true quantiles over the entire range of $t$. 
This behavior is also consistent across the different decay parameters $\beta=\{0.5,1.0,2.0\}$. Moreover, from looking at the second and third rows in Figure~\ref{fig:results_svd_sketching_poly}, we see that this performance holds when the true quantiles range over three different orders of magnitude $(10^{-1}, 10^{-2}, 10^{-3})$.
Hence, even in situations where the sketching errors are larger, the bootstrap is useful because it can tell the user that a higher precision SVD algorithm may be needed to reach a given error tolerance.

\begin{figure*}[!b]
	
	\centering
	\begin{subfigure}{1\textwidth}	
		\centering
		\DeclareGraphicsExtensions{.pdf}
		\begin{overpic}[width=0.31\textwidth]{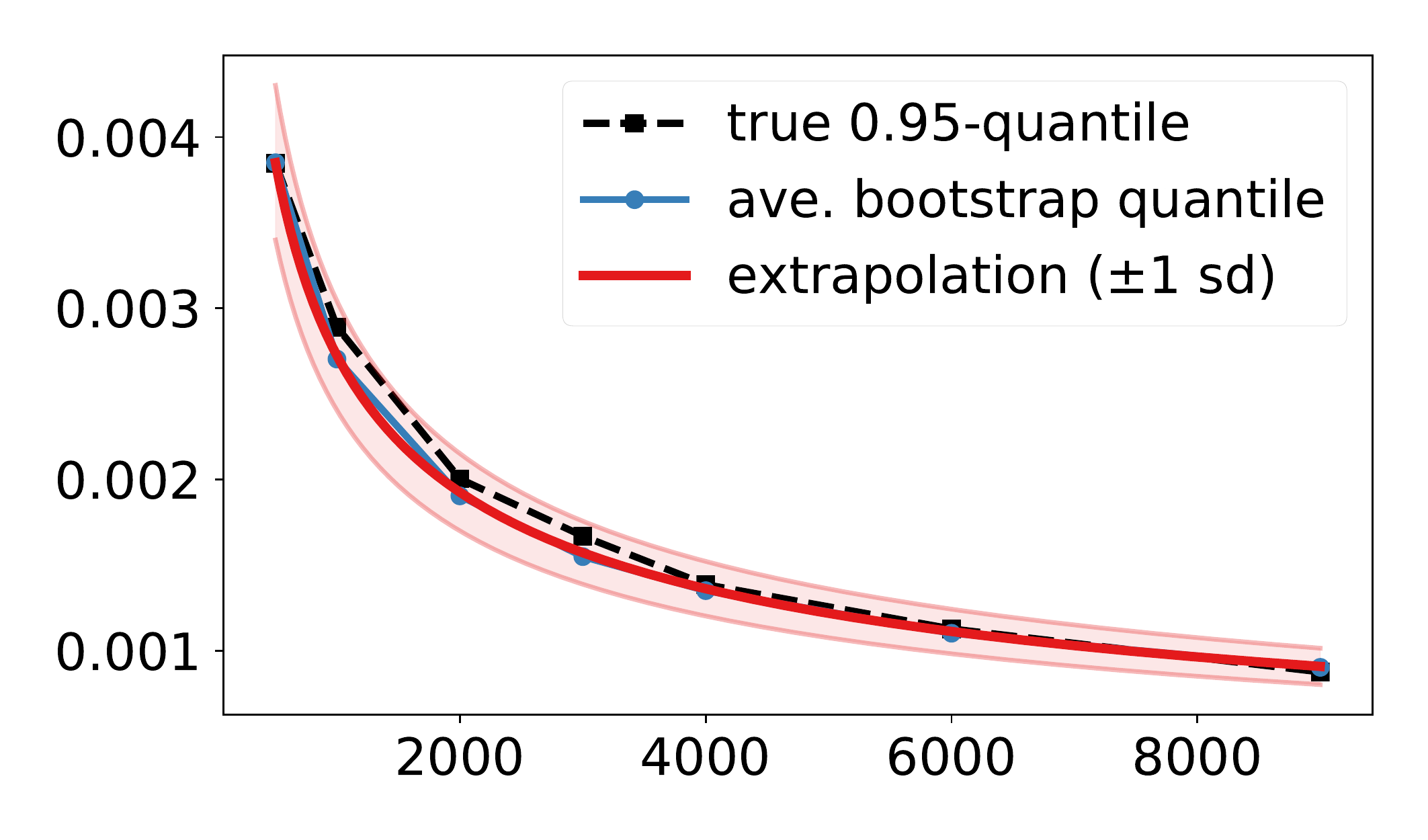} 
			\put(-6,26){\rotatebox{90}{\footnotesize $\tilde\e_{_{U}}(t)$}}
		\end{overpic}
		~
		\DeclareGraphicsExtensions{.png}
		\begin{overpic}[width=0.31\textwidth]{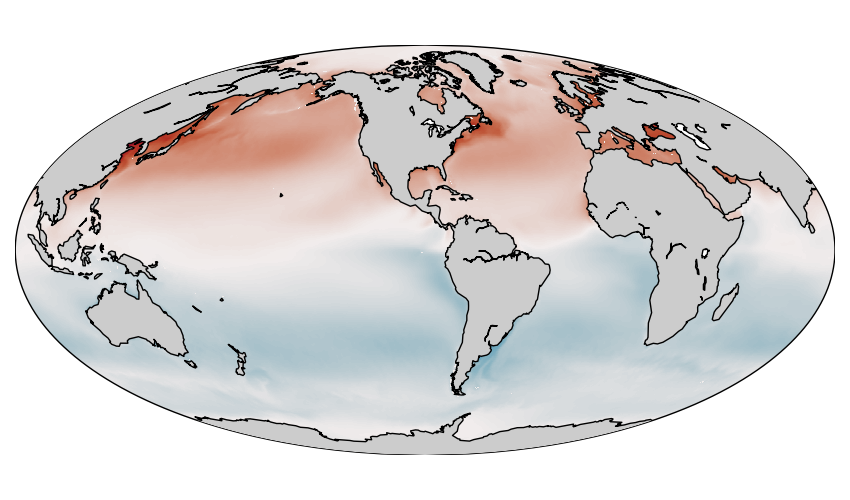} 
			\put(18,58){\color{black}{\footnotesize exact left singular vector}} 				
		\end{overpic}
		~	
		\begin{overpic}[width=0.31\textwidth]{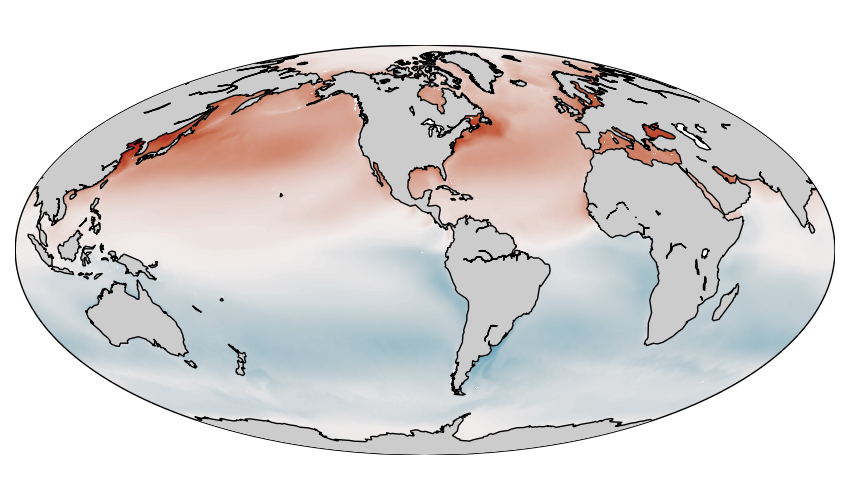} 
			\put(17,58){\color{black}{\footnotesize sketched left singular vector}}
		\end{overpic}	
	\end{subfigure}	
	
	\centering
	\begin{subfigure}{1\textwidth}	
		\centering
		\DeclareGraphicsExtensions{.pdf}
		\begin{overpic}[width=0.31\textwidth]{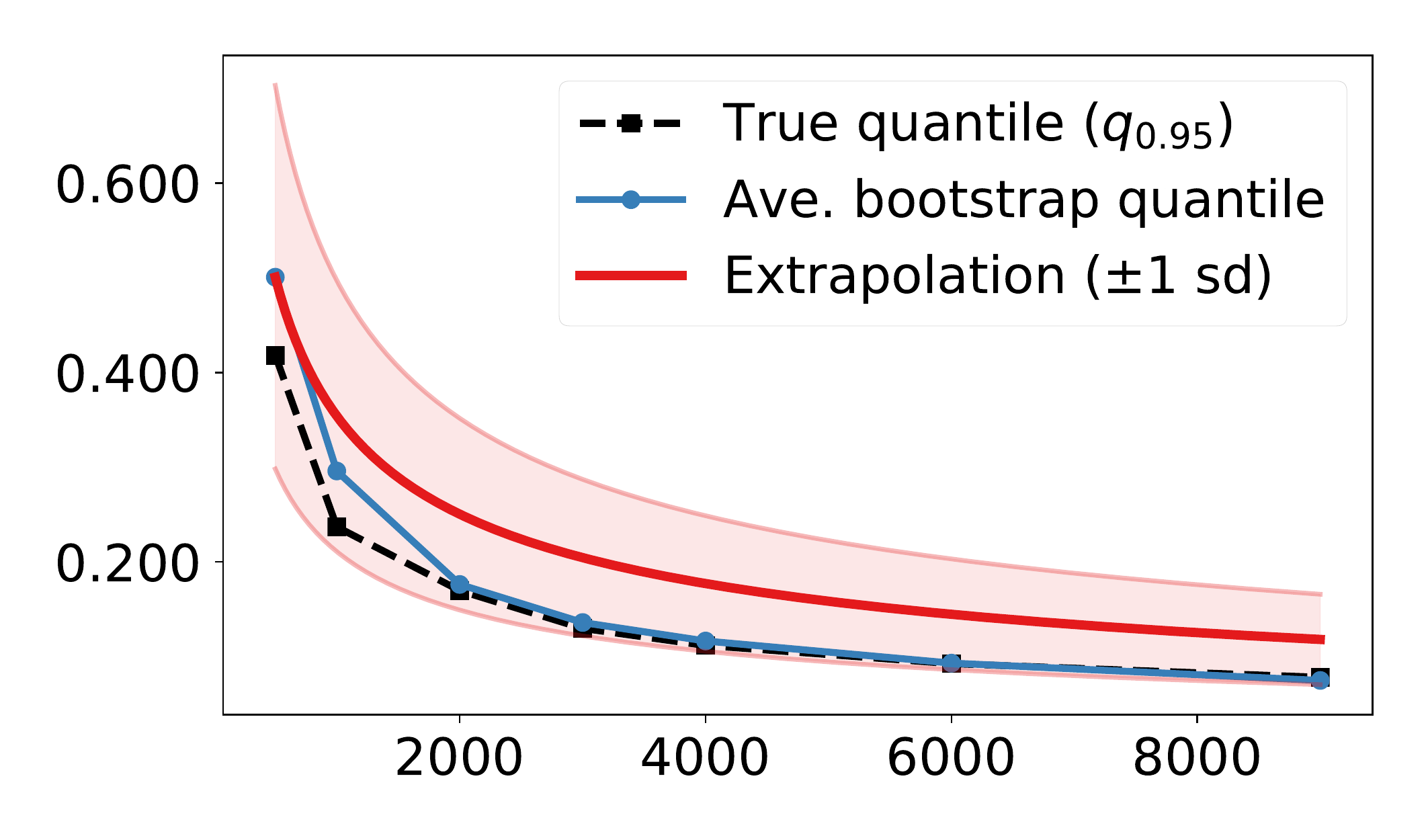} 
			\put(-6,26){\rotatebox{90}{\footnotesize $\tilde\e_{_{U}}(t)$}}
			\put(42,-2){\color{black}{\footnotesize sketch size $t$}}   
		\end{overpic}
		~
		\DeclareGraphicsExtensions{.png}
		\begin{overpic}[width=0.31\textwidth]{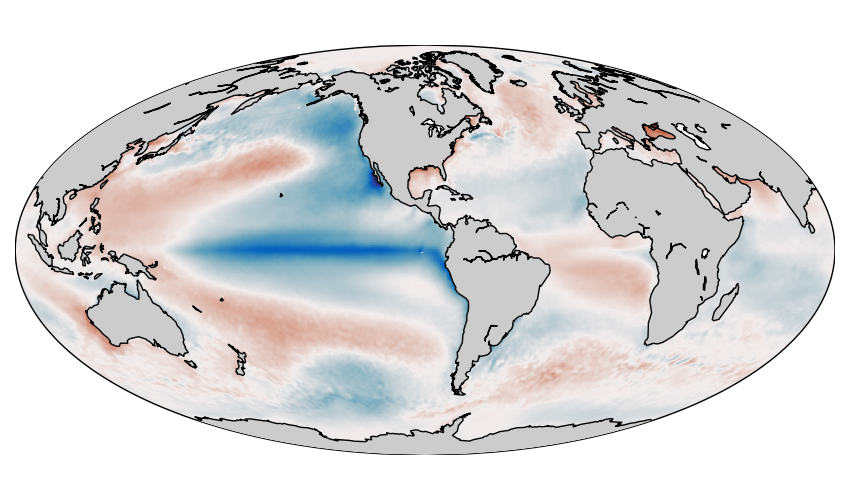} 
		\end{overpic}
		~	
		\begin{overpic}[width=0.31\textwidth]{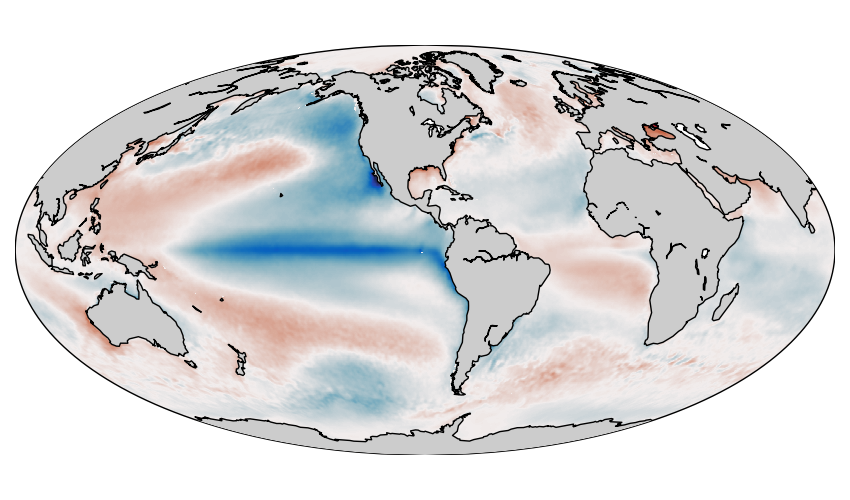} 
		\end{overpic}
	\end{subfigure}	
	
	\vspace{+.2cm}	
	\caption{Bootstrap error estimates for 1st and 4th sketched left singular vectors of the SST dataset ($691150\times 14001$), using squared-length sampling. The error variable in the top left plot corresponds to the index set $\mathcal{J}=\{1\}$ and the error variable in the bottom left plot corresponds to $\mathcal{J}=\{4\}$. The simulations involve $500$ trials and $30$ bootstraps per trial. In addition we show the deterministic mode and a single instance of a sketched mode using $t=3000$. We can see that with high probability the error (sine distance) is less than $0.002$ for $\mathcal{J}=\{1\}$, and less than $0.2$ for $\mathcal{J}=\{4\}$. Here, regions that are red indicate warm sea surface temperatures, while blue regions indicate cold temperature regions.} 
	\label{fig:results_sst_real}
\end{figure*}

\subsection{Examples from applications}\label{sec:exapp}

Now we turn to some examples arising from applications in climate science and fluid dynamics.

\paragraph{Sea surface temperature data.} 
In the analysis of sea-surface temperature (SST) data, principal components (henceforth called ``modes'') play an important role in visualizing the structure of climate patterns.
Due to the massive scale of such data, it is impractical to use classical SVD algorithms, but fortunately, it is often possible to gain clear physical insights from approximate computations. Consequently, this application is well-suited to sketching algorithms~\citep{erichson2018sparse,doi:10.1137/18M1215013}.

We consider satellite-based recordings of SST data  collected during the years 1981 to 2019, comprising $d=14001$ temporal snapshots~\citep{reynolds2007daily}. Each snapshot measures the daily temperature means at $n=691150$ spatial grid points across the globe. In total, the data requires $72$GB in storage.

The left panel of Figure~\ref{fig:results_sst_real} shows the performance of the bootstrap estimate $\hat q_{_U}(t)$, where the experiments were organized in the same way as in Section~\ref{sec:results_synthetic}, and the results are plotted in the same format. 
In addition to the fact that the extrapolated estimates are accurate, there are two other aspects of this example that are especially encouraging: (1) The intial sketch size $t_0=500$ corresponds to an extremely small fraction ($500/691150\approx 0.0007$) of the data. (2) When Algorithm~\ref{alg:bootstrap} was distributed across 30 machines, it was possible to generate $B=30$ bootstrap samples at $t_0=500$ in \emph{less than 4 seconds}. Thus, this is fast enough to provide the user with error estimates on a time scale that is compatible with \emph{interactive data analysis}.

The right panel of Figure \ref{fig:results_sst_real} gives a visual comparison between the exact and sketched versions of the 1st and 4th mode.
More specifically, the modes are visualized by projecting their $691150$ entries onto a set of geospatial coordinates. In this situation, the bootstrap would tell the user that a sketch size of $t=3000$ corresponds to a sine distance of less than $0.002$ with 95\% probability for the 1st mode, which conforms with the fact that the exact and sketched modes are nearly indistinguishable to the human eye.

\begin{figure*}[!b]
	
	\centering
	\begin{subfigure}{1\textwidth}	
		\centering
		\DeclareGraphicsExtensions{.pdf}
		\begin{overpic}[width=0.31\textwidth]{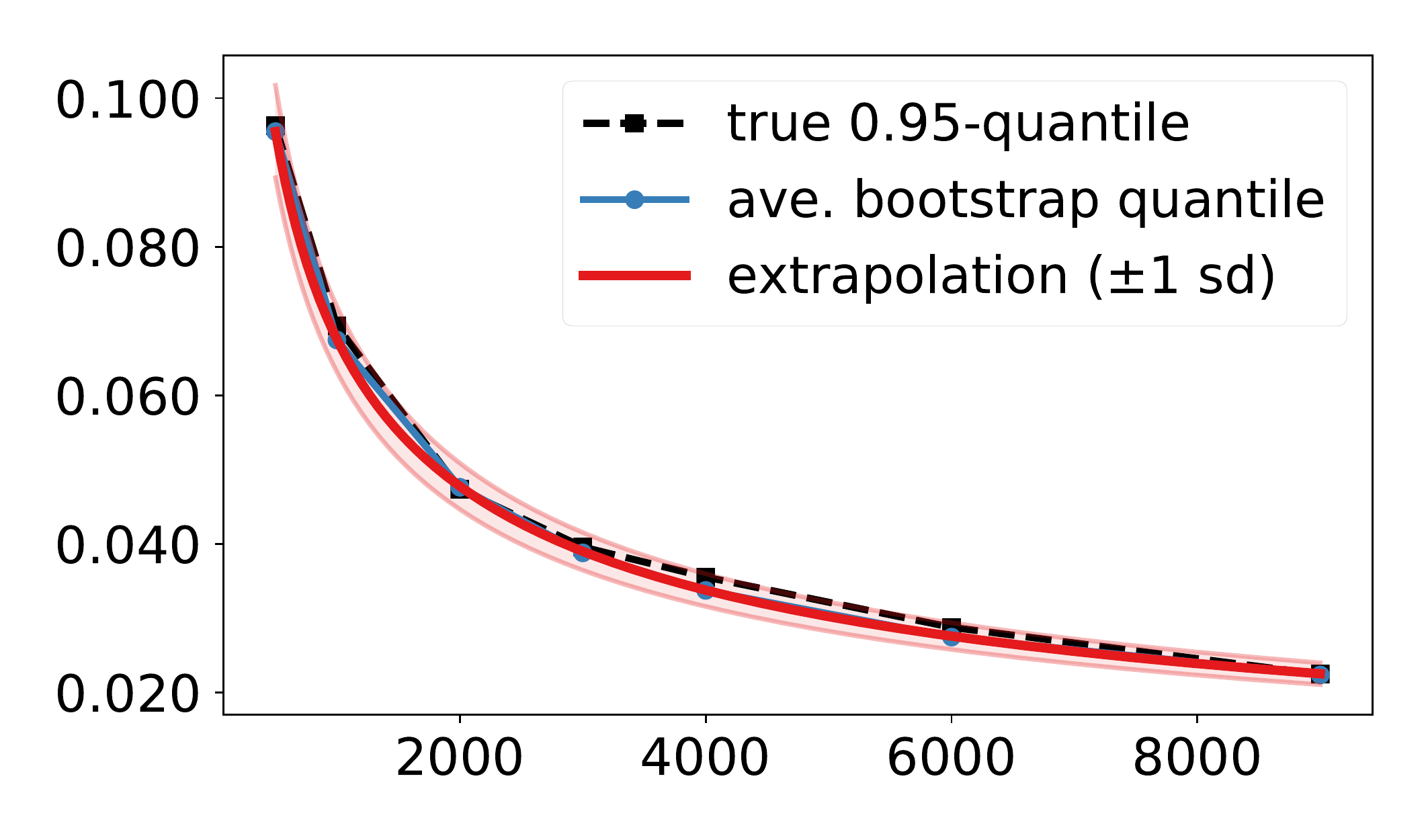} 
			\put(-6,26){\rotatebox{90}{\footnotesize $\tilde\e_{_{V}}(t)$}}
		\end{overpic}
		~
		\DeclareGraphicsExtensions{.png}
		\begin{overpic}[width=0.31\textwidth]{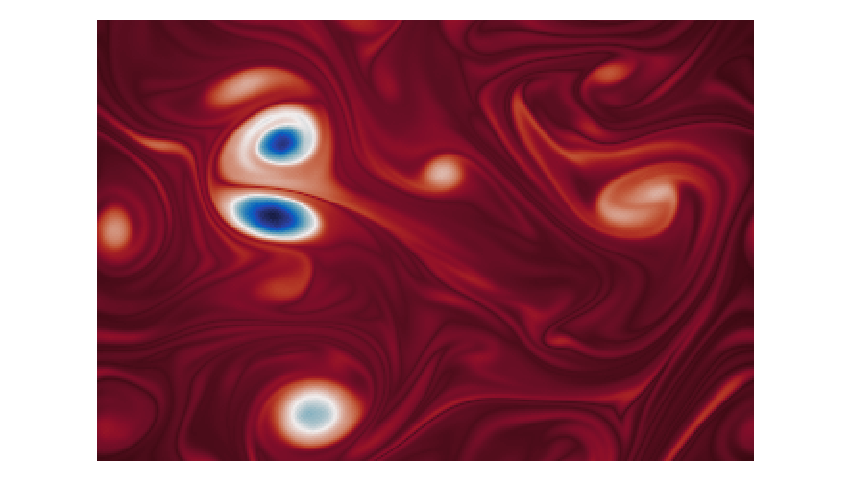} 
			\put(3,8){\rotatebox{90}{\footnotesize exact eigenvector}}				
		\end{overpic}
		~	
		\begin{overpic}[width=0.31\textwidth]{figures/turb_flow_mode_sketch_6000} 
			\put(3,5){\rotatebox{90}{\footnotesize sketched eigenvector}}			 				
		\end{overpic}	
	\end{subfigure}	
	
	\centering
	\begin{subfigure}{1\textwidth}	
		\centering
		\DeclareGraphicsExtensions{.pdf}
		\begin{overpic}[width=0.31\textwidth]{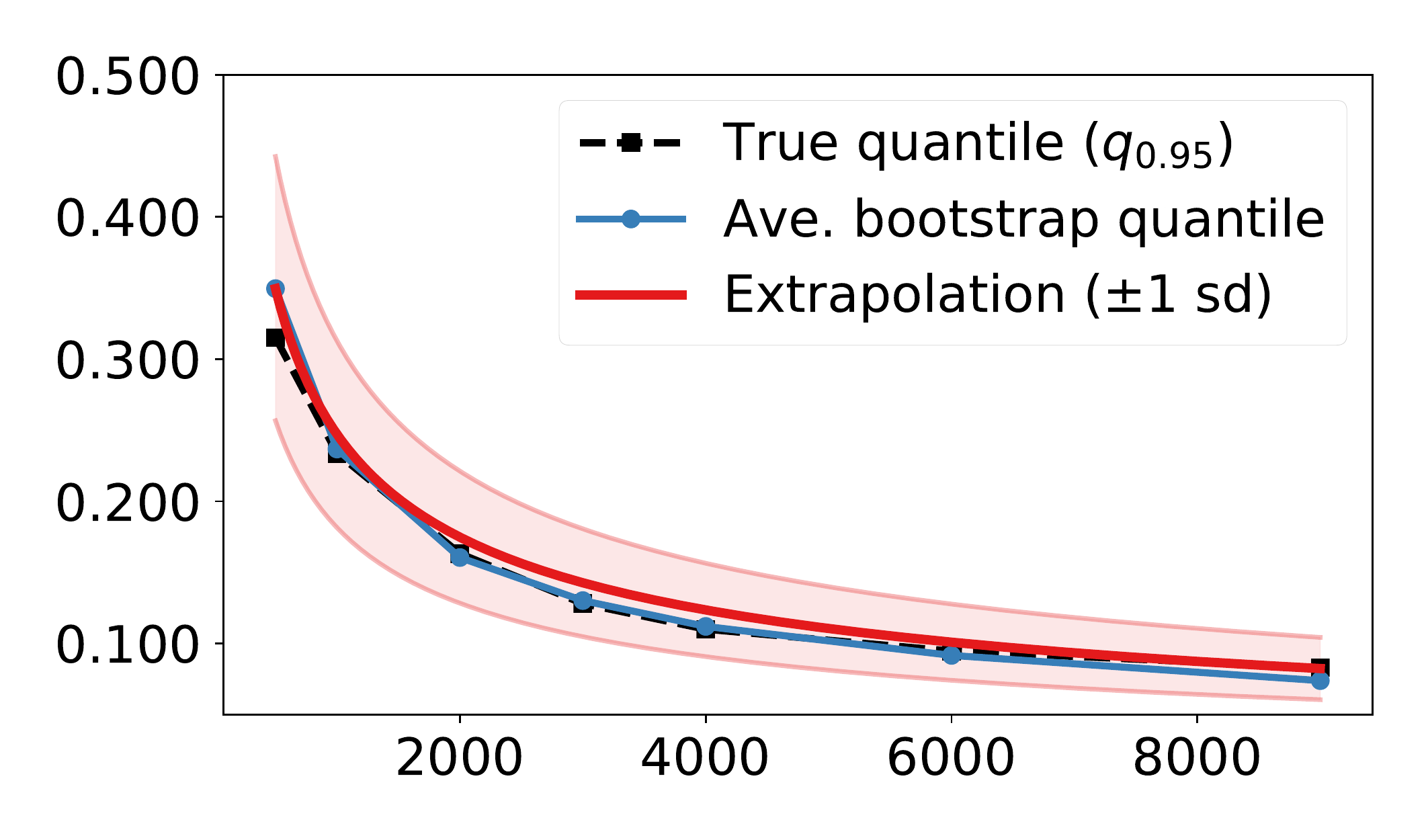} 
			\put(-6,26){\rotatebox{90}{\footnotesize $\tilde\e_{_{V}}(t)$}}
			\put(42,-2){\color{black}{\footnotesize sketch size $t$}}   
		\end{overpic}
		~
		\DeclareGraphicsExtensions{.png}
		\begin{overpic}[width=0.31\textwidth]{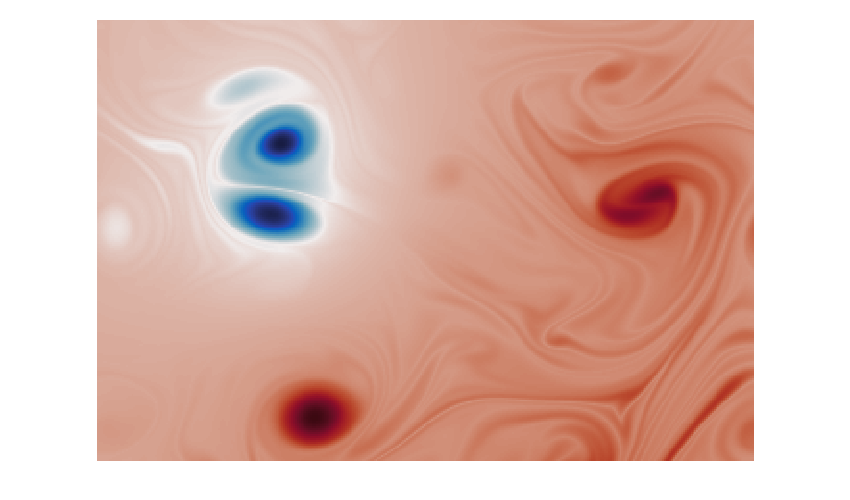} 
			\put(3,8){\rotatebox{90}{\footnotesize exact eigenvector}} 					
		\end{overpic}
		~	
		\begin{overpic}[width=0.31\textwidth]{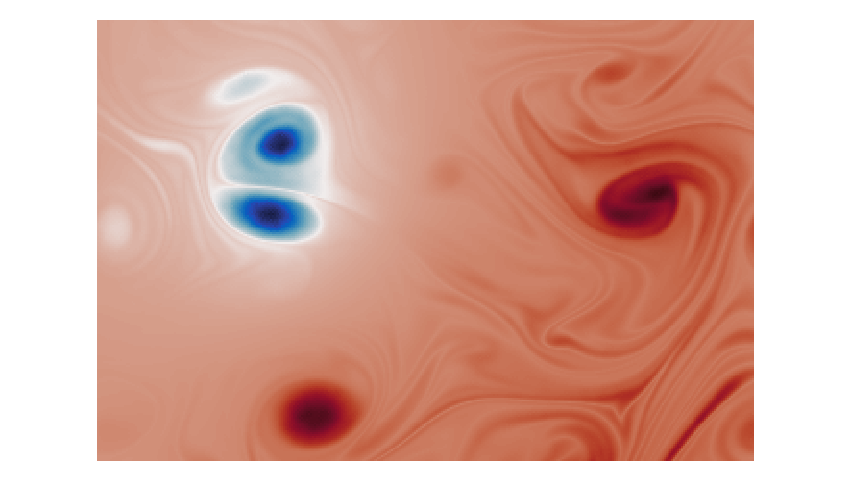} 
			\put(3,5){\rotatebox{90}{\footnotesize sketched eigenvector}}			 			
		\end{overpic}	
	\end{subfigure}	
	\vspace{+.2cm}	
	\caption{Bootstrap error estimates for the dominant two sketched eigenvectors of a dense adjacency matrix ($116964\times 116964$), using squared-length sampling. Here, the error variables correspond to the index set $\mathcal{J}=\{1\}$ (top) and $\mathcal{J}=\{2\}$ (bottom), respectively. The simulations involve $500$ trials and $30$ bootstraps per trial. In addition we show the (cropped) deterministic and sketched mode using $t=6000$. We can see that with high probability the error (sine distance) is less than $0.03$ for $\mathcal{J}=\{1\}$, and less than $0.1$ for $\mathcal{J}=\{2\}$.}
	\label{fig:results_adjacency_real}
\end{figure*}

\paragraph{Large adjacency matrices in fluid dynamics.} \ 
Computing the eigenvectors of very large adjacency matrices is a frequently encountered problem in many application domains. When these matrices are dense, eigenvector computations are especially costly, which makes sketching algorithms a natural approach.
As an illustration of this type of situation, we consider a dense symmetric adjacency matrix that encodes dynamics in a fluid flow system. In this context, the eigenvectors carry information about the strength of vortices in the system. (We refer to~\cite{bai2019randomized} for further background.)
Specifically, the adjacency matrix is of size $n\times n$ with $n = 116964$, which requires $101$GB of storage.

The left panel of Figure~\ref{fig:results_adjacency_real} shows the performance of $\hat q_{_V}(t)$, where we note that $v_1$ corresponds to the top eigenvector, since $A$ is symmetric. (The experiments here were designed in the same way as in Section~\ref{sec:results_synthetic}.)
From looking at the left panel, we see that the extrapolated bootstrap estimates are accurate over a large range of sketch sizes. Also, by distributing Algorithm~\ref{alg:bootstrap} across 30 machines, it was possible to generate $B=30$ bootstrap samples at $t_0=500$  in only 11.5 seconds (including overhead costs). Indeed, this is a remarkably short amount of time for error estimation in the context of 101GB matrix.

The right panel of Figure~\ref{fig:results_adjacency_real} gives a visual comparison of the exact and sketched eigenvectors, where blue/red correspond to strong/weak vortices. As in the case of the SST data, this comparison shows that Algorithm~\ref{alg:bootstrap} can provide the user with reliable confirmation that the sketched approximation is of high quality.

\section{Conclusion}

In this work, we developed a fully data-driven bootstrap method that numerically estimates the \emph{actual} error of sketched singular vectors/values.
From a practical standpoint, this allows the user to inspect the quality of a rough initial sketched SVD, and then adaptively predict how much extra work is needed to reach a given error tolerance.
Also, our numerical results show that the estimates are accurate for choices of $A$ in a range of conditions, including some large-scale applications related to fluid dynamics and climate science.

Computationally, our method readily scales to very large problems by taking advantage of inherent speedups based on parallelism and extrapolation. 
In fact, these speedups are so substantial that even when $A$ is on the order of 100GB, it is possible to obtain high quality error estimates \emph{within a matter of seconds} after a sketched SVD has been computed.

Theoretically, we have shown in Theorem~\ref{thm:main} that the quantile estimates $\hat q_{_U}\!(t)$, $\hat q_{_{\Sigma}}\!(t)$, and $\hat q_{_V}\!(t)$ are consistent, in the sense that as the size of the problem becomes large, they bound the error variables with a probability approaching the desired value of $1-\alpha$.

\section*{Acknowledgements}
MEL gratefully acknowledges funding support from NSF (DMS-1613218 and DMS-1915786). 
MWM would like to acknowledge DARPA, NSF, ONR, and Intel for providing partial support of this work. NBE gratefully acknowledges Amazon Web Services for supporting this project with EC2 credits. 
Further, we would like to acknowledge the NOAA for providing the SST data (\href{https://www.esrl.noaa.gov/psd/}{https://www.esrl.noaa.gov/psd/}).

\bibliographystyle{icml2020}
\bibliography{sketching.bib}

\appendix

\section{Proofs}

The main aspects of the proof of Theorem~\ref{thm:main} are presented in Section~\ref{sec:mainproof}, and the arguments in this section will refer to lower level results that are stated and proved in Sections~\ref{sec:intermediate},~\ref{sec:right},~\ref{sec:left}, and~\ref{sec:svals}. Next, in Section~\ref{sec:examples}, we provide detailed examples of matrices that satisfy both of the assumptions RP and RS. Lastly, in Section~\ref{sec:suppnum}, we present additional experimental results that go beyond the settings considered in the main~text. 

\subsection{Proof of Theorem~\ref{thm:main}}\label{sec:mainproof} 
We decompose the main parts of the proof into Sections~\ref{sec:limV},~\ref{sec:limU}, and~\ref{sec:limSigma} corresponding to the three limits (in the order of~\ref{eqn:limV},~\eqref{eqn:limU}, and~\eqref{eqn:limSigma}). 
 In addition, we provide a summary of the notation and terminology for the proofs immediately below.

\subsubsection{Notation and terminology for proofs}\label{sec:notation}

\paragraph{Items related to matrices.} For any real matrices $L$ and $M$ of the same size, we frequently use the inner product $\llangle L,M\rrangle :=\tr(L\ttop M)$. Also, for any real matrix $M$, the Frobenius norm $\|M\|_F$ is equal to $\sqrt{\tr(M\ttop M)}$, and the operator norm $\|M\|\op$ is equal to the maximum singular value of $M$. The set of symmetric matrices in $\R^{d\times d}$ is denoted $\mathcal{S}^{d\times d}$, and for any $M\in\mathcal{S}^{d\times d}$, its ordered eigenvalues are written as $\lambda_j(M)\geq \lambda_{j+1}(M)$. Likewise, the ordered singular values of a general real (possibly rectangular) matrix $R$ are denoted $\sigma_j(R)\geq \sigma_{j+1}(R)$. For a sketch of $A_n$, we write $\tilde A_n= S_nA_n$, and similarly, the matrix $\tilde A_n^*$ is defined as having rows that are sampled with replacement from the rows of $\tilde A_n$. When the context is clear, we will sometimes use the shorthand notation
\begin{equation*}
\begin{split}
\sigma_j &=\sigma_j(A_n),\\[0.2cm]
\tilde \sigma_j&=\sigma_j(\tilde A_n),\\[0.2cm]
\tilde \sigma_j^*&=\sigma_j(\tilde A_n^*).
\end{split}
\end{equation*}
Similarly, the $j$th left and right singular vectors of $\tilde A_n$ are denoted as $\tilde u_j$ and $\tilde v_j$, and likewise for $\tilde u_j^*$ and $\tilde v_j^*$ with respect to $\tilde A_n^*$. Hence, the dependence on $n$ will be generally suppressed for these vectors. In addition, for the normalized Gram matrices associated with $A_n$, $\tilde A_n$ and $\tilde A_n^*$, we define
\begin{align}\label{eqn:Gdefs}
G_n &:=\ts\frac{1}{n}A_n\ttop A_n,\\[0.2cm]
\tilde{G}_n &:=\ts\frac{1}{n}\tilde A_n\ttop \tilde A_n,\\[0.2cm]
\tilde{G}_n^* &:=\ts\frac{1}{n}(\tilde A_n^*)\ttop (\tilde A_n^*).
\end{align}
Lastly, recall that under Assumptions RP and RS, the matrix $G_n$ converges to a positive definite matrix $\mathsf{G}_{\infty}\in\R^{d\times d}$ as $n\to\infty$. Accordingly, sans-serif font will be reserved for other limiting objects, such as the leading eigenvectors $\mathsf{v}_1,\mathsf{v}_2\in\R^d$ of $\mathsf{G}_{\infty}$, as described in Assumptions RS.

\paragraph{Items related to probability.} If $Y$ is a random matrix, we write $\mathcal{L}(Y)$ to refer to its distribution, and if $Z$ is another random matrix, we write $\mathcal{L}(Y|Z)$ to refer to the conditional distribution of $Y$ given $Z$. 
If $\{Y_n\}$ is a sequence of random matrices converging in probability to another random matrix $Y_{\infty}$ as $n\to\infty$, we write either $Y_n=Y_{\infty}+o_{\P}(1)$, or $Y_n\xrightarrow{\P}Y_{\infty}$.  Next, if $Y_n$ converges to $Y_{\infty}$ in distribution, we write $\mathcal{L}(Y_n)\xrightarrow{d}\mathcal{L}(Y_{\infty})$. In addition, it is important to define a notion of convergence for conditional distributions. Specifically, if $\{Z_n\}$ is a sequence of random matrices, we will need to define the convergence of the conditional distributions $\mathcal{L}(Y_n|Z_n)$. To do this, first note that ordinary convergence in distribution can be equivalently expressed in terms of various metrics on the space of probability measures. That is, the limit $\mathcal{L}(Y_n)\xrightarrow{d}\mathcal{L}(Y_{\infty})$ is equivalent to $\varrho(\mathcal{L}(Y_n),\mathcal{L}(Y_{\infty}))\to 0$, where $\varrho$ is a metric such as the L\'evy-Prohorov metric, or the bounded Lipschitz metric~(cf.~\citep[Sec.~11.3]{Dudley}). Likewise, for conditional distributions we write `$\mathcal{L}(Y_n|Z_n)\xrightarrow{ \ d \ }\mathcal{L}(Y_{\infty})$ in probability' if the sequence of scalar random variables $\{\varrho(\mathcal{L}(Y_n|Z_n),\mathcal{L}(Y_{\infty}))\}$ converges to 0 in probability.

\subsubsection{Proof of the limit~\eqref{eqn:limV}}\label{sec:limV} We begin with a reduction that is often used in the literature on bootstrap methods. Specifically, it is known that~\eqref{eqn:limV} can be reduced to showing that
\begin{equation}
\mathcal{L}\Big(\sqrt{t_n}\max_{j\in\mathcal{J}} \rho_{\sin}(\tilde v_j,v_j)\Big) \xrightarrow{ \ d \ } \mathcal{L}(\xi_{_V}),\label{eqn:reduceV}
\end{equation}
and
\begin{equation}
\mathcal{L}\Big(\sqrt{t_n} \max_{j\in\mathcal{J}} \rho_{\sin}(\tilde v_j^*,\tilde v_j)\Big| S_n\Big) \xrightarrow{ \ d \ } \mathcal{L}(\xi_{_V}) \text{ \ \ in probability},\label{eqn:reduceVboot}
\end{equation}
for some random variable $\xi_{_V}$\! whose distribution function is continuous. (For further details, please see Theorem 1.2.1, as well as Remark 1.2.1, and the discussion on p.5 of the book~\cite{Politis:1999}.) Next, as a step towards showing the limits~\eqref{eqn:reduceV} and~\eqref{eqn:reduceVboot}, we will use some algebraic identities involving the projection matrices associated with $v_j$, $\tilde v_j$, and $\tilde v_j^*$, which we denote as 
\begin{align}\label{eqn:Pdefs}
P_j&:=v_jv_j\ttop, \ \ \ \ \ \ \tilde P_j:=\tilde v_j\tilde v_j\ttop,  \ \ \ \ \  \tilde P_j^*:=(\tilde v_j^*)(\tilde v_j^*)\ttop,
 \end{align}
 where the fact that these matrices depend on $n$ has been suppressed.
 The relevant identities are
\begin{align}
\sqrt{t_n}\rho_{\sin}(\tilde v_j, v_j) &
\ = \ \ts\frac{1}{\sqrt{2}}\big\|\sqrt{t_n}(\tilde P_j-P_j)\big\|_F,\label{eqn:Vident}\\[0.2cm]
\sqrt{t_n}\rho_{\sin}(\tilde v_j^*, \tilde v_j)  & 
\ = \ \ts\frac{1}{\sqrt{2}}\big\|\sqrt{t_n}(\tilde P_j^*-\tilde P_j)\big\|_F.\label{eqn:Videntboot}
\end{align}
As a consequence of these identities, we may write
\begin{align}\label{eqn:functionalident}
 \sqrt{t_n}\max_{j\in\mathcal{J}} \rho_{\sin}(\tilde v_j,v_j) & \ = \  f\Big(\sqrt{t_n}(\tilde P_1-P_1),\dots,\sqrt{t_n}(\tilde P_k-P_k)\Big),\\[0.2cm]
  \sqrt{t_n}\max_{j\in\mathcal{J}} \rho_{\sin}(\tilde v_j^*,\tilde v_j) & \ = \ f\Big(\sqrt{t_n}(\tilde P_1^*-\tilde P_1),\dots,\sqrt{t_n}(\tilde P_k^*-\tilde P_k)\Big)
 \end{align}
 where $f:(\R^{d\times d})^k\to \R$ is defined by $f(C_1,\dots,C_k)=\max_{j\in\mathcal{J}}\ts\frac{1}{\sqrt 2}\|C_j\|_F$. In turn, by the continuous mapping theorem and the Cram\'er-Wold theorem~\citep[Theorem 3.27 and Corollary 4.5]{Kallenberg}), the limits~\eqref{eqn:reduceV} and~\eqref{eqn:reduceVboot} will hold if we can show that for any fixed matrices $M_1,\dots,M_k\in\R^{d\times d}$, there is an associated Gaussian random vector, say $(Z_1(M_1),\dots,Z_k(M_k))\in\R^k$, such that
 \begin{align}
 \mathcal{L}\Big(\llangle\sqrt{t_n}(\tilde P_1-P_1),M_1\rrangle,\dots,\llangle \sqrt{t_n}(\tilde P_k-P_k),M_k\rrangle\Big) & \ \xrightarrow{ \ d \ } \ \mathcal{L}(Z_1(M_1),\dots,Z_k(M_k)), \text{ \ \ and }\label{eqn:initialprojclt}\\[0.2cm]
   \mathcal{L}\Big(\llangle\sqrt{t_n}(\tilde P_1^*-\tilde P_1),M_1\rrangle,\dots,\llangle \sqrt{t_n}(\tilde P_k^*-\tilde P_k),M_k\rrangle\Big| S_n\Big) & \ \xrightarrow{ \ d \ } \ \mathcal{L}(Z_1(M_1),\dots,Z_k(M_k)) \text{ \ \ \  in probability}.
  \end{align}
These limits are established in Lemmas~\ref{lem:projclt} and~\ref{lem:bootprojclt} below, where we handle certain key technical challenges. In addition, we must verify the condition that the limiting random variable $\xi_V$ in~\eqref{eqn:reduceV} and~\eqref{eqn:reduceVboot} has a continuous distribution function. To do this, first note that the limit~\eqref{eqn:initialprojclt} allows us to view the tuple of matrices $\big(\sqrt{t_n}(\tilde P_j-P_j)\big)_{j\in\mathcal{J}}$ as converging in distribution to a Gaussian vector in the space $(\R^{d\times d})^{|\mathcal{J}|}$. Also, it is a basic fact that the norm of a Gaussian vector with a non-zero covariance matrix yields a random variable whose distribution function is continuous. So, given that the function $f$ restricts to a norm on $(\R^{d\times d})^{|\mathcal{J}|}$, it suffices to show that the mentioned Gaussian vector in $(\R^{d\times d})^{|\mathcal{J}|}$ has positive variance when projected into at least one direction. In other words, to show that $\xi_V$ has a continuous distribution function, it is enough to show that there is at least one index $j\in\mathcal{J}$ and matrix $M_j\in\R^{d\times d}$ such that $\llangle \sqrt{t_n}(\tilde P_j-P_j),M_j\rrangle$ has a limiting Gaussian distribution with positive variance --- and this is handled in Lemma~\ref{lem:projclt}. Altogether, this completes the proof of the first limit~\eqref{eqn:limV} in Theorem~\ref{thm:main}.\\

\noindent\emph{Remark.} The proofs of the second and third limits~\eqref{eqn:limU} and~\eqref{eqn:limSigma} will require different versions of the Lemmas~\ref{lem:projclt} and~\ref{lem:bootprojclt}, and these are given later on in Lemmas~\ref{lem:Uclt} and~\ref{lem:svals}.  \qed

\subsubsection{Proof of the limit~\eqref{eqn:limU}}\label{sec:limU}
By the reduction argument used at the beginning of Section~\ref{sec:limV}, it suffices to show that
\begin{equation}
\mathcal{L}\Big(\sqrt{t_n}\max_{j\in\mathcal{J}} \rho_{\sin}(\tilde u_j,u_j)\Big) \xrightarrow{ \ d \ } \mathcal{L}(\xi_{_U}),\label{eqn:reduceU}
\end{equation}
and
\begin{equation}
\mathcal{L}\Big(\sqrt{t_n} \max_{j\in\mathcal{J}} \rho_{\sin}(\tilde A_n \tilde v_j^*,\tilde A_n \tilde v_j)\Big| S_n\Big) \xrightarrow{ \ d \ } \mathcal{L}(\xi_{_U}) \text{ \ \ in probability},\label{eqn:reduceUboot}
\end{equation}
for some random variable $\xi_{_U}$\! whose distribution function is continuous. Next, to develop a counterparts of the relation~\eqref{eqn:Vident}, define the projection matrices 
$$\Pi_j:=u_ju_j\ttop = \ts \frac{A_n P_jA_n\ttop}{\tr( P_j A_n\ttop A_n)}\text{ \ \ \ \  and \ \ \ \ } \tilde\Pi_j:=\tilde u_j\tilde u_j\ttop =\ts \frac{A_n \tilde P_jA_n\ttop}{\tr(\tilde P_j A_n\ttop A_n)}.$$
Similarly, to develop a counterpart of~\eqref{eqn:Videntboot}, define the vectors $\breve u_j:=\ts\frac{\tilde A \tilde v_j}{\|\tilde A\tilde v_j\|_2}$ and $\breve u_j^*:=\ts\frac{\tilde A \tilde v_j^*}{\|\tilde A\tilde v_j^*\|_2}$, and their associated projections
$$ \ \ \ \ \ \breve\Pi_j := \ \breve u_j\breve u_j\ttop \, = \, \ts\frac{\tilde A_n\tilde P_j\ttop \tilde A_n\ttop}{\tr(\tilde P_j \tilde A_n\ttop \tilde A_n)} 
\text{ \ \ \ \ \ and \ \ \ \ \ }
 \breve\Pi_j^*:= \ (\breve u_j^*)(\breve u_j^*)\ttop 
 \, = \, \ts\frac{\tilde A\tilde P_j^*\tilde A_n\ttop}{\tr(\tilde P_j^* \tilde A_n\ttop\tilde A_n)}.
 $$
\emph{Remark.}  For a finite $n$, it is possible that the denominators $\tr(\tilde P_jA_n\ttop A_n)$, $\tr(\tilde P_j\tilde A\ttop \tilde A)$, or $\tr(\tilde P_j^*\tilde A_n\ttop \tilde A_n)$ may be zero, and if this occurs, we instead define $\tilde \Pi_j$, $\breve \Pi_j$, or $\breve\Pi_j^*$ to be the zero matrix. However, the probability of such events will turn out to go to zero asymptotically, and hence, such events will be unimportant. This same type of consideration will occur at other points in the proofs, and so in order to avoid repetition, we will not make further mention of zero denominators that occur with vanishing probability as $n\to\infty$.\\

 In the above notation, it is straightforward to check the identities
 \begin{align}
 \rho_{\sin}(\tilde u_j,u_j) & \ = \ \ts\frac{1}{\sqrt 2} \|\tilde \Pi_j-\Pi_j\|_F \\[0.2cm]
 \rho_{\sin}(\breve u_j^*,\breve u_j) & \ = \  \ts\frac{1}{\sqrt 2} \|\breve\Pi_j^*-\breve\Pi_j\|_F.
 \end{align}
 Since the Frobenius norm of a symmetric matrix only depends on the non-zero eigenvalues, we may replace the matrices $(\tilde \Pi_j-\Pi_j)$ and $(\tilde \Pi_j^*-\tilde\Pi_j)$ above  with different matrices whose non-zero eigenvalues are the same. In particular, the matrix $A_n MA_n\ttop$ has the same non-zero eigenvalues as $(A_n\ttop A_n)^{1/2}M (A_n\ttop A_n)^{1/2}$ for any $M\in\R^{d\times d}$. So, if we recall the definition $G_n=\ts\frac{1}{n}A_n\ttop A_n$ from~\eqref{eqn:Gdefs}, it follows that the matrix
\begin{equation}\label{eqn:deltajdef}
 \tilde \Delta_j:= G_n^{1/2}\Big(\ts\frac{\tilde P_j}{\tr(\tilde P_j G_n)}-\ts\frac{ P_j}{\tr( P_jG_n)}\Big)G_n^{1/2}
 \end{equation}
has the same non-zero eigenvalues as $(\tilde \Pi_j-\Pi_j)$, and therefore
  \begin{equation}\label{eqn:frobident}
  \rho_{\sin}(\tilde u_j,u_j) \ = \ \ts\frac{1}{\sqrt 2}\|\tilde\Delta_j\|_F.
 \end{equation}	 
Similarly, if we recall the definition $\tilde G_n=\ts\frac{1}{n}\tilde A_n\ttop \tilde A_n$ and define 
\begin{equation}\label{eqn:deltajstardef}
\tilde\Delta_j^*:=\tilde{G}_n^{1/2}\Big(\ts\frac{\tilde P_j^*}{\tr(\tilde P_j^* \tilde{G}_n)}-\ts\frac{\tilde P_j}{\tr(\tilde P_j\tilde{G}_n)}\Big)\tilde{G}_n^{1/2},
\end{equation}
then we have
  \begin{equation}\label{eqn:frobidentboot}
  \rho_{\sin}(\breve u_j*,\breve u_j) \ = \ \ts\frac{1}{\sqrt 2}\| \tilde\Delta_j^*\|_F.
 \end{equation}	 
 The key significance of working with the $d\times d$ matrices $\tilde\Delta_j$ and $\tilde\Delta_j^*$ is that they remain of a fixed size asymptotically, whereas the $n\times n$ matrices $(\tilde\Pi_j-\Pi_j)$ and $(\breve\Pi_j^*-\breve \Pi_j)$ expand as $n\to\infty$.
 
  At this stage, the identities~\eqref{eqn:frobident} and~\eqref{eqn:frobidentboot} will play the role that~\eqref{eqn:Vident} and~\eqref{eqn:Videntboot} did earlier. In turn, we may apply the previous reasoning based on the continuous mapping theorem and  the Cram\'er-Wold theorem. In this way, the limits~\eqref{eqn:reduceU} and~\eqref{eqn:reduceUboot} will hold if we can show that for any fixed matrices $M_1,\dots,M_k\in\R^{d\times d}$, there is an associated Gaussian vector, say $(\zeta_1(M_1),\dots,\zeta_k(M_k))\in\R^k$, such that
 \begin{align}
 \footnotesize
 \mathcal{L}\Big(\llangle[\big] \sqrt{t_n} \tilde\Delta_1,M_1\rrangle[\big],\dots,\llangle[\big] \sqrt{t_n}\tilde\Delta_k,M_k\rrangle[\big]\Big) & \ \xrightarrow{ \ d \ } \ \mathcal{L}(\zeta_1(M_1),\dots,\zeta_k(M_k)), \text{ \ \ and }\label{eqn:limdelta}\\[0.2cm]
   \mathcal{L}\Big(\sqrt{t_n}\tilde\Delta_1^*,M_1\rrangle,\dots,\llangle \sqrt{t_n}\tilde\Delta_k^*,M_k\rrangle\Big| S_n\Big) & \ \xrightarrow{ \ d \ } \ \mathcal{L}(\zeta_1(M_1),\dots,\zeta_k(M_k)) \text{ \ \ \  in probability}.\label{eqn:limdeltastar}
  \end{align}
  These limits are established in Lemma~\ref{lem:Uclt} below.
Lastly, to ensure that the limiting random variable $\xi_{_U}$ in~\eqref{eqn:reduceU} and~\eqref{eqn:reduceUboot} has a continuous distribution function, the reasoning in Section~\ref{sec:limV} shows that it is sufficient to exhibit at least one index $j\in\mathcal{J}$ and matrix $M_j\in\R^{d\times d}$ such that $\var(\zeta_j(M_j))>0$. This is also done in Lemma~\ref{lem:Uclt}.\qed

\subsubsection{Proof of the limit~\eqref{eqn:limSigma}}\label{sec:limSigma}
As in the previous two subsections, the proof can be reduced to showing that
\begin{align}
\mathcal{L}\Big(\ts\frac{\sqrt{t_n}}{\sqrt n}\displaystyle\max_{j\in\mathcal{J}} |\tilde \sigma_j-\sigma_j|\Big) & \ \xrightarrow{ \ d \ } \ \mathcal{L}(\xi_{_{\Sigma}}), \text{ \ \  and}\label{eqn:reduceSigma}\\[0.2cm]
\mathcal{L}\Big(\ts\frac{\sqrt{t_n}}{\sqrt n}\displaystyle \max_{j\in\mathcal{J}}|\tilde \sigma_j^*-\tilde\sigma_j|\,\Big| S\Big) & \ \xrightarrow{ \ d \ } \ \mathcal{L}(\xi_{_{\Sigma}}), \text{ \ \ in probability}\label{eqn:reduceSigmaboot}
\end{align}
for some random variable $\xi_{_{\Sigma}}$ whose distribution function is continuous. (Here, we use the normalizing factor $\ts\frac{\sqrt{t_n}}{\sqrt n}$, rather than the $\sqrt{t_n}$ used in~\eqref{eqn:reduceV} and~\eqref{eqn:reduceVboot}, because the singular values depend on the scaling of the matrix $\ts\frac{1}{n}A_n\ttop A_n$ --- whereas the singular vectors do not.)
Proceeding as before, the continuous mapping theorem and the Cram\'er-Wold theorem imply that~\eqref{eqn:reduceSigma} and~\eqref{eqn:reduceSigmaboot} will hold if we can show that the following limits hold for any constants $c_1,\dots,c_k\in\R$,
\begin{equation}
\mathcal{L}\Big(\sum_{j=1}^k \ts\frac{\sqrt{t_n}}{\sqrt n} c_j(\tilde \sigma_j-\sigma_j)\Big) \ \xrightarrow{ \ d \ } \mathcal{L}(\zeta(c_1,\dots,c_k)),
\end{equation}
and 
\begin{equation}
\mathcal{L}\Big(\tsum_{j=1}^k\ts\frac{\sqrt{t_n}}{\sqrt n} c_j(\sigma_j(\tilde A_n^*)-\sigma_j(\tilde A_n)) \, \Big|\,S_n\Big) \ \xrightarrow{ \ d \ } \ \mathcal{L}(\zeta(c_1,\dots,c_k)) \text{ \ \  in probability},
\end{equation}
where $\zeta(c_1,\dots,c_k)$ is a Gaussian scalar random variable. These limits are established in Lemma~\ref{lem:svals}. In addition, we can show that $\xi_{_{\Sigma}}$ has a continuous distribution function in the same way as was done for $\xi_{_V}$ and $\xi_{_U}$, which amounts to showing that $\zeta(1,0,\dots,0)$ has positive variance --- and this is shown in Lemma~\ref{lem:svals} as well. This completes the proof.\qed

\subsection{Intermediate results}\label{sec:intermediate}

The results in this section are relevant to the proofs of all three limits in Theorem~\ref{thm:main}. The first  is a well known result, usually called Slutsky's lemma~\citep[Lemma 2.8]{vanderVaart:2000}, whereas the second is a conditional version of it that is tailored to the current paper. Hence, we only provide a proof of the conditional version. Lastly, in Lemmas~\ref{lem:coreclt} and~\ref{lem:corebootclt}, we provide a CLT for $\sqrt{t_n}(\tilde{G}_n-G_n)$, as well as its bootstrap counterpart $\sqrt{t_n}(\tilde{G}_n^*-\tilde{G}_n)$.

\begin{fact}[Slutsky's lemma]\label{lem:slutsky}
For each $n\geq 1$, let $T_n\in\R^{d_1\times d_2}$ and $R_n\in\R^{d_1'\times d_2'}$ be random matrices whose dimensions remain fixed as $n\to\infty$. In addition, suppose there is a random matrix $T_{\infty}\in\R^{d_1\times d_2}$ and a constant matrix $R_{\infty}\in\R^{d_1'\times d_2'}$ such that
$ \mathcal{L}(T_n) \xrightarrow{ \ d \ } \mathcal{L}(T_{\infty}) $
and
$R_n\to R_{\infty} \text{ in probability}.$
Then, for any continuous function $g:\R^{d_1\times d_2}\times \R^{d_1'\times d_2'}\to \R$, the following limit holds
\begin{equation}
\mathcal{L}\big(g(T_n,R_n)\big) \ \xrightarrow{ \ d \ } \ \mathcal{L}\big(g(T_{\infty},R_{\infty})\big).
\end{equation}
\end{fact}

\begin{lemma}[Conditional Slutsky's lemma]\label{lem:condslutsky}
For each $n\geq 1$, let $\mathcal{D}_n=\{X_{1,n},\dots,X_{t_n,n}\}$ be a set of random variables, and let $\mathcal{D}_n^*=\{X_{1,n}^*,\dots,X_{t_n,n}^*\}$ be sampled with replacement from $\mathcal{D}_n$. Also, for each $n\geq1$, let $T_n^*=T_n(\mathcal{D}_n^*)$ be a real random matrix of size $d_1\times d_2$ computed from~$\mathcal{D}_n^*$. In addition, let $ R_n\in\R^{d_1'\times d_2'}$ be a random matrix that may depend on both $\mathcal{D}_n$ and $\mathcal{D}_n^*$. Lastly, suppose that there is a random matrix $T_{\infty}\in\R^{d_1\times d_2}$ and a constant matrix $R_{\infty}\in\R^{d_1'\times d_2'}$ such that
\begin{equation}\label{eqn:slutsky1}
 \mathcal{L}(T_n^*|\mathcal{D}_n) \ \xrightarrow{ \ d \ } \ \mathcal{L}(T_{\infty}) \ \ \text{ in probability},
 \end{equation}
and for any $\e>0$,
\begin{equation}\label{eqn:slutsky2}
\P\big( \|R_n-R_{\infty}\|_F>\e\, \big|\mathcal{D}_n\big) \ \to \ 0 \text{ \ \ in probability}.
\end{equation}
Then, for any continuous function $g:\R^{d_1\times d_2}\times \R^{d_1'\times d_2'}\to \R$, the following limit holds
\begin{equation}
\mathcal{L}\big(g(T_n^*,R_n)\big|\mathcal{D}_n\big) \ \xrightarrow{ \ d \ } \ \mathcal{L}\big(g(T_{\infty},R_{\infty})\big) \ \text{ \ \ in probability}.
\end{equation}
\end{lemma}

\proof
By the continuous mapping theorem, it suffices to show that the bounded-Lipschitz metric between $\mathcal{L}\big(T_n^*,R_n\big|\mathcal{D}_n\big)$ and $\mathcal{L}\big(T_{\infty},R_{\infty}\big)$ converges to 0 in probability. (Please see the comments in Section~\ref{sec:notation} for additional background.)
Let $\mathcal{F}$ denote the class of functions from $\R^{d_1\times d_2}\times \R^{d_1'\times d_2'}$ to $\R$ that are bounded in magnitude by 1 and are 1-Lipschitz with respect to the Frobenius norm. Then,
\begin{equation}\label{eqn:split}
\begin{split}
\sup_{f\in\mathcal{F}}\Big|\E[f(T_n^*,R_n)|\D_n)] - \E[f(T_{\infty},R_{\infty})]\Big| & \ \leq \ \sup_{f\in\mathcal{F}}\Big|
\E\big[f(T_n^*,R_n) - f(T_n^*,R_{\infty})\big|\D_n\big]\Big|\\[0.2cm]
& \  \ \ \ \ \ \ \ \  \ \ \ \ \ + \ \sup_{f\in\mathcal{F}}\Big|\E[f(T_n^*,R_{\infty})|\D_n)] - \E[f(T_{\infty},R_{\infty})]\Big| .
\end{split}
\end{equation}
Regarding the second term on the right side, note that for each $f\in\mathcal{F}$, the associated function $h(\cdot):=f(\cdot,R_{\infty})$ on $\R^{d_1\times d_2}$ is bounded in magnitude by 1 and is 1-Lipschiz with respect to the Frobenius norm. Hence, the assumption~\eqref{eqn:slutsky1} implies that the second term in the bound~\eqref{eqn:split} converges to 0 in probability.

Regarding the first term of the bound~\eqref{eqn:split}, we can decompose the expectation by writing the constant $1$ as a sum of two indicators, $1=1_{E_n}+1_{E_n^c}$, where we define the event $E_n:=\{\|(T_n^*,R_n)-(T_n^*,R_{\infty})\|_F>\e\}$. Likewise, by noting that $E_n$ is the same as $\{\|R_n-R_{\infty}\|_F>\e\}$, we have
\begin{equation}
\begin{split}
\sup_{f\in\mathcal{F}}\Big|
\E[f(T_n^*,R_n)|\D_n)] - \E[f(T_n^*,R_{\infty})|\D_n]\Big|  & \ \leq \ \e+2\,\P\big(\|R_n-R_{\infty}\|_F>\e \big| \D_n\big)\\[0.1cm]
& \ = \ \e+o_{\P}(1),
\end{split}
\end{equation}
where the second step follows from Assumption~~\eqref{eqn:slutsky2}. Finally, since $\e>0$ can be taken arbitrarily small, this completes the proof. \qed

~\\

\noindent\emph{Remark.} The proofs of the next two results are similar to the proofs of Lemmas 3 and 4 in~\cite{Lopes:arxiv:2018}, but there is an important distinction insofar as the current proofs handle row-sampling matrices --- which were not addressed in that prior work.

\begin{lemma}\label{lem:coreclt}
Suppose that the conditions of Theorem~\ref{thm:main} hold.  Then, for any fixed matrix $M\in\R^{d\times d}$, there is a Gaussian random variable $Z(M)$ such that
$$\mathcal{L}\Big(\llangle M, \sqrt{t_n}(\tilde{G}_n -G_n)\rrangle\Big) \ \xrightarrow{ \ d \ } \ \mathcal{L}(Z(M)).$$
\end{lemma}

\noindent \emph{Proof}. Let $s_{1,n},\dots,s_{t_n,n}\in\R^n$ denote the rows of $\sqrt{t_n}S_n$, and observe the algebraic relation
\begin{equation}
 \llangle M, \sqrt{t_n}(\tilde{G}_n -G_n)\rrangle = \ts\frac{1}{\sqrt{ t_n} \,n}\displaystyle\sum_{i=1}^{t_n} \Big(\llangle M, A_n\ttop s_is_i\ttop  A_n \rrangle -\llangle M, A_n\ttop A_n\rrangle\Big).
\end{equation}
Hence, if we define $X_{i,n}=\ts\frac{1}{n}\Big(s_{i,n}\ttop A_n MA_n\ttop s_{i,n}-\llangle M, A_n\ttop A_n\rrangle\Big)$ for each $i\in\{1,\dots,t_n\}$, then it is straightforward to check that these random variables have mean zero and satisfy
\begin{equation}
 \llangle M, \sqrt{t_n}(\tilde{G}_n -G_n)\rrangle \ = \ \ts\frac{1}{\sqrt{t_n}}\displaystyle\sum_{i=1}^{t_n} X_{i,n}.
 \end{equation}
Since the variables $X_{1,n},\dots,X_{t_n,n}$ are i.i.d.~for each $n$, but have distributions that may vary with $n$, we now apply the Lindeberg CLT for triangular arrays~\citep[Prop.~2.27]{vanderVaart:2000}. This result requires us to verify two conditions as $n\to\infty$. The first is that $\var(X_{1,n})$ converges to a finite limit, and the second is that
\begin{equation}\label{eqn:lind1}
\E\Big[X_{1,n}^21\big\{|X_{1,n}|>\e\sqrt{t_n}\big\}\Big] \ \to \ 0 \text{ \ \ \ \ for every fixed $\e>0$}.
\end{equation}

 We will now show that $\var(X_{1,n})$ converges to a limit separately in the cases where $S_n$ is a row-sampling matrix or a Gaussian random projection. In the row-sampling case, this follows directly from Assumption RS and the fact that
 $$\var(X_{1,n})=\var\Big(\ts\frac{1}{n}s_{1,n}\ttop A_n M A_n\ttop s_{1,n}\Big) = \var(\tilde r_n\ttop M \tilde r_n),$$
 where $\tilde r_n$ is the first row of $\ts\frac{1}{\sqrt n}\tilde A_n$.
Meanwhile, in the case of Gaussian random projections, it is possible to explicitly calculate $\var(X_{1,n})$. Specifically, if $z\sim N(0,I_n)$ is a standard Gaussian vector and $Q\in\R^{n\times n}$ is fixed, then 
\begin{equation}\label{eqn:gaussianqform}
\var(z\ttop Q z)=2\|Q\|_F^2,
\end{equation} which can be found in~\citep[eqn.~9.8.6]{Bai:Silverstein:2010}. Consequently, we have
\begin{equation}
\begin{split}
 \var(X_{1,n}) 
 & \ = \ 2\big\| \ts\frac{1}{n}A_n M A_n\ttop \big\|_F^2\\[0.2cm]
 & \ = \ 2\tr(M\ttop G_n M G_n),
\end{split}
\end{equation}
and since Assumption RP ensures $G_n\to \mathsf{G}_{\infty}$, it follows that $\var(X_{1,n})\to \tr(M\ttop \mathsf{G}_{\infty} M\mathsf{G}_{\infty})$, as needed.\\

To complete the proof, it remains to check the Lindeberg condition~\eqref{eqn:lind1} in the cases of the two types of sketching matrices. First we handle the case of row sampling. Using the Cauchy-Schwarz inquality, followed by a Chernoff bound, we have
\begin{equation}\label{eqn:cs}
\begin{split}
\E\Big[X_{1,n}^21\big\{|X_{1,n}|>\e\sqrt{t_n}\big\}\Big]  
& \ \leq \ \sqrt{\E\big[X_{1,n}^4\big] \E\big[e^{ |X_{1,n}|}\big] \, e^{-\e \sqrt{t_n}}}.
\end{split}
\end{equation}
Using the general inequality $(a+b)^4\leq 8(a^4+b^4)$, we may bound the fourth moment as
\begin{equation}\label{eqn:preexp}
\begin{split}
\E[X_{1,n}^4] & \ \leq \  \, 8\sum_{l=1}^n p_l\big(\ts\frac{1}{np_l} a_l\ttop M a_l\big)^4 \ + \  8 \llangle M, G_n\rrangle^4\\[0.2cm]
& \ \leq \ \, 8\|M\|\op^4 \max_{1\leq l\leq n}\|\ts\frac{1}{\sqrt{np_l}} a_l\|_2^8 \ + \ c_0,
\end{split}
\end{equation}
where we have used the fact that $\llangle M,G_n\rrangle\leq c_0$ for some positive constant $c_0>0$.  Similarly, we have
\begin{equation}
\begin{split}
\E\big[e^{|X_{1,n}|}\big]e^{-\e\sqrt{t_n}}  & \ \leq \ \exp\Big({| \llangle[\tiny] M, G_n\rrangle[\small]|}-\e\sqrt{t_n}\Big) \tsum_{l=1}^n p_l\exp\big(\big|\ts\frac{1}{np_l} a_l\ttop M  a_l\!\big|\big)\\[0.2cm]
& \ \leq \ \exp\!\Big(c_0-\e\sqrt{t_n}+\|M\|\op\max_{1\leq l\leq n} \|\ts\frac{1}{\sqrt{np_l}} a_l\|_2^2\Big).
\end{split}
\end{equation}
Hence, under Assumption RS, there is a constant $c(\e)>0$ such that the bound
$$\E\big[e^{ |X_{1,n}|}\big]e^{-\e\sqrt{t_n}} \ \leq \ e^{c_0} e^{- c(\e) \sqrt{t_n}}$$
holds for all large $t$. Combining with~\eqref{eqn:preexp} and noting that the limit $\max_{1\leq l\leq n}\|\ts\frac{1}{\sqrt{np_l}}a_l\|_2^8 \,e^{-c(\e)\sqrt{t_n}}\to 0$ holds under Assumption RS, it follows that the Lindeberg condition~\eqref{eqn:lind1} indeed holds in the case of row~sampling.

Lastly, to handle the case of Gaussian random projections, it follows from the Cauchy-Schwarz and Chebyshev inequalities that
\begin{equation}\label{eqn:cs2}
\begin{split}
\E\Big[X_{1,n}^21\big\{|X_{1,n}|>\e\sqrt{t_n}\big\}\Big]  & \ \leq \ \sqrt{\E\big[X_{1,n}^4\big] \ts\frac{1}{\e^2 t_n}\var(X_{1,n})}.
\end{split}
\end{equation}
Next, it is known from~\citep[Lemma B.26]{Bai:Silverstein:2010} that if $s_{1,n}$ is a standard Gaussian vector in $\R^n$, then the following bound holds in terms of the matrix $K_n:=\ts\frac{1}{n}A_nMA_n\ttop$ and an absolute constant $c>0$,
\begin{equation}\label{eqn:4thmomentbound}
\begin{split}
\E[X_{1,n}^4] & \  \leq \ c\, \Big(\tr(K_nK_n\ttop)^2 \ + \ \tr\big((K_nK_n\ttop)^2\big)\Big)\\[0.2cm]
& \ = \ c\Big( \tr(M G_n M\ttop G_n) \ + \ \tr\big((MG_n M\ttop G_n)^2\big)\Big).
\end{split}
\end{equation}
In turn, since this bound converges to $\tr(M\mathsf{G}_{\infty}M\ttop \mathsf{G}_{\infty})+\tr\big((M\mathsf{G}_{\infty} M\ttop \mathsf{G}_{\infty})^2\big)$, we conclude that $\E[X_{1,n}^4]$ is bounded, and so~\eqref{eqn:cs2} implies that the Lindeberg condition~\eqref{eqn:lind1} holds in the case of Gaussian random~projections.\qed

~\\

\begin{lemma}\label{lem:corebootclt}
Suppose that the conditions of Theorem~\ref{thm:main} hold, and for each fixed $M\in\R^{d\times d}$, let $Z(M)$ be the Gaussian random variable in the statement of Lemma~\ref{lem:coreclt}. Then, the following limit holds as $n\to\infty$,

$$\mathcal{L}\Big(\llangle[\big] M, \sqrt{t_n}(\tilde{G}_n^* -\tilde{G}_n)\rrangle[\big]\, \Big|S_n\Big)\xrightarrow{ \ d \ } \mathcal{L}(Z(M)), \text{ \ \ \  in probability}.$$
\end{lemma}
\noindent \emph{Proof}. Recall that $s_{1,n},\dots,s_{t_n,n}\in\R^n$ denote the rows of $\sqrt{t_n}S_n$, and let $s_{1,n}^*,\dots,s_{t_n,n}^*$ be $t_n$ i.i.d.~samples drawn with replacement from $s_{1,n},\dots,s_{t_n,n}$.   By analogy with the proof of Lemma~\ref{lem:coreclt}, consider the algebraic~relation
\begin{equation}
 \llangle[\big] M, \sqrt{t_n}(\tilde{G}_n^* -\tilde{G}_n)\rrangle[\big] = \ts\frac{1}{\sqrt{t_n} \,n}\displaystyle\sum_{i=1}^{t_n} \Big(\llangle[\big] M, A_n\ttop (s_{i,n}^*)(s_{i,n}^*)\ttop  A_n \rrangle[\big] -\llangle[\big] M, \tilde A_n\ttop \tilde A_n\rrangle[\big]\Big),
\end{equation}
and define the random variable $X_{i,n}^*=\ts\frac{1}{n}\Big((s_{i,n}^*)\ttop A_n MA_n\ttop (s_{i,n}^*)-\llangle M,\tilde A_n\ttop \tilde A_n\rrangle\Big)$ for each $i\in\{1,\dots,t_n\}$ so that
\begin{equation}\label{eqn:bootlin}
 \llangle[\big] M, \sqrt{t_n}(\tilde{G}_n^* -\tilde{G}_n)\rrangle[\big] \ = \ \ts\frac{1}{\sqrt{t_n}}\displaystyle\sum_{i=1}^{t_n} X_{i,n}^*.
 \end{equation}
 Also observe that $\E[X_{i,n}^*|S_n]=0$ for every $i\in\{1,\dots,t_n\}$.
To complete the proof, it suffices to show that the conditions of the Lindeberg CLT for triangular arrays hold in a conditional sense (cf.~\cite[p.330-331]{vanderVaart:2000}). More precisely, it suffices to show that the conditional variance $\var( \llangle[\big] M, \sqrt{t_n}(\tilde{G}_n^* -\tilde{G}_n)\rrangle|S_n)$ converges to a limit in probability, and that the following limit holds for each fixed $\e>0$,
\begin{equation}\label{eqn:lindboot}
\E\Big[(X_{1,n}^*)^21\big\{|X_{1,n}^*|>\e\sqrt{t_n}\big\}\Big|\,S_n\Big] \ \to \ 0 \text{ \ \ \ \  in probability}.
\end{equation}

To show the latter condition in the cases of either the row-sampling or Gaussian random projections, let $L_n$ denote the left side of~\eqref{eqn:lindboot} and note that the definition of sampling with replacement implies
$$L_n=\ts\frac{1}{t_n}\sum_{i=1}^{t_n} X_{i,n}^21\{X_{i,n}>\e\sqrt{t_n}\}.$$
Consequently, for any $\e>0$, Markov's inequality gives
\begin{equation}
\begin{split}
\P(L_n>\e) & \ \leq \ \ts\frac{1}{\e}\E[L_n] \ = \ \ts\frac{1}{\e} \E\big[X_{1,n}^21\{X_{1,n}>\e\sqrt{t_n}\}\big],
\end{split}
\end{equation}
and so the condition~\eqref{eqn:lindboot} must hold because the limit $\E\big[X_{1,n}^21\{X_{1,n}>\e\sqrt{t_n}\}\big]\to 0$ was established in the proof of Lemma~\ref{lem:coreclt}.\\

Finally, we show that $\var( \llangle M, \sqrt{t_n}(\tilde{G}_n^* -\tilde{G}_n)\rrangle|S_n)$ has a limit in probability, and the argument will apply in the same manner to the cases of row sampling and Gaussian random projections. Define the random variable $\hat\varsigma_{t_n}^2:= \ts\frac{1}{t_n}\tsum_{i=1}^t X_{i,n}^2 - (\ts\frac{1}{t_n}\sum_{i=1}^{t_n} X_{i,n})^2$, and observe that
\begin{equation}
\begin{split}
\var\Big( \llangle[\big] M, \sqrt{t_n}(\tilde{G}_n^* -\tilde{G}_n)\rrangle\Big|S_n\Big) 
& \ = \ \var\big(X_{1,n}^*\big|S_n\big)\\[0.0cm]
& \ = \ \hat\varsigma^2_{t_n},
\end{split}
\end{equation}
which follows from the relation~\eqref{eqn:bootlin} and the fact that, conditionally on $S_n$, the random variable $X_{1,n}^*$ is a sample from the discrete uniform distribution on $\{X_{1,n},\dots,X_{t_n,n}\}$. 
Thus, it remains to show that $\hat\varsigma_{t_n}^2$ converges to a limit in probability. Due to basic facts about the sample variance $\hat\varsigma_{t_n}^2$, it is known that $\E[\hat\varsigma_{t_n}^2]=\frac{t_n-1}{t_n}\var(X_{1,n})$ 
and $\var(\hat \varsigma_{t_n}^2)=\mathcal{O}( \frac{1}{t_n}\E[X_{1,n}^4])$~\citep[p.~164]{Kenney}. Furthermore, the proof of Lemma~\ref{lem:coreclt} shows that $\frac{1}{t_n}\E[X_{1,n}^4]\to 0$ under either Assumption RP or RS, and so it follows that $\hat\varsigma_{t_n}^2$ must converge in probability to the same limit as $\var(X_{1,n})$, which completes the proof.\qed

\subsection{Lemmas for the right singular vectors}\label{sec:right}

In Lemmas~\ref{lem:projclt} and~\ref{lem:bootprojclt} we provide a joint CLT for the projection matrices $\sqrt{t_n}(\tilde P_j-P_j)$, as well as their bootstrap counterparts $\sqrt{t_n}(\tilde P_j^*-\tilde P_j)$ with $j=1,\dots,k$. (Recall that the definitions of definitions of $P_j$, $\tilde P_j$, and $\tilde P_j^*$ are given in~\eqref{eqn:Pdefs}.)

\begin{lemma}\label{lem:projclt}
Suppose that the conditions of Theorem~\ref{thm:main} hold. Then, for any fixed matrices $M_1,\dots,M_k\in\R^{d\times d}$,  there is a Gaussian vector $(Z_1(M_1),\dots,Z_k(M_k))$ in $\R^k$ such that 
\begin{equation}\label{eqn:projprod}
\mathcal{L}\Big(\sqrt{t_n}\llangle \tilde P_{1}-P_{1},M_1\rrangle,\dots, \sqrt{t_n}\llangle \tilde P_{k}-P_{k},M_k\rrangle \Big) \ \xrightarrow{ \ d \ } \ \mathcal{L}\big(Z_1(M_1),\dots,Z_k(M_k)\big). 
\end{equation}
Furthermore, there is a choice of the matrix $M_1\in\R^{d\times d}$ such that the Gaussian variable $Z_1(M_1)$ has positive variance.

\end{lemma}

\noindent \emph{Proof.} 
By the Cram\'er-Wold theorem, the limit~\eqref{eqn:projprod} can be established by showing that for any constants $c_1,\dots,c_k\in\R$, the sum $\sum_{j=1}^k  c_j\sqrt{t_n} \llangle \tilde P_j-P_j, M_j\rrangle$ converges in distribution to a Gaussian random variable. We will show this first, and then at the end of the proof, we will exhibit a choice of $M_1$ for which $\var(Z_1(M_1))>0$.

Recall that $\mathcal{S}^{d\times d}\subset\R^{d\times d}$ denotes the subspace of symmetric matrices, and for each $j\in\{1,\dots,k\}$, let $\psi_j:\mathcal{S}^{d\times d}\to \R$ denote the function that satisfies 
$\psi_j(G_n)=\llangle P_j,M_j\rrangle$
and $\psi_j(\tilde{G}_n)=\llangle \tilde P_j,M_j\rrangle$, so that
$$\sqrt{t}\llangle \tilde P_j-P_j,M_j\rrangle \ = \ \sqrt{t_n}(\psi_j(\tilde{G}_n)-\psi_j(G_n)).$$
To apply the mean-value theorem to the difference on the right, we may rely on the fact from matrix calculus that $\psi_j$ is continuously differentiable in an open neighborhood of any symmetric matrix whose $j$th eigenvalue is isolated~\cite[Theorem 8.9]{Magnus:2019}. Since all the eigenvalues of $\mathsf{G}_{\infty}$ are isolated, we may let $\mathcal{U}\subset\mathcal{S}^{d\times d}$ denote an open neighborhood on which all the functions $\psi_1,\dots,\psi_k$ are continuously differentiable.
Also, we need to define a random variable $\tilde R_{n,j}$ and a random matrix $\tilde D_{j,n}\in\mathcal{S}^{d\times d}$ in the following two cases: either (1) both of the matrices $G_n$ and $\tilde{G}_n$ lie in $\mathcal{U}$, or (2) at least one of the matrices $G_n$ or $\tilde{G}_n$ falls outside of $\mathcal{U}$. In the first case, let $\tilde R_{n,j}=0$, and let $\tilde D_{j,n}$ denote the differential (gradient) $\psi_j'(\breve G_{j,n})\in\mathcal{S}^{d\times d}$ evaluated at a random matrix $\breve G_{j,n}$ that is a convex combination of $G_n$ and $\tilde{G}_n$. In the second case, let $\tilde R_{n,j}=\sqrt{t_n}(\psi_j(\tilde{G}_n)-\psi_j(G_n))$ and let $\tilde D_{j,n}=0$.
Based on these definitions, the mean-value theorem ensures that the following relation always holds
\begin{equation}\label{eqn:mvt}
\sqrt{t_n}\Big(\psi_j(\tilde{G}_n)-\psi_j(G_n)\Big) \ = \ \llangle[\big] \tilde D_{j,n}, \sqrt{t_n}(\tilde{G}_n-G_n)\rrangle[\big] \ + \ \tilde R_{n,j}.
\end{equation}
Hence, if we let $\tilde D_n=\sum_{j=1}^k c_j \tilde D_{j,n}$ and $\tilde R_n=\sum_{j=1}^k  c_j\tilde R_{n,j}$, then
\begin{equation}
 \sum_{j=1}^k c_j \sqrt{t_n} \llangle \tilde P_j-P_j, M_j\rrangle \ = \ \llangle[\big] \tilde D_n, \sqrt{t_n}(\tilde{G}_n-G_n)\rrangle[\big]+\tilde R_n.
\end{equation}
Based on this relation, as well as Lemma~\ref{lem:coreclt} and Slutsky's lemma (Fact~\ref{lem:slutsky}), the proof of the limit~\eqref{eqn:projprod} reduces to establishing the following limits
\begin{equation}\label{eqn:Rnlim}
\tilde R_n \to 0 \text{ \ \ \ \ \ \ \ \ in probability},
\end{equation}
and 
\begin{equation}\label{eqn:Dnlim}
\tilde D_n\to \mathsf{D}_{\infty} \text{ \ \ \ \ \  in probability},
\end{equation}
for some constant matrix $\mathsf{D}_{\infty}\in\mathcal{S}^{d\times d}$. These limits are established below.

Let $\mathcal{E}_n$ denote the event that both $\tilde{G}_n$ and $G_n$ lie in the neighborhood $\mathcal{U}$. Given that $\tilde R_n$ can only be non-zero when $\mathcal{E}_n^c$ occurs, we have
$$\P(|\tilde R_n|>\e) \ \leq \ \P(\mathcal{E}_n^c).$$
Furthermore, since we assume that $G_n$ converges to $\mathsf{G}_{\infty}$, and since it is shown in Lemma~\ref{lem:Hlln} below that $\tilde{G}_n$ converges in probability to $\mathsf{G}_{\infty}$, it follows that $\P(\mathcal{E}_n^c)\to 0$. This establishes the limit~\eqref{eqn:Rnlim}. With regard to the limit~\eqref{eqn:Dnlim}, observe that our definitions give the relation
$$\tilde D_n = \sum_{j=1}^k c_j 1_{\mathcal{E}_n}\psi_j'(\breve G_{j,n}) ,$$
where $1_{\mathcal{E}_n}$ is the indicator of the event $\mathcal{E}_n$.
Based on the mentioned limits of $\tilde{G}_n$ and $G_n$, it follows that the convex combination $\breve G_{j,n}$ must converge in probability to $\mathsf{G}_{\infty}$. In addition, since the differential $\psi_j'$ is continuous on $\mathcal{U}$, it follows that $\psi_j'(\breve G_{j,n})$ converges in probability to the constant matrix $\psi_j'(\mathsf{G}_{\infty})$. Hence, if we put
\begin{equation}\label{eqn:Dinfty}
\mathsf{D}_{\infty}:= \sum_{j=1}^k c_j \psi_j'(\mathsf{G}_{\infty}),
\end{equation}
then the limit~\eqref{eqn:Dnlim} holds, and the proof of~\eqref{eqn:projprod} is complete.

Now, we turn to showing that $\var(Z_1(M_1))>0$. Let $\mathsf{v}_1$ and $\mathsf{v}_2$ denote the pair of eigenvectors of $\mathsf{G}_{\infty}$ mentioned in Assumption RS, and consider the particular choice of the matrix 
$$M_1=\mathsf{v}_2\mathsf{v}_1\ttop.$$
 Then, our previous argument leads to
$$\sqrt{t_n}\llangle[\big] \tilde P_{1}-P_{1}, M_1\rrangle[\big] \ = \ \llangle[\big] \psi_{1}'(\mathsf{G}_{\infty}), \sqrt{t_n}(\tilde{G}_n-G_n)\rrangle[\big] \ + \ o_{\P}(1),$$
where we recall that the definition of $\psi_{1}$ depends on $M_1$.
 Based on an analytical formula for the matrix differential of an eigenprojection~\cite[Theorem 8.9]{Magnus:2019}, the differential $\psi_{1}'(\mathsf{G}_{\infty})$ can be obtained explicitly. In particular, the inner product on the right side above may be calculated as follows, where $\lambda_1$ and $\lambda_2$ denote the eigenvalues of $\mathsf{G}_{\infty}$ corresponding to $\mathsf{v}_1$ and $\mathsf{v}_2$, and the symbol $\dagger$ refers to the Moore-Penrose inverse,
\begin{equation}\label{eqn:keydiff}
\begin{split}
\llangle[\big] \psi_{1}'(\mathsf{G}_{\infty}), \sqrt{t_n}(\tilde{G}_n-G_n)\rrangle[\big] &  \ = \ \llangle[\bigg]\mathsf{v}_1\mathsf{v}_1\ttop M_1\big(\lambda_1I_d-\mathsf{G}_{\infty}\big)^{\dagger}+\big(\lambda_1I_d-\mathsf{G}_{\infty}\big)^{\dagger}M_1 \mathsf{v}_1\mathsf{v}_1\ttop \ ,\ \sqrt{t_n}(\tilde{G}_n-G_n)\rrangle[\bigg],\\[0.2cm]
& \ =  \ \ts\frac{1}{\lambda_1-\lambda_2}\llangle[\big]M_1\ttop, \sqrt{t_n}(\tilde{G}_n-G_n)\rrangle[\big].
\end{split}
\end{equation}
where the second step is obtained by noting $\ttop \mathsf{v}_1\mathsf{v}_1\ttop M_1= 0$, as well as $(\lambda_1I_d-\mathsf{G}_{\infty})^{\dagger}\mathsf{v}_2=\frac{1}{\lambda_1-\lambda_2}\mathsf{v}_2$.  In turn, by rearranging the last expression in~\eqref{eqn:keydiff}, we have
\begin{equation}
\begin{split}
\llangle[\big] \psi_{1}'(\mathsf{G}_{\infty}), \sqrt{t_n}(\tilde{G}_n-G_n)\rrangle[\big] 
&  \ = \ \ts\frac{1}{\lambda_1-\lambda_2} \, \ts\frac{1}{\sqrt{t_n}}\displaystyle\sum_{i=1}^{t_n} \Big(s_{i,n}\ttop \big(\ts\frac{1}{n}A_n M_1\ttop A_n\ttop\big) s_{i,n} -\tr(\ts\frac{1}{n}A_n M_1\ttop A_n\ttop)\Big)
\end{split}
\end{equation}
and since the terms of the sum are mean-zero and i.i.d., we have
$$\var\Big(\llangle[\big] \psi_{1}'(\mathsf{G}_{\infty}), \sqrt{t_n}(\tilde{G}_n-G_n)\rrangle[\big]\Big) \ = \ \ts\frac{1}{(\lambda_1-\lambda_2)^2}\var\Big(s_{1,n}\ttop \big(\ts\frac{1}{n}A_n M_1\ttop A_n\ttop\big) s_{1,n} \Big).$$

 Altogether, it remains to separately check that the right side of this display has a positive limit in the cases of row sampling and Gaussian random projections. In the row sampling case, this follows directly from Assumption RS since $\var(s_{1,n}\ttop (\ts\frac{1}{n}A_n M_1\ttop A_n\ttop) s_{1,n})=\var(\tilde r_n\ttop (\mathsf{v}_1\mathsf{v}_2\ttop) \tilde r_n)\to \ell(\mathsf{v}_1\mathsf{v}_2\ttop)$. 
Alternatively, in the case when $S_n$ is a Gaussian random projection, we may use the formula~\eqref{eqn:gaussianqform}
 which leads to
\begin{equation}\label{eqn:gaussianvarcalc}
\begin{split}
 \var\Big(s_{1,n}\ttop \big(\ts\frac{1}{n}A_n M_1\ttop A_n\ttop\big) s_{1,n}\Big)  &\ = \ 2\Big\|\ts\frac{1}{n}A_n M_1\ttop A_n\ttop \Big\|_F^2\\[0.2cm]
  & \ = \ 2\tr\!\Big(M_1 G_n M_1\ttop G_n\Big)\\[0.2cm]
  & \ = \ 2\tr\!\Big(M_1\mathsf{G}_{\infty} M_1\ttop \mathsf{G}_{\infty}\Big)+o(1)\\[0.2cm]
  & \ = \ 2\Big(\mathsf{v}_1\ttop \mathsf{G}_{\infty}\mathsf{v}_1\Big) \Big(\mathsf{v}_2\ttop \mathsf{G}_{\infty} \mathsf{v}_2\Big)+o(1)\\[0.2cm]
  & \ = \ 2\lambda_1\lambda_2+o(1).
 \end{split}
\end{equation}
This clearly leads to a positive limit for the variance of $\llangle[\big] \psi_{1}'(\mathsf{G}_{\infty}), \sqrt{t_n}(\tilde{G}_n-G_n)\rrangle[\big]$, which completes the~proof.\qed

~\\

\begin{lemma}\label{lem:bootprojclt}
Suppose the conditions of Theorem~\ref{thm:main} hold, and for any fixed matrices $M_1,\dots,M_k\in\R^{d\times d}$, let $(Z_1(M_1),\dots,Z_k(M_k))$ be the Gaussian vector in the statement of Lemma~\ref{lem:projclt}. Then, as $n\to\infty$, 
\begin{equation}\label{eqn:projprodboot}
\mathcal{L}\Big(\sqrt{t_n}\llangle \tilde P_{1}^*-\tilde P_{1},M_1\rrangle,\dots, \sqrt{t_n}\llangle \tilde P_{k}^*-\tilde P_{k},M_k\rrangle\,\Big|S_n \Big) \ \xrightarrow{ \ d \ } \ \mathcal{L}\big(Z_1(M_1),\dots,Z_k(M_k)\big) \ \ \text{ in probability}.
\end{equation}
\end{lemma}

\noindent\emph{Proof}. The argument is similar to the proof of Lemma~\ref{lem:projclt}, but with some differences that we explain here. By the Cram\'er-Wold theorem, it suffices to show that the following limit holds for any fixed numbers $c_1,\dots,c_k\in\R$,
$$\mathcal{L}\Big(\tsum_{j=1}^k  c_j\sqrt{t_n} \llangle \tilde P_j^*-\tilde P_j, M_j\rrangle\Big|S_n\Big) \ \xrightarrow{ \ d \ } \ \mathcal{L}\Big(\tsum_{j=1}^k c_j Z_j(M_j)\Big) \text{ \ \ \ \ in probability}.$$ 
To show this, we will combine Lemmas,~\ref{lem:condslutsky},~\ref{lem:coreclt}, and~\ref{lem:projclt}.

Let the fixed matrix $\mathsf{D}_{\infty}\in\R^{d\times d}$ denote the limit in~\eqref{eqn:Dinfty} that depends on $c_1,\dots,c_k$, and observe that the proof of Lemma~\ref{lem:projclt} shows that
$$\mathcal{L}\Big(\llangle[\big] \mathsf{D}_{\infty}, \sqrt{t_n}(\tilde{G}_n-G_n)\rrangle[\big]\Big) \xrightarrow{ \ d \ } \mathcal{L}\Big(\tsum_{j=1}^k c_j Z_j(M_j)\Big).$$
Next, we claim that the following expansion holds 
\begin{equation}\label{eqn:bootexpansion}
\tsum_{j=1}^k  c_j\sqrt{t_n} \llangle \tilde P_j^*-\tilde P_j, M_j\rrangle \ = \ \llangle[\big] \tilde D_n^*, \sqrt{t_n}(\tilde{G}_n^*-\tilde{G}_n)\rrangle[\big]+\tilde R_n^*
\end{equation}
where $\tilde D_n^*\in\R^{d\times d}$ is a random matrix and $\tilde R_n^*$ is a random scalar that satisfy the following limits for any fixed $\e>0$,
\begin{equation}\label{eqn:limDstar}
\P\Big(\|\tilde D_n^*-\mathsf{D}_{\infty}\|_F>\e\,\Big|\, S_n\Big) \ \to \ 0 \text{ \ \ \ in probability},
\end{equation}
\begin{equation}\label{eqn:limRstar}
\ \ \ \ \ \ \ \P\Big(|R_n^*-0|>\e\,\Big| \, S_n\Big) \ \to \ 0 \text{ \ \ \ in probability}.
\end{equation}
As a consequence of this claim, and the fact that $\mathcal{L}(\llangle \mathsf{D}_{\infty}, \sqrt{t_n}(\tilde{G}_n^*-\tilde{G}_n)\rrangle| S_n)$ conditionally converges in distribution to the same limit as $\mathcal{L}(\llangle \mathsf{D}_{\infty}, \sqrt{t_n}(\tilde{G}_n-G_n)\rrangle )$ (by Lemma~\ref{lem:coreclt}), the proof will be completed by the conditional version of Slutsky's lemma (Lemma~\ref{lem:condslutsky}). Thus, it remains to verify the three parts~\eqref{eqn:bootexpansion},~\eqref{eqn:limDstar}, and~\eqref{eqn:limRstar} of the claim.

For each $j\in\{1,\dots,k\}$, let $\psi_j:\mathcal{S}^{d\times d}\to \R$ be as defined in the proof of Lemma~\ref{lem:projclt}, and let $\mathcal{U}\subset\mathcal{S}^{d\times d}$ again denote the open neighborhood of $\mathsf{G}_{\infty}$ on which all the functions $\psi_1,\dots,\psi_k$ are continuously differentiable. Also, let $\mathcal{E}_n'$ denote the event that $\tilde{G}_n^*$ lies in $\mathcal{U}$, and recall that $\mathcal{E}_n$ denotes the event that both $\tilde{G}_n$ and $G_n$ lie in $\mathcal{U}$. To establish the expansion~\eqref{eqn:bootexpansion}, we now define $\tilde D_n^*$ and $\tilde R_n^*$ as follows. When $\mathcal{E}_n'\cap \mathcal{E}_n$ holds, the mean-value theorem ensures that for each $j\in\{1,\dots,k\}$ we have
\begin{equation}
\begin{split}
 \sqrt{t_n}\llangle \tilde P_j^*-\tilde P_j, M_j\rrangle  
 & \ = \ \sqrt{t_n}\big(\psi_j(\tilde{G}_n^*)-\psi_j(\tilde{G}_n)\big)\\[0.2cm]
 & \ = \ \llangle \tilde D_{j,n}^*, \sqrt{t_n}(\tilde{G}_n^*-\tilde{G}_n)\rrangle,
 \end{split}
\end{equation}
where $\tilde D_{j,n}^*$ is a shorthand for the differential $\psi_j'(\breve G_{j,n}^*)\in\mathcal{S}^{d\times d}$ evaluated at a point $\breve G_{j,n}^*$ that is a convex combination of $\tilde G_{j,n}^*$ and $\tilde G_{j,n}$.   Accordingly, when $\mathcal{E}_n'\cap\mathcal{E}_n$ holds, we define $\tilde D_n^*=\sum_{j=1}^k c_j\tilde{D}_{j,n}^*$ and $\tilde R_n^*=0$. Oppositely, when the event $(\mathcal{E}_n'\cap\mathcal{E}_n)^c$ holds, we put $\tilde D_n^*=0$ and $\tilde R_n^*=\sum_{j=1}^k c_j\sqrt{t_n}\big(\psi_j(\tilde{G}_n^*)-\psi_j(\tilde{G}_n)\big)$. Based on these definitions, it follows that the expansion~\eqref{eqn:bootexpansion} always holds.

Turning to the limit~\eqref{eqn:limRstar}, the event $\{|\tilde R_n^*-0|>\e\}$ can only occur when $(\mathcal{E}_n'\cap\mathcal{E}_n)^c$ occurs. Consequently, a union bound gives
\begin{equation}\label{eqn:unionmarkov}
\begin{split}
\E\Big[\P\Big(|R_n^*-0|>\e\,\Big| \, S_n\Big)\Big] \ \leq  \ \P((\mathcal{E}_n')^c)+\P(\mathcal{E}_n^c).
\end{split}
\end{equation}
Furthermore, we know from the proof of Lemma~\ref{lem:projclt} that $\P(\mathcal{E}_n^c)\to 0$, and also, it is straightforward to check that $\P((\mathcal{E}_n')^c)\to 0$. Thus,~\eqref{eqn:unionmarkov}~implies~{\eqref{eqn:limRstar} via Markov's inequality.

Lastly, handling the limit~\eqref{eqn:limDstar} can be reduced to showing $\tilde{G}_n^*\xrightarrow{\P}\mathsf{G}_{\infty}$  (unconditionally), which is done in Lemma~\ref{lem:Hlln} below. This is sufficient because the limit $\tilde{G}_n\xrightarrow{\P} \mathsf{G}_{\infty}$ (cf. Lemma~\ref{lem:Hlln}) and the continuity of $\psi_j'(\cdot)$ on the neighborhood $\mathcal{U}$ imply $\psi_j'(\breve G_{j,n}^*)\xrightarrow{\P}\psi_j'(\mathsf{G}_{\infty})$ for all $j\in\{1,\dots,k\}$, which leads to $\P(\|\tilde D_{n}^*-\mathsf{D}_{\infty}\|_F>\e )\to 0$. In turn, this implies~\eqref{eqn:limDstar} via Markov's inequality, and the proof is complete.\qed
~\\

\begin{lemma}\label{lem:Hlln}
Suppose that the conditions of Theorem~\ref{thm:main} hold. Then, the following limits hold as $n\to\infty$,
\begin{equation}\label{eqn:firstlln}
\tilde{G}_n\xrightarrow{\P}\mathsf{G}_{\infty}
\end{equation}
and
\begin{equation}\label{eqn:secondlln}
\tilde{G}_n^*\xrightarrow{\P}\mathsf{G}_{\infty}.
\end{equation}
\end{lemma}
\proof The first limit~\eqref{eqn:firstlln} is a direct consequence of Lemma~\ref{lem:coreclt}.
To handle the second limit~\eqref{eqn:secondlln}, observe that Chebyshev's inequality (conditional on $S_n$) gives
\begin{equation}
\begin{split}
\P\big(\|\tilde{G}_n^*-\tilde{G}_n\|_F>\e\big|S_n\big) &\ \leq \ \ts\frac{1}{\e^2}\E\big[\big\|\tilde{G}_n^*-\tilde{G}_n\|_F^2\big|S_n\big]\\[0.2cm]
& \ = \ \ts \frac{1}{\e^2}\E\Big[\big\| \ts\frac{1}{t_n} \tsum_{i=1}^t  \ts\frac{1}{n}A_n\ttop (s_{i,n}^*)(s_{i,n}^*)\ttop A_n  -\ts\frac{1}{n}A_n\ttop S_n\ttop S_nA_n\big\|_F^2\Big|S_n\Big]\\[0.2cm]
& \ = \  \ts \frac{1}{\e^2t_n}\E\Big[\big\|  A_n\ttop (s_{1,n}^*)(s_{1,n}^*)\ttop A_n  - \ts\frac{1}{n}A_n\ttop S_n\ttop  S_nA_n\big\|_F^2\Big|S_n\Big]\\[0.2cm]
& \ = \ \ts\frac{1}{\e^2t_n}\ts\frac{1}{t_n}\sum_{i=1}^{t_n}\big\| \ts\frac{1}{n}A_n\ttop s_{i,n}s_{i,n}\ttop A_n  - \ts\frac{1}{n}A_n\ttop S_n\ttop S_nA_n\big\|_F^2,
\end{split}
\end{equation}
where the third line follows from the general fact that if $Y_1,\dots,Y_t$ are independent~mean-zero random matrices, then $\E[\|Y_1+\dots+Y_{t_n}\|_F^2] = \sum_{i=1}^{t_n} \E[\|Y_i\|_F^2]$, and the fourth line follows from the definition of sampling with replacement. So, taking an expectation over $S_n$ on both sides of the previous display, we have
\begin{equation}
\begin{split}
\P\big(\|\tilde{G}_n^*-\tilde{G}_n\|_F>\e\big) & \ \leq \ \ts\frac{1}{\e^2t_n}\E\Big[\big\|  \ts\frac{1}{n}A_n\ttop s_{1,n}s_{1,n}\ttop A_n  - \ts\frac{1}{n}A_n\ttop S_n\ttop  S_nA_n\big\|_F^2\Big]\\[0.2cm]
& \ \leq  \ \ts\frac{2}{\e^2t_n}\E\Big[\big\|\ts\frac{1}{n}A_n\ttop s_{1,n}s_{1,n}\ttop A_n-G_n\big\|_F^2\Big]+ \ts\frac{2}{\e^2t_n}\E\Big[\big\|G_n-\ts\frac{1}{n}A_n\ttop S_n\ttop S_nA_n\big\|_F^2\Big]\\[0.2cm]
& \ = \ \big(\ts\frac{2}{\e^2 t_n}+\ts\frac{2}{\e^2 t^2_n}\big)\Big(\E\big[(\ts\frac{1}{n}s_{1,n}\ttop A_nA_n\ttop s_{1,n})^2\big]-\|G_n\|_F^2\Big)\\[0.2cm]
& \ = \ \big(\ts\frac{2}{\e^2 t_n}+\ts\frac{2}{\e^2 t^2_n}\big)\Big(\var\!\big(\ts\frac{1}{n}s_{1,n}\ttop A_nA_n\ttop s_{1,n}\big)+\big(\tr(G_n)^2-\|G_n\|_F^2\big)\Big),
\end{split}
\end{equation}
where the third line relies on expanding $A_n S_n\ttop S_n A_n$ as a sum and using the identity $\E[\|Y_1+\dots+Y_{t_n}\|_F^2] = \sum_{i=1}^{t_n} \E[\|Y_i\|_F^2]$ that was mentioned just a moment ago.
Finally, the proof of Lemma~\ref{lem:coreclt} shows that the quantity $\var\!(\ts\frac{1}{n}s_{1,n}\ttop A_nA_n\ttop s_{1,n})+(\tr(G_n)^2-\|G_n\|_F^2)$ converges to a finite limit (under either Assumption RP or RS). Hence, the $\mathcal{O}(1/t_n)$ prefactor requires $\P\big(\|\tilde{G}_n^*-\tilde{G}_n\|_F>\e\big)$ to converge to 0, as needed. \qed
~\\

%
%

\subsection{Lemma for the left singular vectors}\label{sec:left}
The following lemma is needed for proving the limit~\eqref{eqn:limU} in Theorem~\ref{thm:main}.
\begin{lemma}\label{lem:Uclt}
Suppose that the conditions of Theorem~\ref{thm:main} hold, and for each $j\in\{1,\dots,k\}$, let $\tilde \Delta_j$ and $\tilde \Delta_j^*$ be as defined in~\eqref{eqn:deltajdef} and~\eqref{eqn:deltajstardef}. Then, for any fixed matrices $M_1,\dots,M_k\in\R^{d\times d}$, there is an associated Gaussian random vector $(\zeta_1(M_1),\dots,\zeta_k(M_k))\in\R^k$ such that
 \begin{align}
 \footnotesize
 \mathcal{L}\Big(\llangle[\big] \sqrt{t_n} \tilde\Delta_1,M_1\rrangle[\big],\dots,\llangle[\big] \sqrt{t_n}\tilde\Delta_k,M_k\rrangle[\big]\Big) & \ \xrightarrow{ \ d \ } \ \mathcal{L}(\zeta_1(M_1),\dots,\zeta_k(M_k)), \text{ \ \ and }\label{eqn:limdelta2}\\[0.2cm]
 %
   \mathcal{L}\Big(\sqrt{t_n}\tilde\Delta_1^*,M_1\rrangle,\dots,\llangle \sqrt{t_n}\tilde\Delta_k^*,M_k\rrangle\Big| S_n\Big) & \ \xrightarrow{ \ d \ } \ \mathcal{L}(\zeta_1(M_1),\dots,\zeta_k(M_k)) \text{ \ \ \  in probability}.\label{eqn:limdeltastar2}
  \end{align}
  Furthermore, there is a choice of the matrix $M_1\in\R^{d\times d}$ such that the Gaussian variable $\zeta_1(M_1)$ has positive~variance.
\end{lemma}
\noindent\emph{Proof}. It is straightforward to check that the algebraic relation
\begin{equation}\label{eqn:Kreln}
\llangle \sqrt{t_n}\tilde \Delta_j,M_j\rrangle \ = \ \llangle \sqrt{t_n}(\tilde P_j-P_j),\tilde K_j\rrangle
\end{equation}
holds for every $j\in\{1,\dots,k\}$, where we let
$$\tilde K_j=\ts\frac{G_n^{1/2}M_jG_n^{1/2}}{\tr(\tilde P_j G_n)}-\ts\frac{\llangle P_j,G_n^{1/2}M_jG_n^{1/2}\rrangle }{\tr(\tilde P_jG_n)\tr(P_j G_n)}\, G_n.$$
In addition, since $\mathsf{G}_{\infty}$ has isolated eigenvalues, the function that maps a symmetric matrix to its $j$th eigenprojection is continuous in an open neighborhood of $\mathsf{G}_{\infty}$~\cite[Theorem 8.9]{Magnus:2019}. Next, if we let $(\lambda_1,\mathsf{v}_1),\dots,(\lambda_d,\mathsf{v}_d)$ denote the eigenvalue-eigenvectors pairs of $\mathsf{G}_{\infty}$ with corresponding eigenprojections $\mathsf{P}_j:=\mathsf{v}_j\mathsf{v}_j\ttop$,  then the limits
$$P_j=\mathsf{P}_j+o(1) \ \ \ \ \text{ and } \ \ \ \  \tilde P_j=\mathsf{P}_{j}+o_{\P}(1),$$
follow from $G_n=\mathsf{G}_{\infty}+o(1)$ and $\tilde{G}_n=\mathsf{G}_{\infty}+o_{\P}(1)$ (by Lemma~\ref{lem:Hlln}). (Also recall that the dependence of $P_j$ and $\tilde P_j$ on $n$ is suppressed.)
 In turn, we have
$$\tilde K_j =\ts\frac{1}{\lambda_j}\mathsf{G}_{\infty}^{1/2}M_{j}\mathsf{G}_{\infty}^{1/2}-\ts\frac{1}{\lambda_j}\llangle \mathsf{P}_j,M_j\rrangle \mathsf{G}_{\infty}+o_{\P}(1).$$
So, if we combine this expression for $\tilde K_j$ with~\eqref{eqn:Kreln},  Lemma~\ref{lem:projclt}, and Slutsky's lemma, it follows that the limit~\eqref{eqn:limdelta2} holds.\\

To show there is a choice of $M_1$ for which $\llangle \sqrt{t_n}\tilde \Delta_1, M_1\rrangle$ converges to a distribution with a positive variance, consider the choice $M_1=\mathsf{v}_2\mathsf{v}_1\ttop$. In this case, it can be checked that
$$\tilde K_{1} =\ts\frac{\sqrt{\lambda_2}}{\sqrt{\lambda_1}} M_1+o_{\P}(1),$$
and in the proof of Lemma~\ref{lem:projclt} it is shown that the limiting distribution of $\llangle \sqrt{t_n}(\tilde P_j-P_j),M_1\rrangle$ has positive variance (under either Assumption RS or RP). Thus, by Slutsky's lemma, the random variable $\llangle \sqrt{t_n}\tilde \Delta_1,M_1\rrangle$ must also have a limiting distribution with positive variance.\\

Lastly, it remains to establish the limit~\eqref{eqn:limdeltastar2}. For each $j\in\{1,\dots,k\}$, define the random matrix
\begin{equation}
\tilde K_j^* = \ts\frac{\tilde{G}_n^{1/2}M_j\tilde{G}_n^{1/2}}{\tr(\tilde P_j^* \tilde{G}_n)}-\ts\frac{\llangle \tilde P_j,\tilde{G}_n^{1/2}M_j\tilde{G}_n^{1/2}\rrangle }{\tr(\tilde P_j^*\tilde{G}_n)\tr(\tilde P_j\tilde{G}_n)}\, \tilde{G}_n,
\end{equation}
which leads to the algebraic relation
\begin{equation}\label{eqn:deltastarreln}
 \llangle \sqrt{t_n}\tilde \Delta_j^*,M_j\rrangle \ = \ \llangle \sqrt{t_n}(\tilde P_j^*-\tilde P_j),\tilde K_j^*\rrangle.
\end{equation}
In addition, using the reasoning that led to the limit $\tilde P_j=\mathsf{P}_j+o_{\P}(1)$ and the fact that $\tilde{G}_n^*=\mathsf{G}_{\infty}+o_{\P}(1)$ (by Lemma~\ref{lem:Hlln}), it can be checked that $\tilde P_j^*=\mathsf{P}_j+o_{\P}(1)$, which leads to
\begin{equation}
\tilde K_j^* \ = \ \ts\frac{1}{\lambda_j}\mathsf{G}_{\infty}^{1/2}M_{j}\mathsf{G}_{\infty}^{1/2}-\ts\frac{1}{\lambda_j}\llangle \mathsf{P}_j,M_j\rrangle \mathsf{G}_{\infty}+o_{\P}(1).
\end{equation}
Finally, by combining this with the relation~\eqref{eqn:deltastarreln}, Lemma~\ref{lem:bootprojclt}, and the conditional of version of Slutsky's lemma (Lemma~\ref{lem:condslutsky}), it follows that the limit~\eqref{eqn:limdeltastar2} holds.\qed


\subsection{Lemma for the singular values}\label{sec:svals}
The following lemma gives a joint CLT for $\sqrt{t_n}(\sigma_j(\tilde A_n)-\sigma_j(A_n))$ with $j=1,\dots,k$, as well as for the bootstrap counterparts $\sqrt{t_n}(\sigma_j(\tilde A_n^*)-\sigma_j(\tilde A_n))$.

\begin{lemma}\label{lem:svals}
Suppose that the conditions of Theorem~\ref{thm:main} hold. Then, for any fixed real numbers $c_1,\dots,c_k$, there is a Gaussian random variable $\zeta(c_1,\dots,c_k)$  such that as $n\to\infty$,
\begin{equation}\label{eqn:svalclt}
\mathcal{L}\Big(\tsum_{j=1}^k\ts\frac{\sqrt{t_n}}{\sqrt n} c_j(\sigma_j(\tilde A_n)-\sigma_j(A_n))\Big) \ \xrightarrow{ \ d \ } \ \mathcal{L}(\zeta(c_1,\dots,c_k)),
\end{equation}
and
\begin{equation}\label{eqn:svalcltboot}
\mathcal{L}\Big(\tsum_{j=1}^k\ts\frac{\sqrt{t_n}}{\sqrt n} c_j(\sigma_j(\tilde A_n^*)-\sigma_j(\tilde A_n)) \, \Big|\,S_n\Big) \ \xrightarrow{ \ d \ } \ \mathcal{L}(\zeta(c_1,\dots,c_k)) \text{ \ \  in probability}.
\end{equation}
Lastly, the random variable  $\zeta(1,0,\dots,0)$ has positive variance.
\end{lemma}

\noindent \emph{Proof}. First, we prove the limit~\eqref{eqn:svalclt}. For any fixed positive semidefinite matrix $M\in\mathcal{S}^{d\times d}$ and index $j\in\{1,\dots,k\}$, let $\varphi_j(M)=\sqrt{\lambda_j(M)}$ so that we have the relation
$$\ts\frac{\sqrt{t_n}}{\sqrt n}(\sigma_j(\tilde A_n)-\sigma_j(A_n))  \ = \ \sqrt{t_n}\big(\varphi_j(\tilde{G}_n)-\varphi_j(G_n)\big).$$
Since the limiting matrix $\mathsf{G}_{\infty}$ has isolated eigenvalues, it is a fact from matrix calculus that there is an open neighborhood $\mathcal{V}\subset\mathcal{S}^{d\times d}$ of $\mathsf{G}_{\infty}$ such that all of the functions $\varphi_1,\dots,\varphi_k$ are continuously differentiable on $\mathcal{V}$~\cite[Theorem 8.9]{Magnus:2019}. Consequently, if we let $\varphi_j'(\mathsf{G}_{\infty})\in\mathcal{S}^{d\times d}$ denote the differential of $\varphi_j$ at $\mathsf{G}_{\infty}$, and define the matrix $\mathsf{J}_{\infty}= \tsum_{j=1}^k c_j \,\varphi_j'(\mathsf{G}_{\infty}) $, then the argument in the proof of Lemma~\ref{lem:projclt} can be re-used to show that the following expansion holds
$$\tsum_{j=1}^k \ts\frac{\sqrt{t_n}}{\sqrt n}c_j (\sigma_j(\tilde A_n)-\sigma_j(A_n))
\ = \ \llangle[\big] \mathsf{J}_{\infty} , \, \sqrt{t_n}(\tilde{G}_n-G_n)\rrangle[\big]+o_{\P}(1).$$
 In turn, it follows directly from Lemma~\ref{lem:coreclt} that the limit~\eqref{eqn:svalclt} holds.

Next, to handle the limit~\eqref{eqn:svalcltboot}, an argument that is analogous to the proof of Lemma~\ref{lem:bootprojclt} can be used. In particular, the same reasoning can be used to show that there is a random matrix $\tilde J_n^*\in\mathcal{S}^{d\times d}$ and a remainder variable $\tilde W_n^*$, such that the following equation holds
\begin{equation}
\tsum_{j=1}^k \ts\frac{\sqrt{t_n}}{\sqrt n}c_j (\sigma_j(\tilde A_n^*)-\sigma_j(\tilde A_n)) 
\ =  \ \llangle[\big] \tilde J_n^*, \sqrt{t_n}(\tilde{G}_n^*-\tilde{G}_n)\rrangle[\big]+\tilde W_n^*.
\end{equation}
In addition, for any $\e>0$, the limits  $\P\big(\|\tilde J_n^*-\mathsf{J}_{\infty}\|_F>\e\big| S_n\big) =o_{\P}(1)$ and $\P(|\tilde W_n^*-0|>\e|S_n)=o_{\P}(1)$ can be shown to hold as well. Thus, the conditional version of Slutsky's lemma (Lemma~\ref{lem:condslutsky}), combined with Lemma~\ref{lem:bootprojclt}, lead to the the desired limit~\eqref{eqn:svalcltboot}.

Lastly, to prove that $\zeta(1,0,\dots,0)$ has positive variance, it suffices to show that the random variable $\llangle \varphi_1'(\mathsf{G}_{\infty}),\sqrt{t_n}(\tilde{G}_n-G_n)\rrangle$ converges to a Gaussian random variable with positive variance. Using an analytical formula for the differential $\varphi_1'(\mathsf{G}_{\infty})$ available in~\cite[Theorem 8.9]{Magnus:2019}, we~have
\begin{equation}
\begin{split}
 \llangle[\big] \varphi_1'(\mathsf{G}_{\infty}),\sqrt{t_n}(\tilde{G}_n-G_n)\rrangle[\big] & \ = \ \llangle[\big] \ts\frac{1}{2\sqrt{\lambda_1}} \mathsf{v}_1\mathsf{v}_1\ttop , \sqrt{t_n}(\tilde{G}_n-G_n)\rrangle[\big]\\[0.1cm]
 & \ = \ \ts\frac{1}{2\sqrt{\lambda_1}}\ts\frac{1}{\sqrt{t_n}}\displaystyle\sum_{i=1}^{t_n} \Big(s_{i,n}\ttop \big(\ts\frac{1}{n}A_n \mathsf{v}_1\mathsf{v}_1\ttop A_n\ttop\big) s_{i,n} -\tr(\ts\frac{1}{n}A_n \mathsf{v}_1\mathsf{v}_1\ttop A_n\ttop)\Big).
 \end{split}
\end{equation}
Since the last sum consists of i.i.d.~zero-mean random variables, the variance of the limiting Gaussian distribution will be positive if the sequence  $\var(s_{1,n}\ttop \big(\ts\frac{1}{n}A_n \mathsf{v}_1\mathsf{v}_1\ttop A_n\ttop\big) s_{1,n})$ has a positive limit. In the case of a row-sampling sketching matrix, the positive limit follows from Assumption RS, and in the case of a Gaussian random projection, this can be verified by essentially repeating the calculation~\eqref{eqn:gaussianvarcalc}.\qed

\subsection{Examples satisfying Assumptions RP and RS}\label{sec:examples}

\noindent \textbf{Example 1.}  For any positive definite matrix $\mathsf{G}_{\circ}\in\R^{d\times d}$, we may define an associated sequence of vectors $a_1,a_2,\dots$ in $\R^d$ as follows. If $1\leq l\leq d$, define $a_l$ to be the $l$th row of the matrix $\sqrt{d}\mathsf{G}_{\circ}^{1/2}$, and if $l>d$, define the successive vectors in a cyclical manner, $a_{d+1}=a_1,a_{d+2}=a_2,\dots,a_{2d}=a_d, a_{2d+1}=a_1$, and so on. In this notation, let the rows of $A_n\in\R^{n\times d}$ consist of the first $n$ such vectors.  When $n$ is an exact multiple of $d$, we have
$$\ts\frac{1}{n}A_n\ttop A_n = \ts\frac{1}{d}\sum_{l=1}^d a_la_l\ttop = \, \mathsf{G}_{\circ}.$$
More generally, as $n\to\infty$ with $d$ held fixed, it is simple to check that $\ts\frac{1}{n}A_n\ttop A_n\to \mathsf{G}_{\circ}$, and so the matrix $\mathsf{G}_{\circ}$ plays the role of $\mathsf{G}_{\infty}$ in this example. Likewise, any choice of $\mathsf{G}_{\circ}$ whose eigenvalues $\lambda_1(\mathsf{G}_{\circ}),\dots,\lambda_d(\mathsf{G}_{\circ})$ all have multiplicity 1 will ensure that Assumption RP holds.\\

Next, we consider Assumption RS. It is straightforward to check that for any fixed $n$ and fixed $C\in\R^{d\times d}$ we have
\begin{equation}\label{eqn:basicvar}
\var(\tilde r_n\ttop C \tilde r_n) = \tsum_{l=1}^n p_l \big(\ts\frac{1}{np_l} a_l\ttop Ca_l\big)^2 - \Big(\ts\frac{1}{n}\sum_{l=1}^n a_l\ttop Ca_l\Big)^2.
\end{equation}
If we focus on the case of uniform sampling with $p_l=1/n$ for all $l\in\{1,\dots,n\}$, then we have the following limit as $n\to\infty$,
$$\var(\tilde r_n\ttop C\tilde r_n) \ \to \ \ell(C)= \ts\frac{1}{d}\sum_{l=1}^d (a_l\ttop Ca_l)^2- \Big(\ts\frac{1}{d}\sum_{l=1}^d a_l\ttop Ca_l\Big)^2.$$
To consider the choices $C=\mathsf{v}_1\mathsf{v}_1\ttop$ or $C=\mathsf{v}_1\mathsf{v}_2\ttop$ in the case of uniform sampling, we may use the algebraic~identity 
\begin{equation}\label{eqn:usefulident}
a_l\ttop \mathsf{v}_1 =\sqrt d e_l\ttop \mathsf{G}_{\circ}^{1/2} \mathsf{v}_1 = \sqrt{d\lambda_1(\mathsf{G}_{\circ})} e_l\ttop \mathsf{v}_1
\end{equation}
to evaluate the limit function $\ell(\cdot)$ as
$$\ell(\mathsf{v}_1\mathsf{v}_1\ttop)=\lambda_1^2(\mathsf{G}_{\circ}) \Big(d\tsum_{l=1}^d (e_l\ttop \mathsf{v}_1)^4-1\Big)
 \text{ \ \ \ and \ \ \ } 
\ell(\mathsf{v}_1\mathsf{v}_2\ttop) = \lambda_1(\mathsf{G}_{\circ})\lambda_2(\mathsf{G}_{\circ})d\tsum_{l=1}^d (e_l\ttop \mathsf{v}_1)^2(e_l\ttop \mathsf{v}_2)^2.$$
Based on these formulas, it follows that we have $\ell(\mathsf{v}_1\mathsf{v}_1\ttop)>0$ and $\ell(\mathsf{v}_1\mathsf{v}_2\ttop)>0$ under two rather generic conditions: First, we have $\ell(\mathsf{v}_1\mathsf{v}_1\ttop)>0$ as long as $\mathsf{v}_1$ is not parallel to a vector of the form $(\ts\frac{\pm 1}{\sqrt d},\dots,\ts\frac{\pm1}{\sqrt d})$, which can be checked by noting that the Cauchy-Schwarz inequality 
$$1= \tsum_{l=1}^d 1 \cdot(e_l\ttop \mathsf{v}_1)^2 \leq \sqrt{d}\sqrt{\tsum_{l=1}^d (e_l\ttop \mathsf{v}_1)^4}$$ 
holds with equality precisely when $\mathsf{v}_1$ is parallel to a vector of the stated form. Second, we have \smash{$\ell(\mathsf{v}_1\mathsf{v}_2\ttop)>0$} as long as there is at least one coordinate in $\{1,\dots,d\}$ where $\mathsf{v}_1$ and $\mathsf{v}_2$ are both non-zero. Lastly, since the set of values $\{\|a_1\|_2,\dots,\|a_d\|_2\}$ is fixed with respect to $n$, it follows that in the case of uniform sampling, the growth condition $\max_{1\leq l\leq n}\ts\frac{1}{\sqrt{n p_l}}\|a_l\|_2=o(t_n^{1/8})$ is satisfied as well.

\paragraph{Example 2.} Although Theorem~\ref{thm:main} is based on a framework in which the matrix $A_n\in\R^{n\times d}$ is deterministic, it is of interest to know if Assumptions RP or RS are likely to hold for particular realizations of $A_n$ generated at random by ``nature''. Likewise, for the purposes of this example only, the matrix $A_n$ will be treated as being generated independently of all sources of algorithmic randomness. Furthermore, since the input matrix to an SVD algorithm is often viewed as having rows that represent data points in $\R^d$, we will consider the case where the rows of $A_n$ are i.i.d.~samples from a centered elliptical distribution~(cf.~\cite{Cambanis:1981}).\footnote{Distributions of this type are commonly used in multivariate data analysis.}
In detail, this means that each vector $a_i$ can be expressed in the form $a_i=\sqrt d \nu_i\mathsf{G}_{\circ}^{1/2}U_i$, where $\mathsf{G}_{\circ}\in\R^{d\times d}$ is a fixed positive definite matrix with isolated eigenvalues, and the pairs $(\nu_1,U_1),(\nu_2,U_2),\dots$ are i.i.d.~elements in $\R\times \R^d$. In addition, each $U_i$ is uniformly distributed on the unit $\ell_2$-sphere, and each $\nu_i$ is a non-negative random variable (independent of $U_i$) with a finite moment generating function and $\E[\nu_i^2]=1$. 

To consider Assumption RP, we will use the relation $\E[a_la_l\ttop]=\mathsf{G}_{\circ}$, which follows from the fact that $\E[U_iU_i\ttop]=\ts\frac{1}{d}I_d$. Therefore, if we apply the law of large numbers to 
$\ts\frac{1}{n}A_n\ttop A_n = \ts\frac{1}{n}\sum_{l=1}^n a_la_l\ttop$, then we have the limit  $\ts\frac{1}{n}A_n\ttop A_n \to \mathsf{G}_{\circ}$ in probability. Thus, Assumption RP holds in probability.

Next, to consider Assumption RS in the case of uniform sampling, the formula~\eqref{eqn:basicvar} gives 
\begin{equation}
\var(\tilde r_n\ttop C \tilde r_n|A_n) = \ts\frac{1}{n}\tsum_{l=1}^n \big(a_l\ttop Ca_l\big)^2 - \Big(\ts\frac{1}{n}\sum_{l=1}^n a_l\ttop Ca_l\Big)^2.
\end{equation}
In order to verify Assumption RS in probability, we will first show that $\var(\tilde r_n\ttop C\tilde r_n|A_n)$ converges in probability to a constant $\ell(C)$. Using a known formula for the expectation of quadratic forms involving elliptical random vectors~\citep[Lemma A.1]{elliptical_LSS}, as well as the law of large numbers, we have
$$\var(\tilde r_n\ttop C\tilde r_n|A_n) \, \xrightarrow{ } \, \ell(C)$$
in probability, where the limit is given by
$$\ell(C) \, = \, \ts\frac{\E[\nu_1^4]}{1+2/d}\Big(\tr(C\mathsf{G}_{\circ})^2+\tr(\mathsf{G}_{\circ}C\mathsf{G}_{\circ}C\ttop)+ \tr(\mathsf{G}_{\circ}C\mathsf{G}_{\circ}C)\Big)-\tr(C\mathsf{G}_{\circ})^2.$$
In turn, this formula for $\ell(C)$ implies
\begin{align}
\ell(\mathsf{v}_1\mathsf{v}_1\ttop) = \lambda_1^2(\mathsf{G}_{\circ})\Big(\ts\frac{3\E[\nu_1^4]}{1+2/d}-1\Big) \ \ \ \ \text{ and } \ \ \ \  \ell(\mathsf{v}_1\mathsf{v}_2\ttop) = \lambda_1\lambda_2\ts\frac{\E[\nu_1^4]}{1+2/d},
%
%
\end{align}
where the positivity of  $\ell(\mathsf{v}_1\mathsf{v}_2\ttop)$ is clear, and the positivity of $\ell(\mathsf{v}_1\mathsf{v}_1\ttop)$ holds when $d\geq 2$, because 
 $\E[\nu_1^4]\geq (\E[\nu_1^2])^2=1$. Lastly, to verify the growth condition involving $\max_{1\leq l\leq n}\ts\frac{1}{\sqrt{np_l}} \|a_l\|_2$, recall that $p_l=1/n$ for uniform sampling, and that $\nu_i$ is assumed to have a moment generating function. Under these conditions, it follows from~\citep[Lemma 2.2.2]{vaartWellner} that $\E[\max_{1\leq l\leq n} \|a_l\|_2]=\mathcal{O}(\log(n))$, which implies that $t_n^{-1/8}\max_{1\leq l\leq n}\|a_l\|_2\to 0$ in probability under the mild condition $\log(n)t_n^{-1/8}\to 0$.

\section{Additional experiments}\label{sec:suppnum}

In this section, we provide two sets of additional experiments that go beyond the settings considered in the main text. Specifically, these extra experiments look at happens when the decay profile of the singular values is changed to a different type, or when the index set $\mathcal{J}=\{1\}$ is changed to $\mathcal{J}=\{1,2,3\}$.

\subsection{Synthetic matrices with alternative decay profile}

As an alternative to the singular value decay profile $\sigma_j=j^{-\beta}$ with parameter values $\beta\in\{0.5,1,2\}$ used for the experiments of the main text, we now look at a profile of the form $\sigma_j=10^{-\gamma j}$, with parameter values $\gamma\in\{0.05,0.1,0.5\}$. In particular, this type of decay profile arises in many applications related to differential equations and dynamical systems. Apart from the change in the decay profile, the experiments here were organized in the same manner as those for the synthetic matrices in the main text, and the results are plotted in the same format.

Figure~\ref{fig:results_svd_sketching_exp_k1} shows how close the bootstrap quantile estimates are to the true quantiles $q_{_U}(t)$, $q_{_{\Sigma}}(t)$, and $q_{_V}(t)$ for sketch sizes $t\in\{500,\dots,6000\}$. Recall that the the dashed black curve represents the true quantiles, the blue curve represents the average of the non-extrapolated estimates, and the red curve represents the average of the extrapolated estimates from $t_0=500$ (with the light red envelope representing $\pm 1$ standard deviation for the extrapolated estimates). The three rows of plots correspond to the singular values (top), right singular vectors (middle), and left singular vectors (bottom). Overall, the plots show that the bootstrap estimates are quite accurate, and in essence, the results for the current setting are on par with those shown in the main text.

\begin{figure*}[t!]\vspace{+0.5cm}
	
	\centering
	\begin{subfigure}{1\textwidth}	
		\centering
		\DeclareGraphicsExtensions{.pdf}
		\begin{overpic}[width=0.31\textwidth]{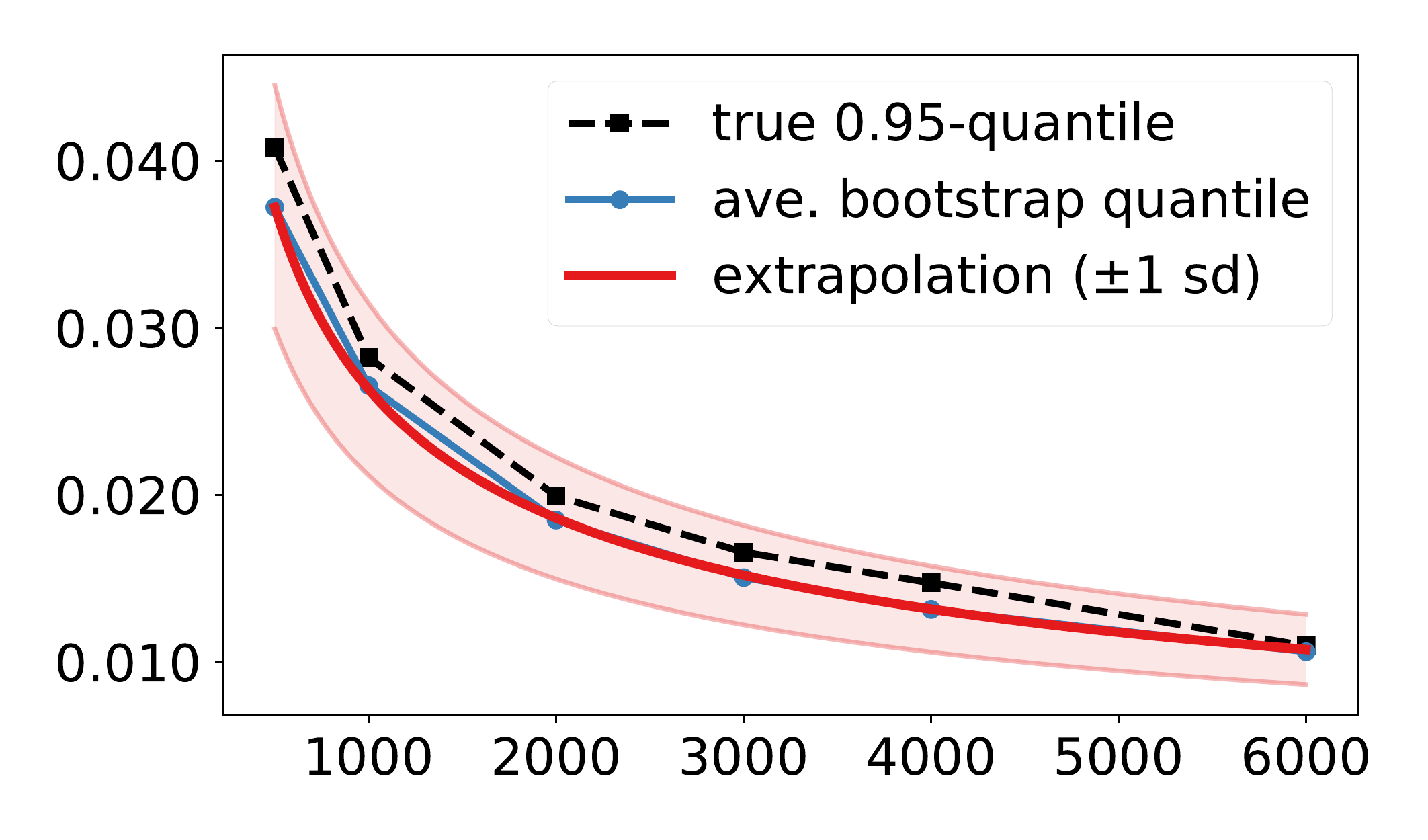} 
			\put(-6,26){\rotatebox{90}{\footnotesize $\tilde\e_{_{\Sigma}}(t)$}}
			\put(45,58){\color{black}{\scriptsize $\gamma=0.05$}} 			
		\end{overpic}\hspace*{-0.2cm}
		~
		\begin{overpic}[width=0.31\textwidth]{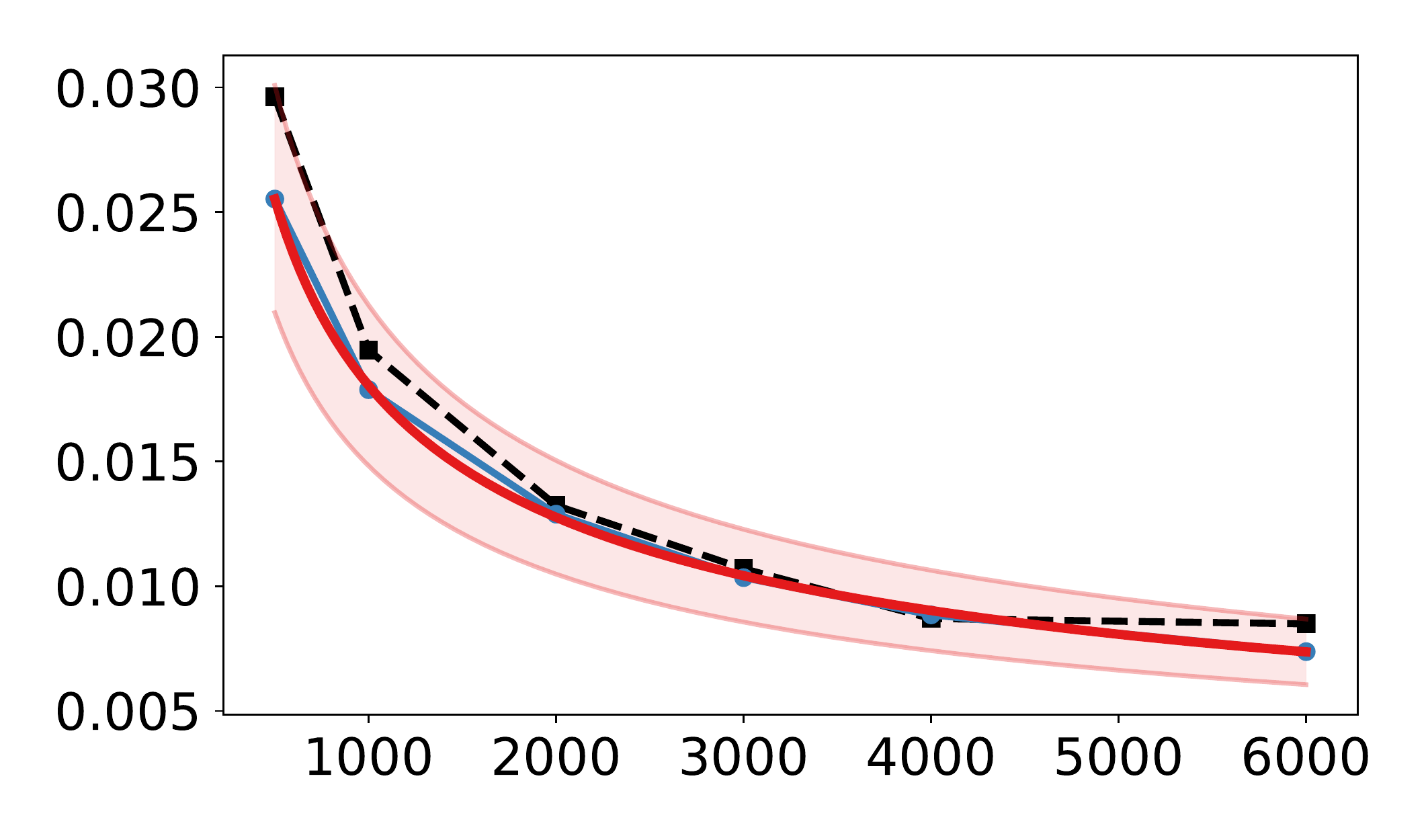} 
			\put(45,58){\color{black}{\scriptsize $\gamma=0.1$}} 			 			
		\end{overpic}\hspace*{-0.2cm}
		~
		\begin{overpic}[width=0.31\textwidth]{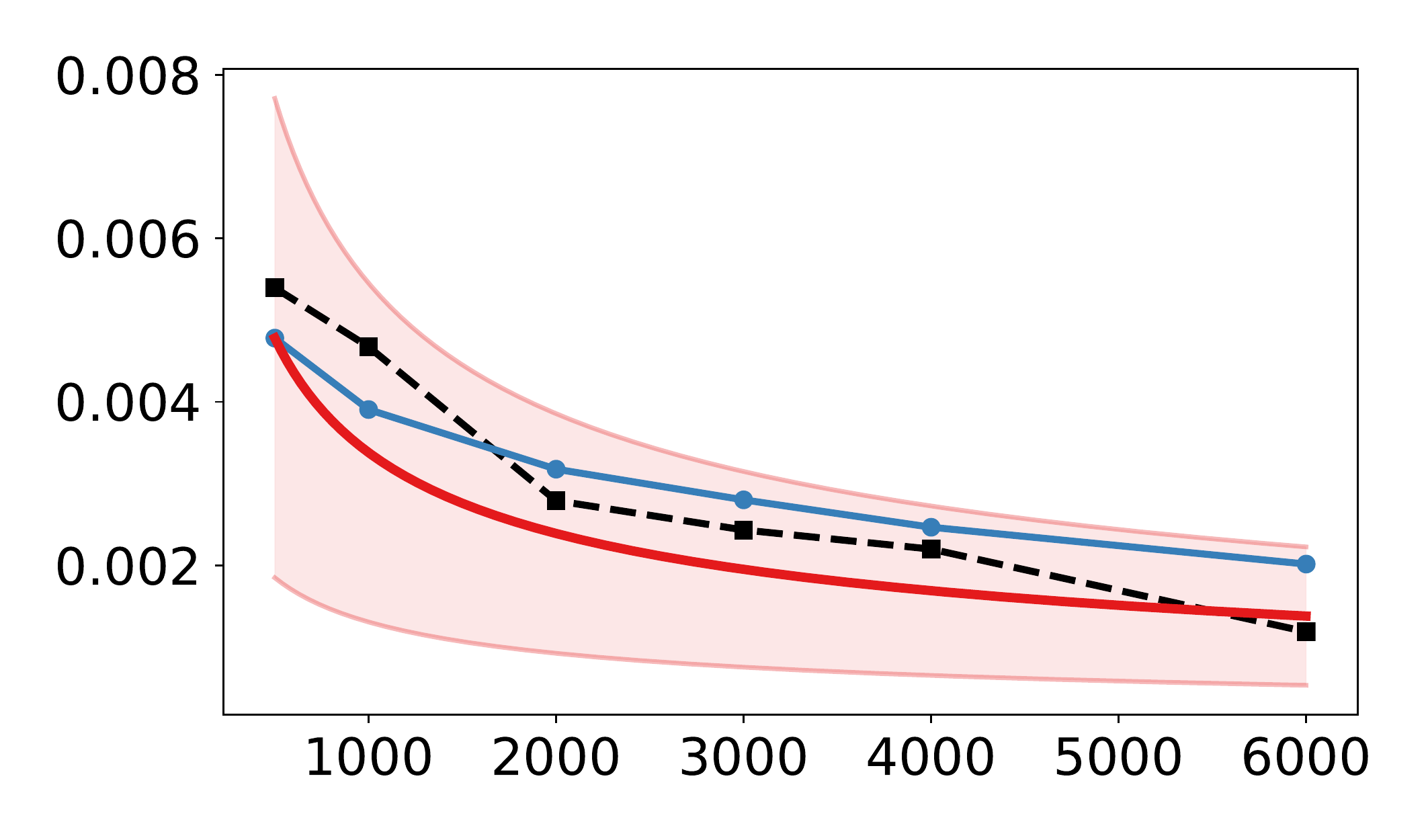} 
			\put(45,58){\color{black}{\scriptsize $\gamma=0.5$}} 
			\put(100,10){\rotatebox{90}{\scriptsize (singular values)}}
		\end{overpic}
	\end{subfigure}\vspace{-0.0cm}	
	
	\begin{subfigure}{1\textwidth}	
		\centering
		\DeclareGraphicsExtensions{.pdf}
		\begin{overpic}[width=0.31\textwidth]{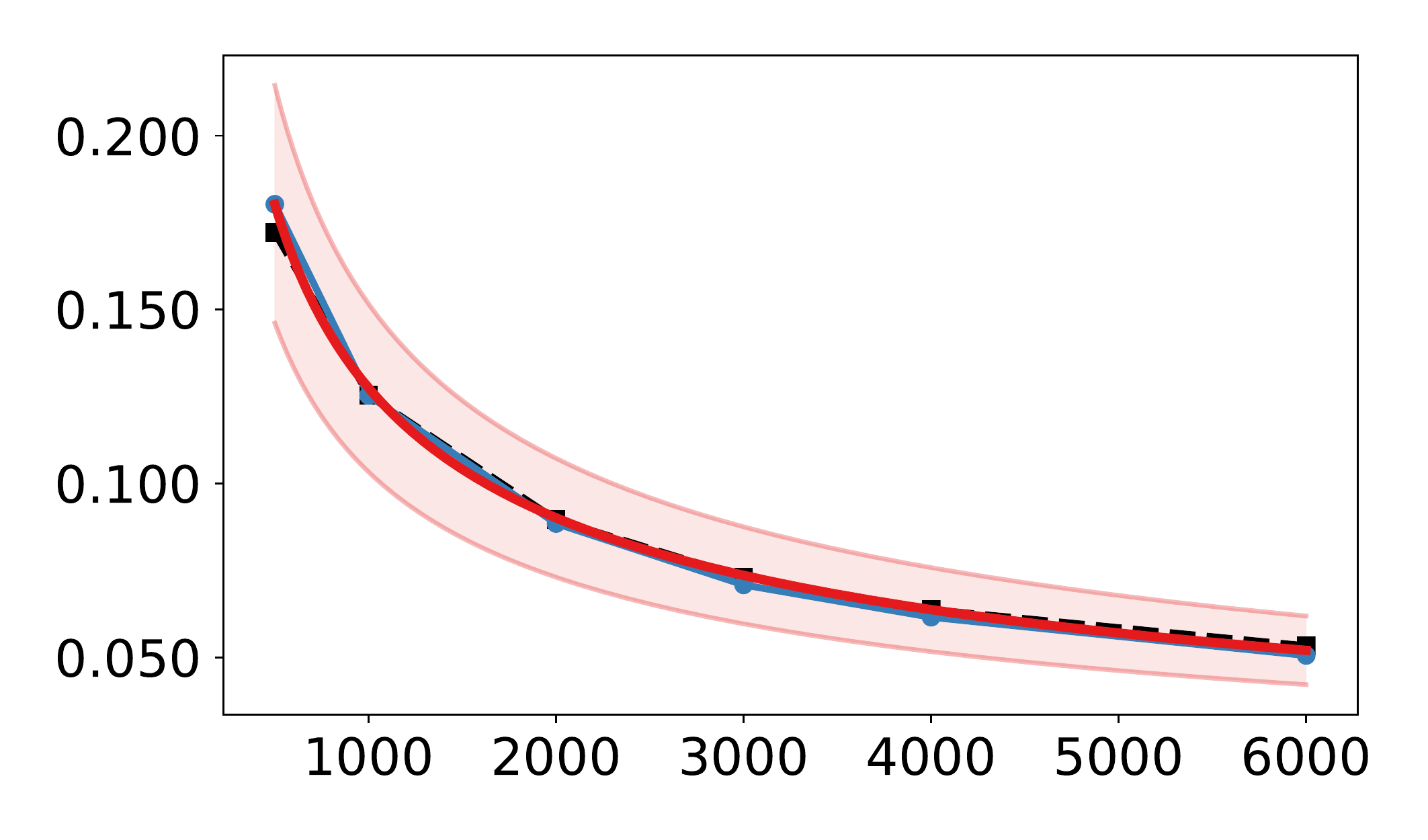} 
			\put(-6,24){\rotatebox{90}{\footnotesize $\tilde\e_{_{V}}(t)$}}
		\end{overpic}\hspace*{-0.2cm}
		~
		\begin{overpic}[width=0.31\textwidth]{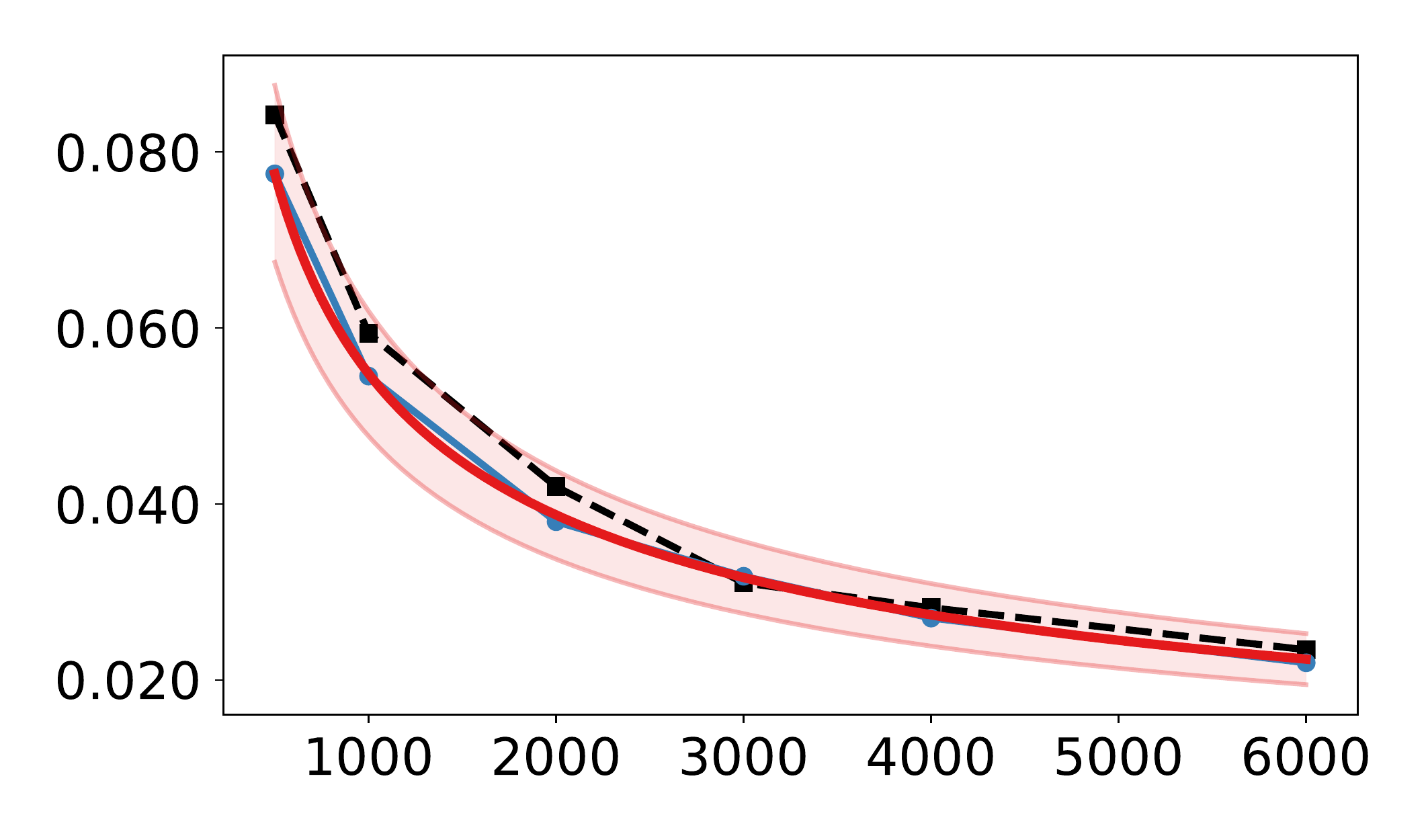} 
		\end{overpic}\hspace*{-0.2cm}
		~
		\begin{overpic}[width=0.31\textwidth]{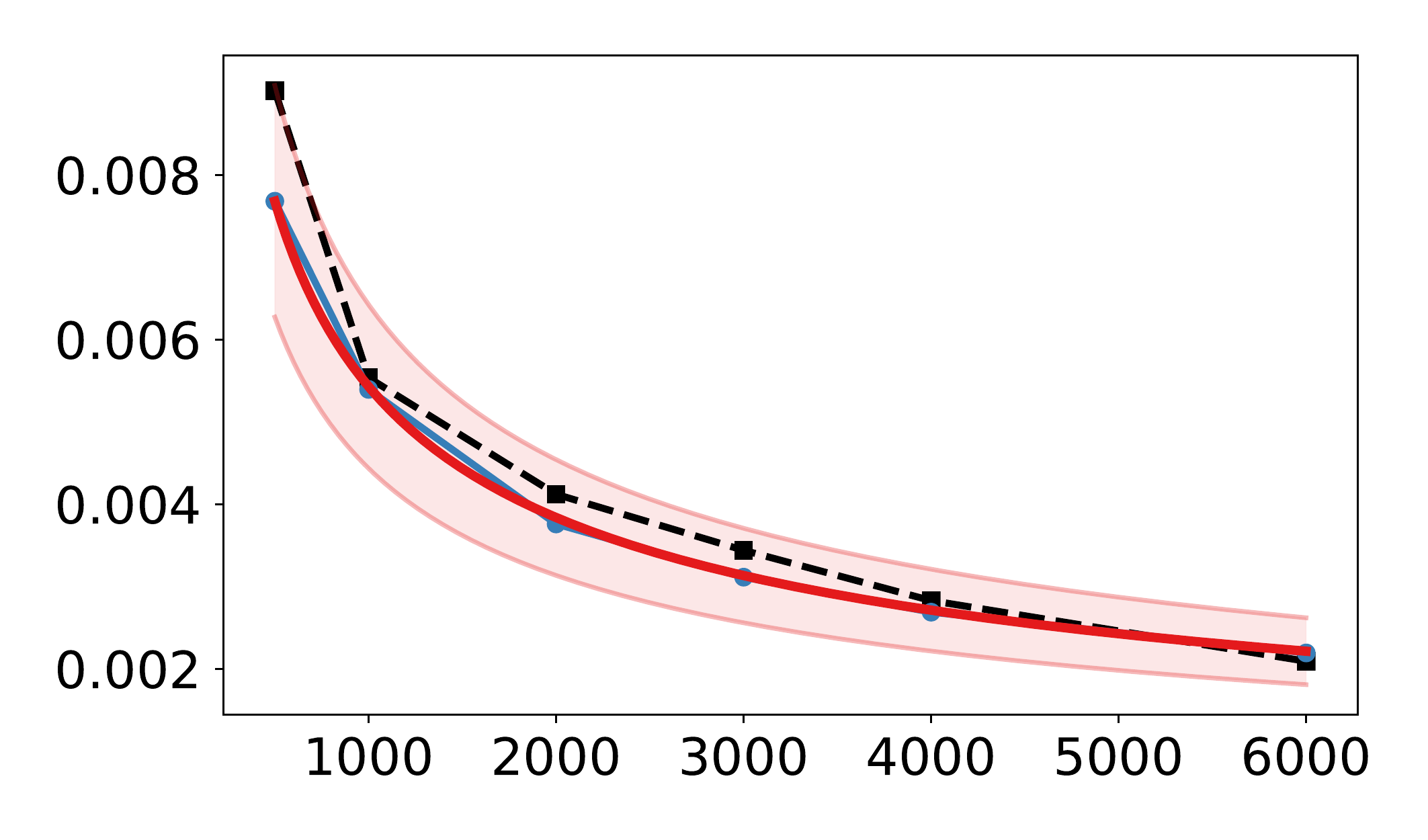} 
			\put(100,3){\rotatebox{90}{\scriptsize (right singular vectors)}}			
		\end{overpic}
	\end{subfigure}\vspace{-0.0cm}	
	
	\begin{subfigure}{1\textwidth}	
		\centering
		\DeclareGraphicsExtensions{.pdf}
		\begin{overpic}[width=0.31\textwidth]{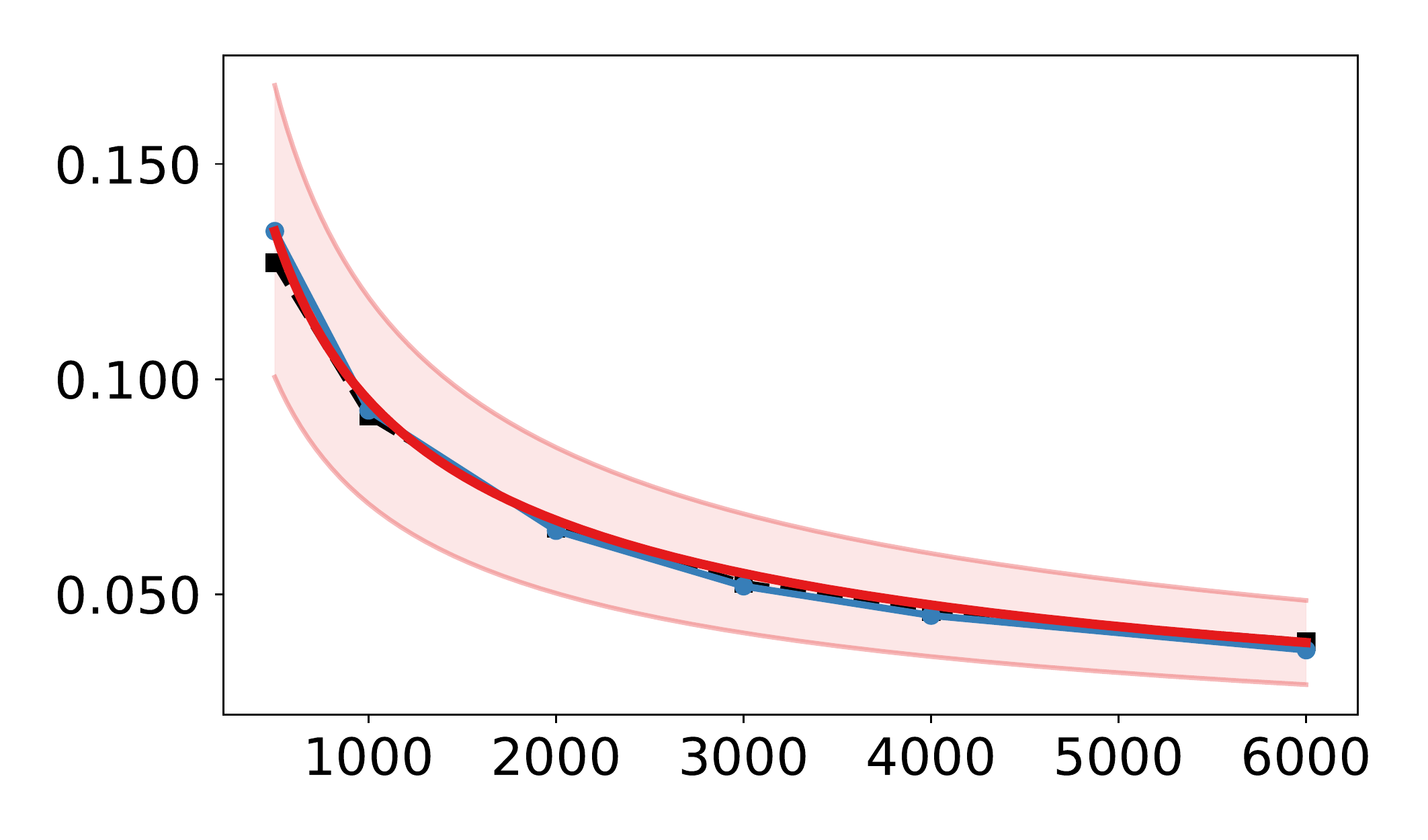} 
			\put(42,-2){\color{black}{\footnotesize sketch size $t$}}   
			\put(-6,24){\rotatebox{90}{ \footnotesize $\tilde\e_{_{U}}(t)$}}
		\end{overpic}\hspace*{-0.2cm}
		~
		\begin{overpic}[width=0.31\textwidth]{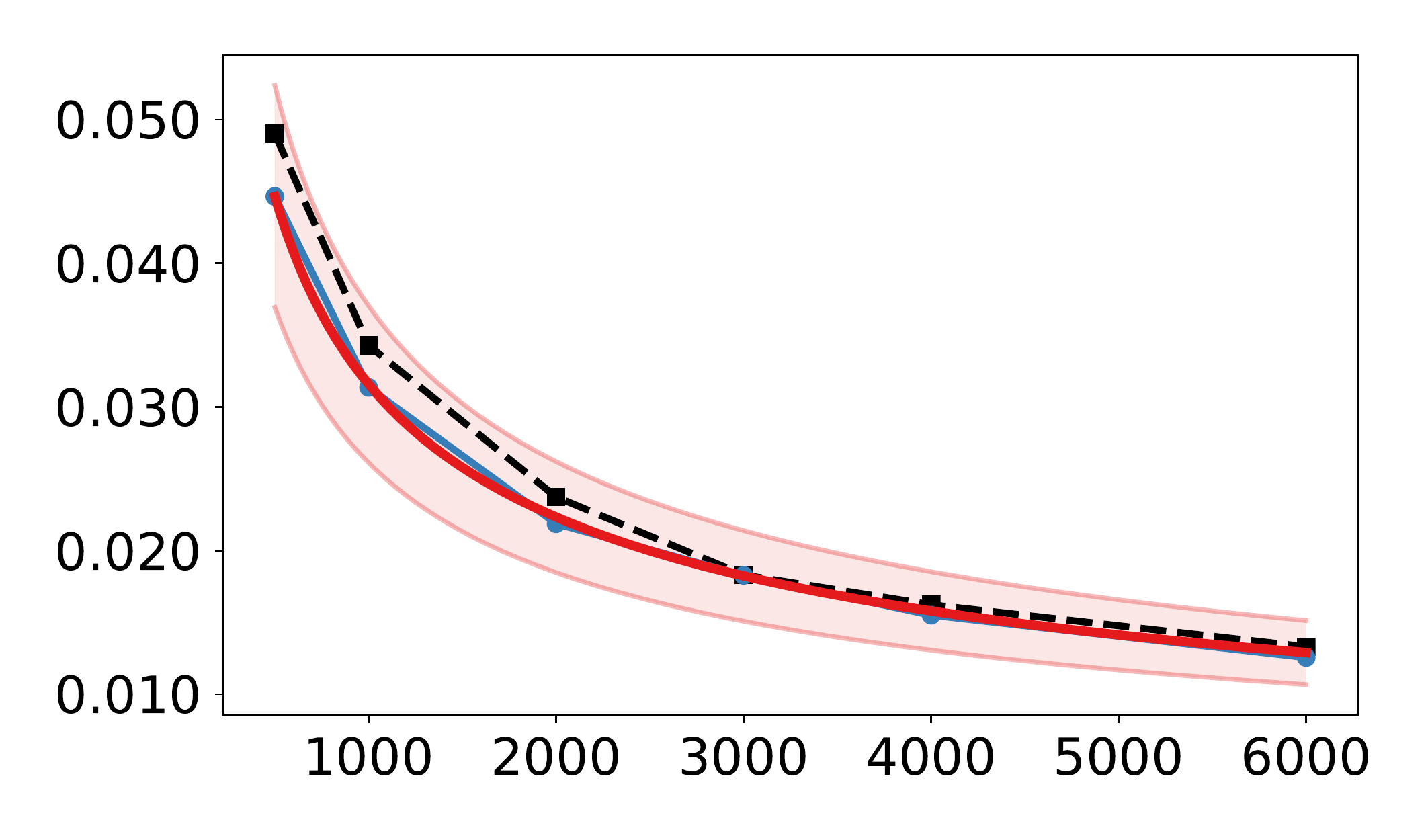} 
			\put(42,-2){\color{black}{\footnotesize sketch size $t$}}   
		\end{overpic}\hspace*{-0.2cm}
		~
		\begin{overpic}[width=0.31\textwidth]{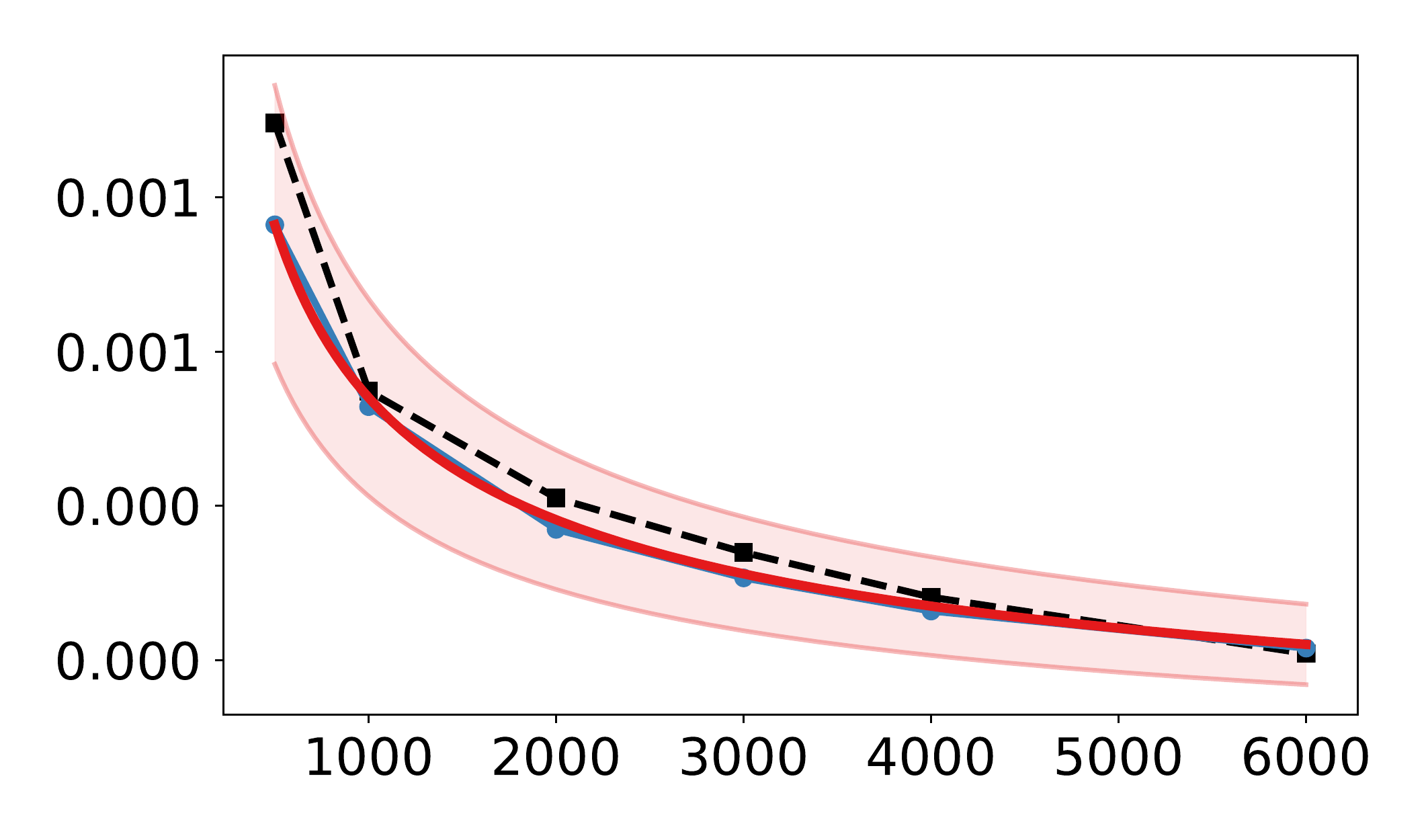} 
			\put(42,-2){\color{black}{\footnotesize sketch size $t$}}   
			\put(100,4){\rotatebox{90}{\scriptsize (left singular vectors)}}			
		\end{overpic}
	\end{subfigure}
	
	\vspace{+.5cm}	
	\caption{We consider artificial matrices of dimension $(n,d)=(10^5,3\times 10^3)$ that have singular value decay profiles of the form $\sigma_j=10^{-\gamma j}$ for $j\in\{1,\dots,d\}$ with $\gamma\in\{0.05, 0.1, 0.5\}$. The error variables correspond to the index set $\mathcal{J}=\{1\}$, and the simulations involve $500$ trials and $30$ bootstraps per trial. The rows correspond to the error quantiles for the singular values (top), right singular vectors (middle), and left singular vectors (bottom).}
	\label{fig:results_svd_sketching_exp_k1}
\end{figure*}
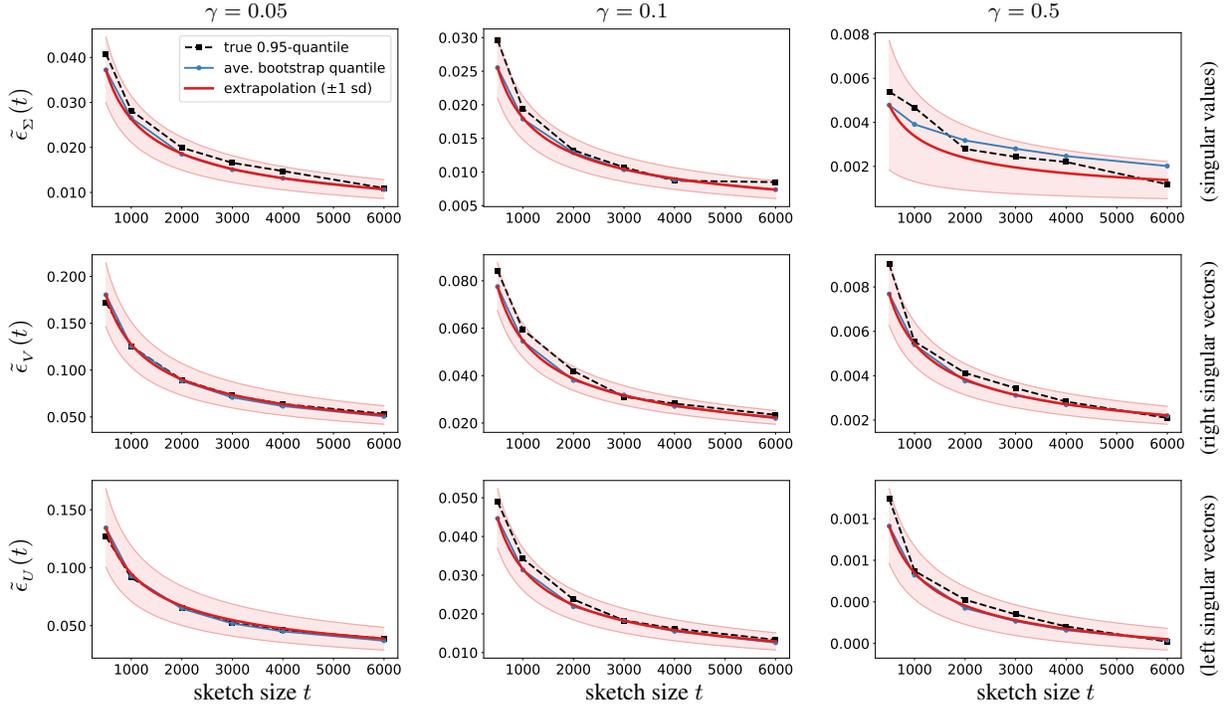

\subsection{Results for the index set $\mathcal{J}=\{1,2,3\}$}\label{app:otherJ}

Recall that the sketching error variables are defined with respect to an index set $\mathcal{J}\subset\{1,\dots,k\}$ according to
\begin{equation*}
\begin{split}
\small
\tilde \e_{_U}\!(t) &= \max_{j\in\mathcal{J}}\rho_{\sin}(\tilde u_j, u_j)\\[0.2cm]
 \tilde \e_{_V}\!(t) &= \max_{j\in\mathcal{J}}\rho_{\sin}(\tilde v_j, v_j),\\[0.2cm]
\tilde \e_{_{\Sigma}}\!(t)&= \!\max_{j\in\mathcal{J}}|\tilde\sigma_j-\sigma_j|.
\end{split}
\end{equation*}
Whereas the synthetic examples in the main text considered the sketching errors for the leading triple $(u_1,\sigma_1,v_1)$ corresponding to $\mathcal{J}=\{1\}$, we now look at the case when $\mathcal{J}=\{1,2,3\}$. In other words, the new experiments in this section correspond to a situation where the user would like to have \emph{simultaneous} control over the sketching errors associated the top three singular vectors/values. Apart from this change in the choice of $\mathcal{J}$, all other aspects of the design and presentation of the experiments remain the same as in the main text. Given that a maximum is now being taken over a larger set of indices, the magnitudes of $\tilde\e_{_{\Sigma}}(t)$, $\tilde\e_{_{U}}(t)$, and $\tilde\e_{_{V}}(t)$ will necessarily be larger. Nevertheless, the important point to notice is that the quality of the bootstrap quantile estimates remains essentially just as good as in the case when $\mathcal{J}=\{1\}$.

\vspace{+0.8cm}
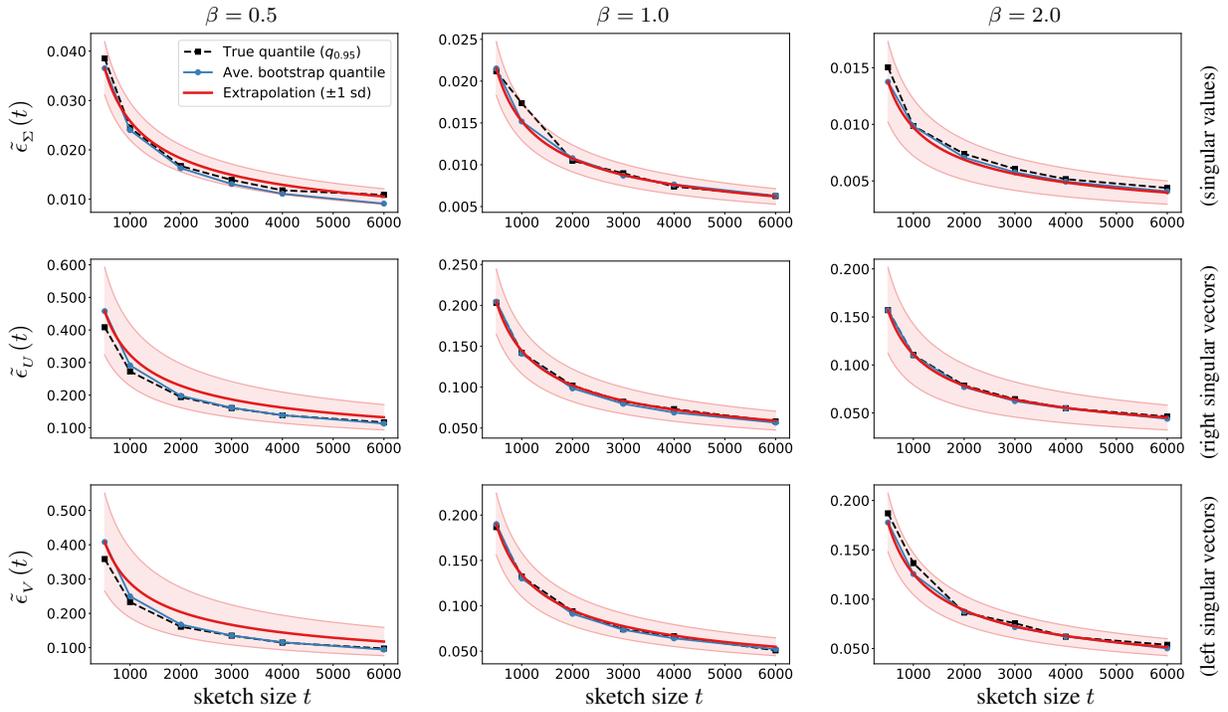
\begin{figure*}[!t]
	
	\centering
	\begin{subfigure}{1\textwidth}	
		\centering
		\DeclareGraphicsExtensions{.pdf}
		\begin{overpic}[width=0.31\textwidth]{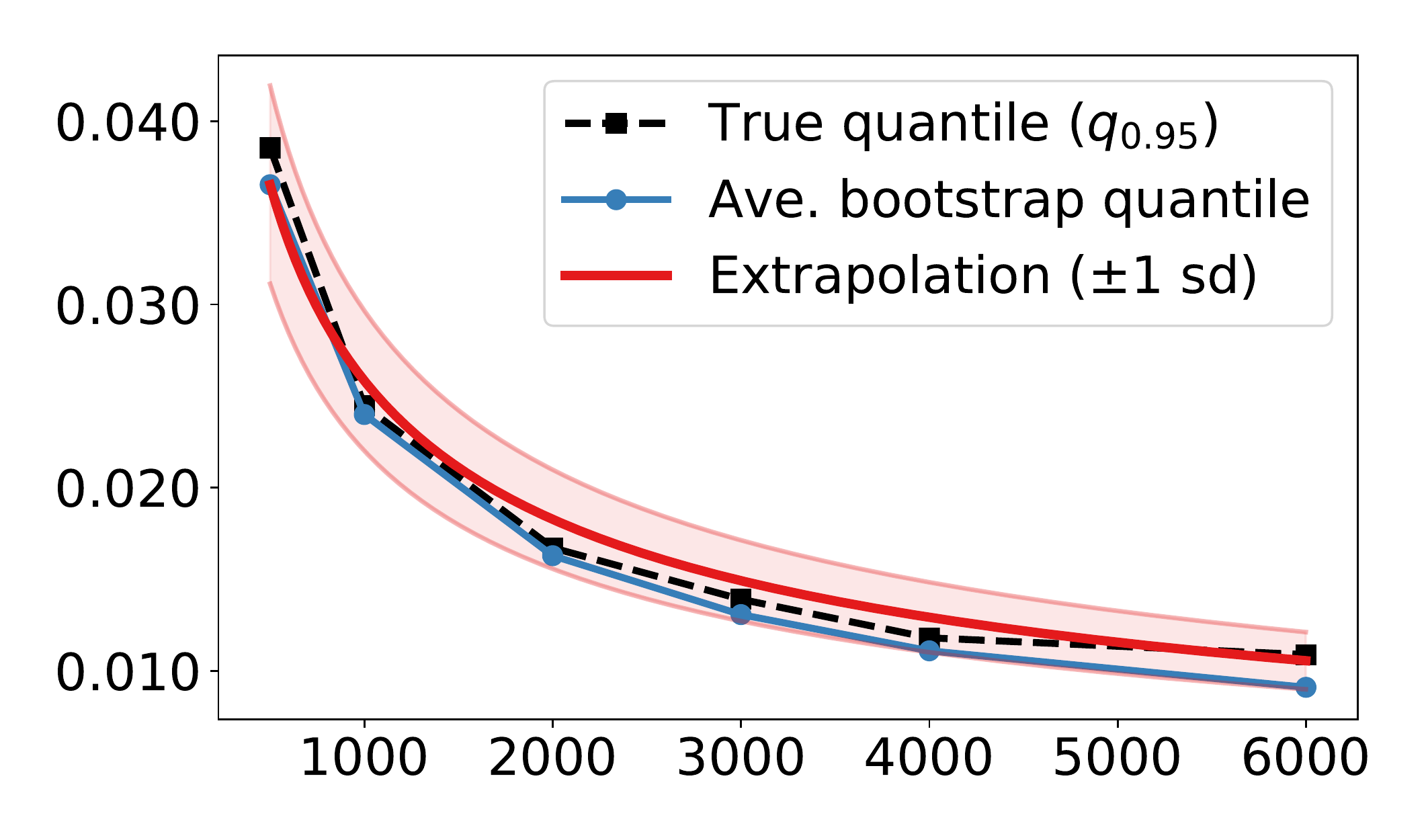} 
			\put(-6,24){\rotatebox{90}{\footnotesize $\tilde\e_{_{\Sigma}}(t)$}}
			\put(45,58){\color{black}{\scriptsize $\beta=0.5$}} 			
		\end{overpic}\hspace*{-0.2cm}
		~
		\begin{overpic}[width=0.31\textwidth]{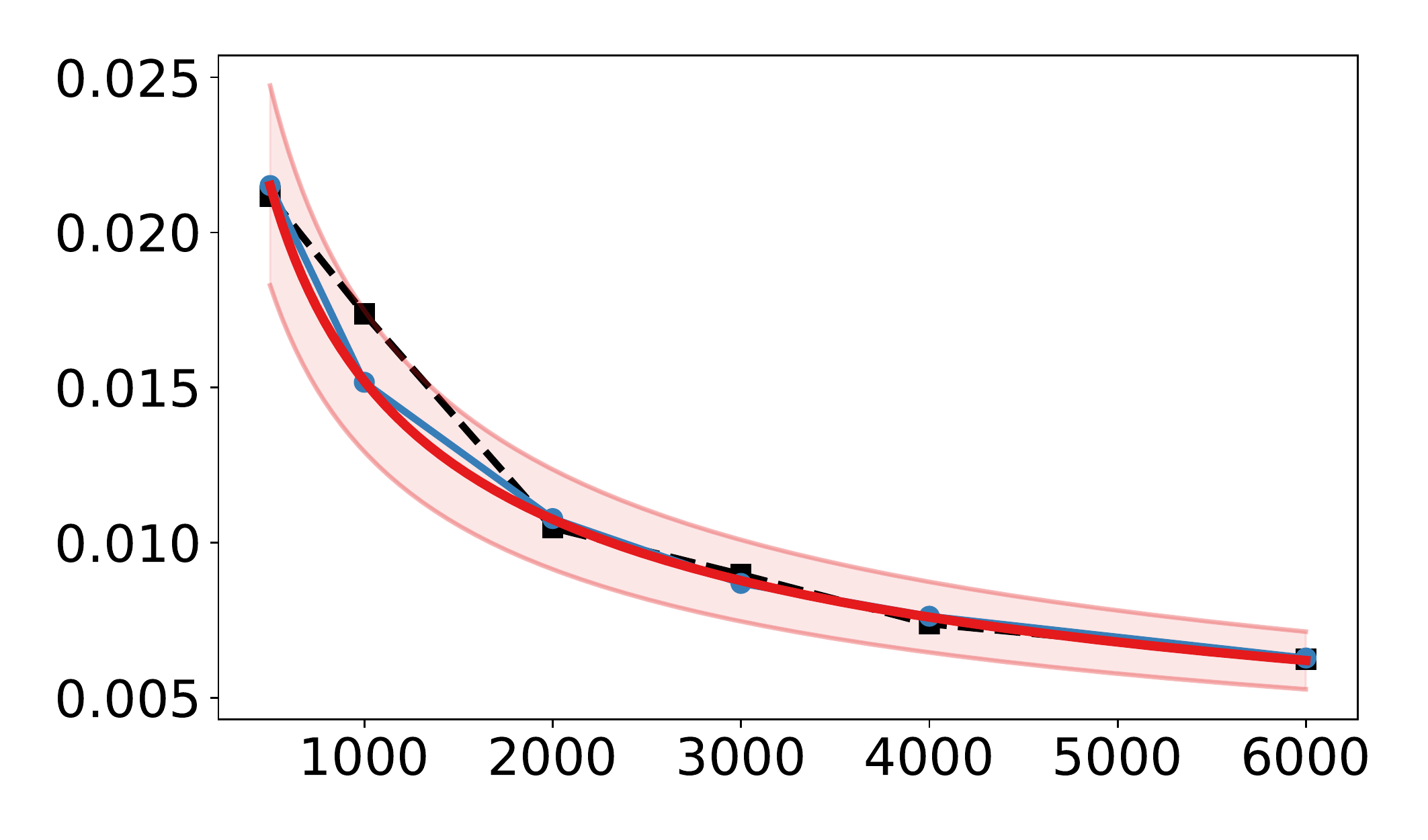} 
			\put(45,58){\color{black}{\scriptsize $\beta=1.0$}} 			 			
		\end{overpic}\hspace*{-0.2cm}
		~
		\begin{overpic}[width=0.31\textwidth]{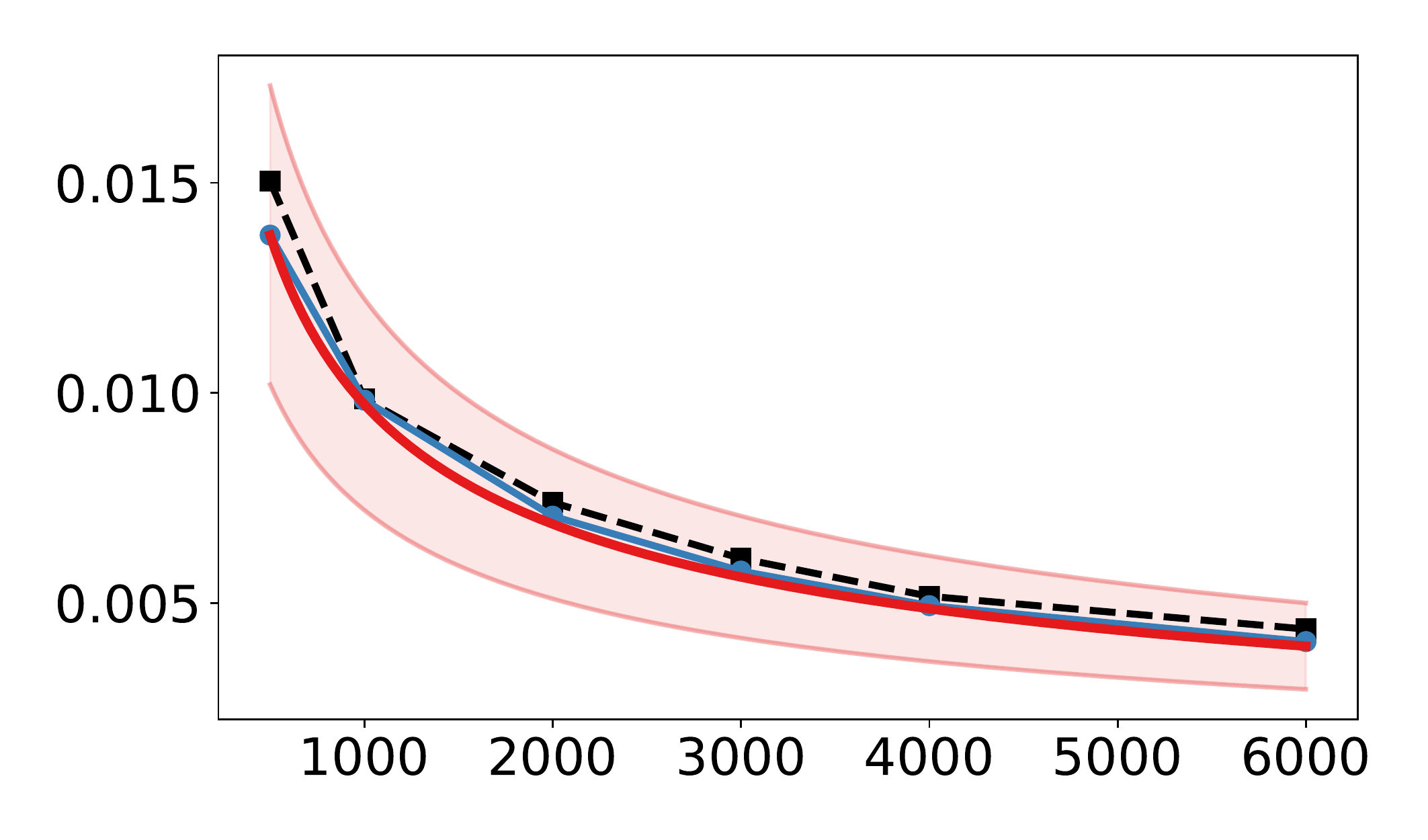} 
			\put(45,58){\color{black}{\scriptsize $\beta=2.0$}} 
			\put(100,10){\rotatebox{90}{\scriptsize (singular values)}}
		\end{overpic}
	\end{subfigure}\vspace{-0.0cm}	
	
	\begin{subfigure}{1\textwidth}	
		\centering
		\DeclareGraphicsExtensions{.pdf}
		\begin{overpic}[width=0.31\textwidth]{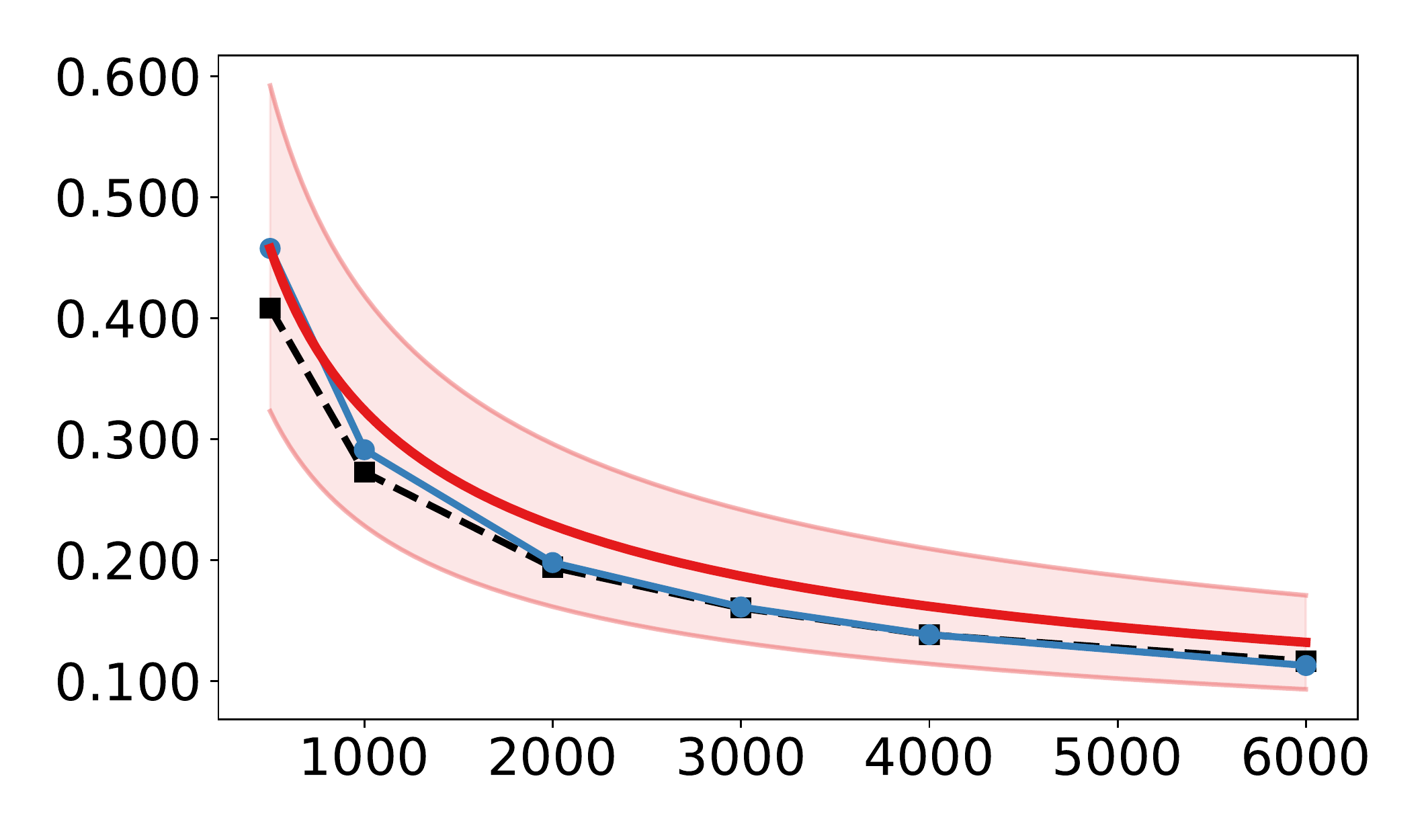} 
			\put(-6,24){\rotatebox{90}{\footnotesize $\tilde\e_{_{U}}(t)$}}
		\end{overpic}\hspace*{-0.2cm}
		~
		\begin{overpic}[width=0.31\textwidth]{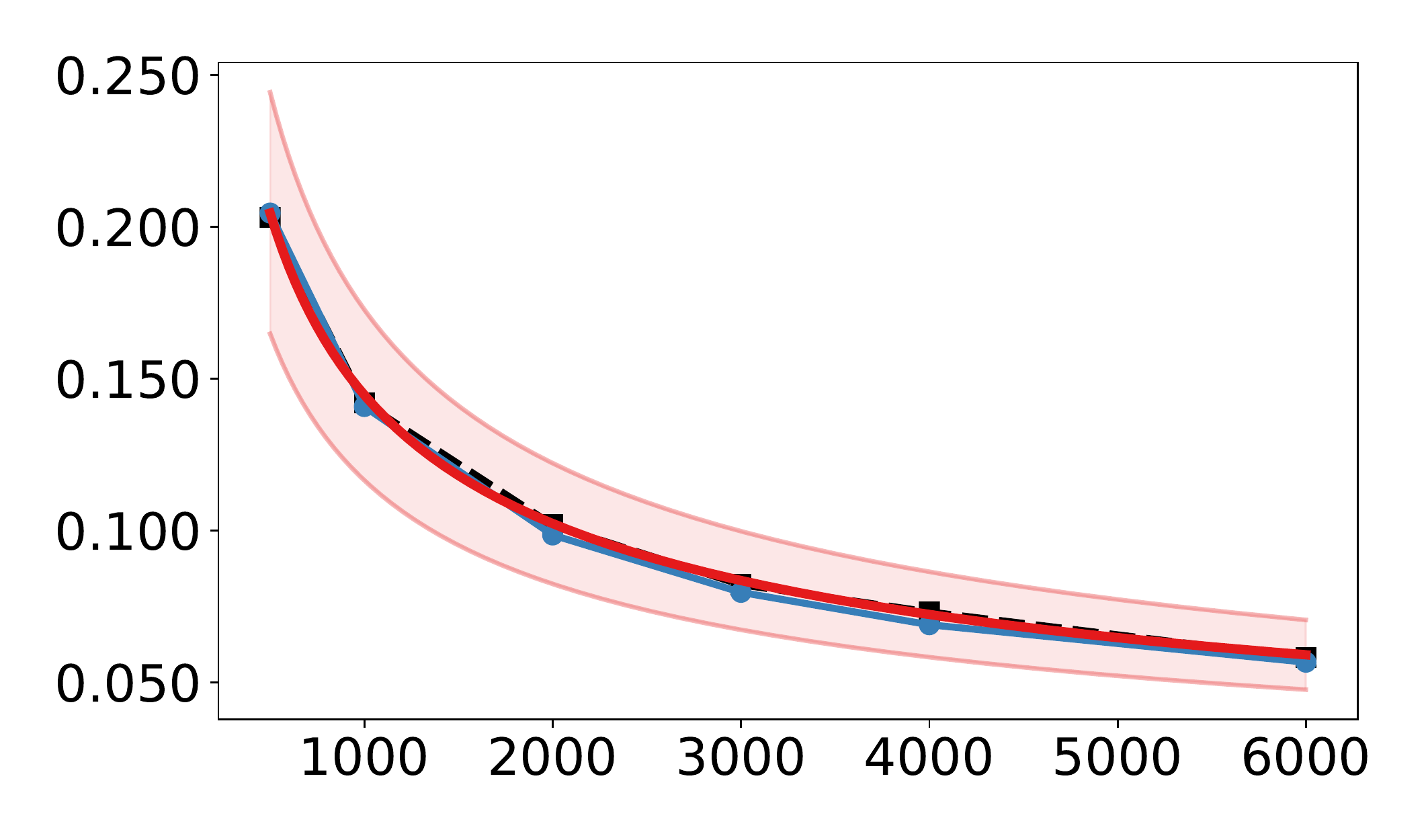} 
		\end{overpic}\hspace*{-0.2cm}
		~
		\begin{overpic}[width=0.31\textwidth]{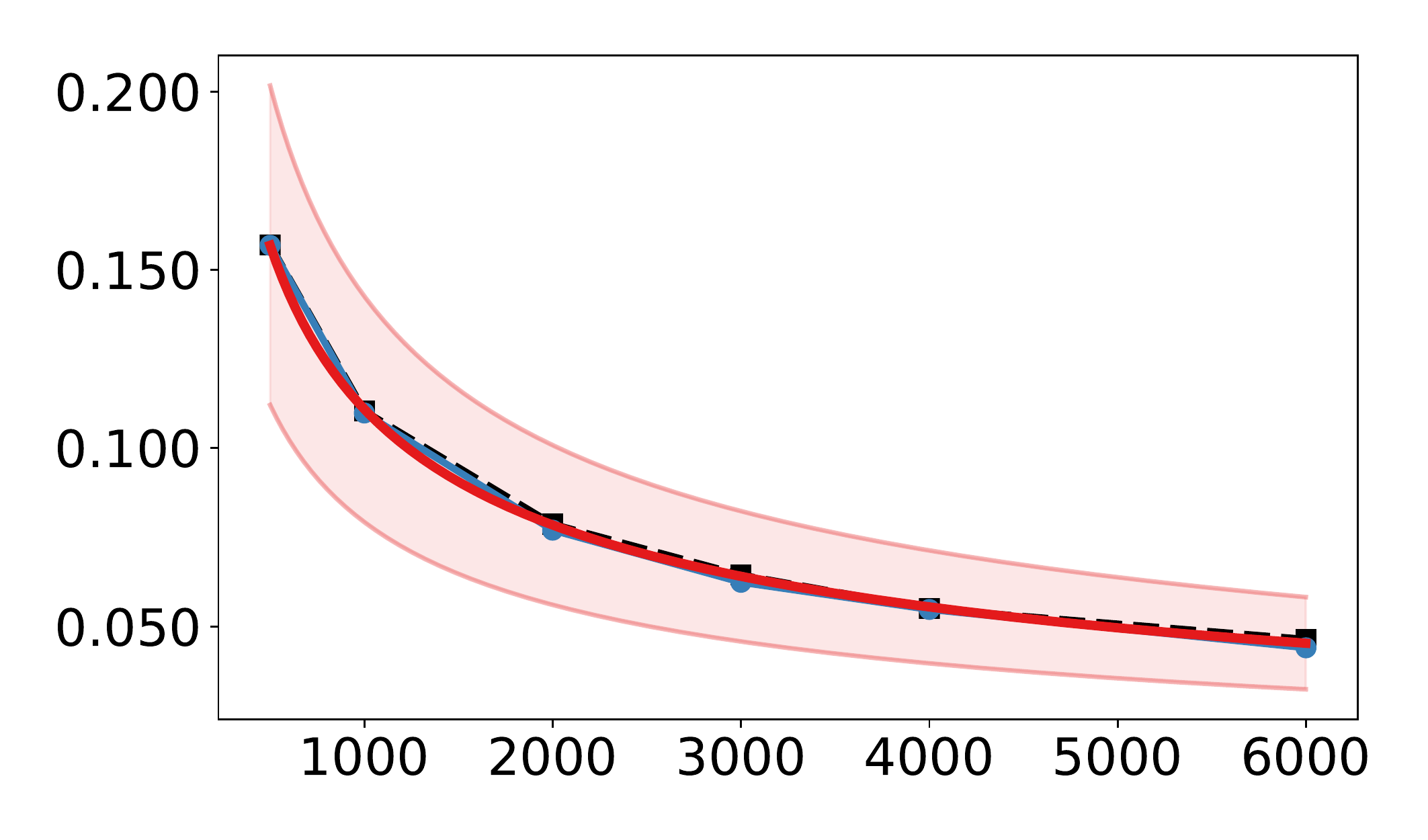} 
			\put(100,3){\rotatebox{90}{\scriptsize (right singular vectors)}}			
		\end{overpic}
	\end{subfigure}\vspace{-0.0cm}	
	
	\begin{subfigure}{1\textwidth}	
		\centering
		\DeclareGraphicsExtensions{.pdf}
		\begin{overpic}[width=0.31\textwidth]{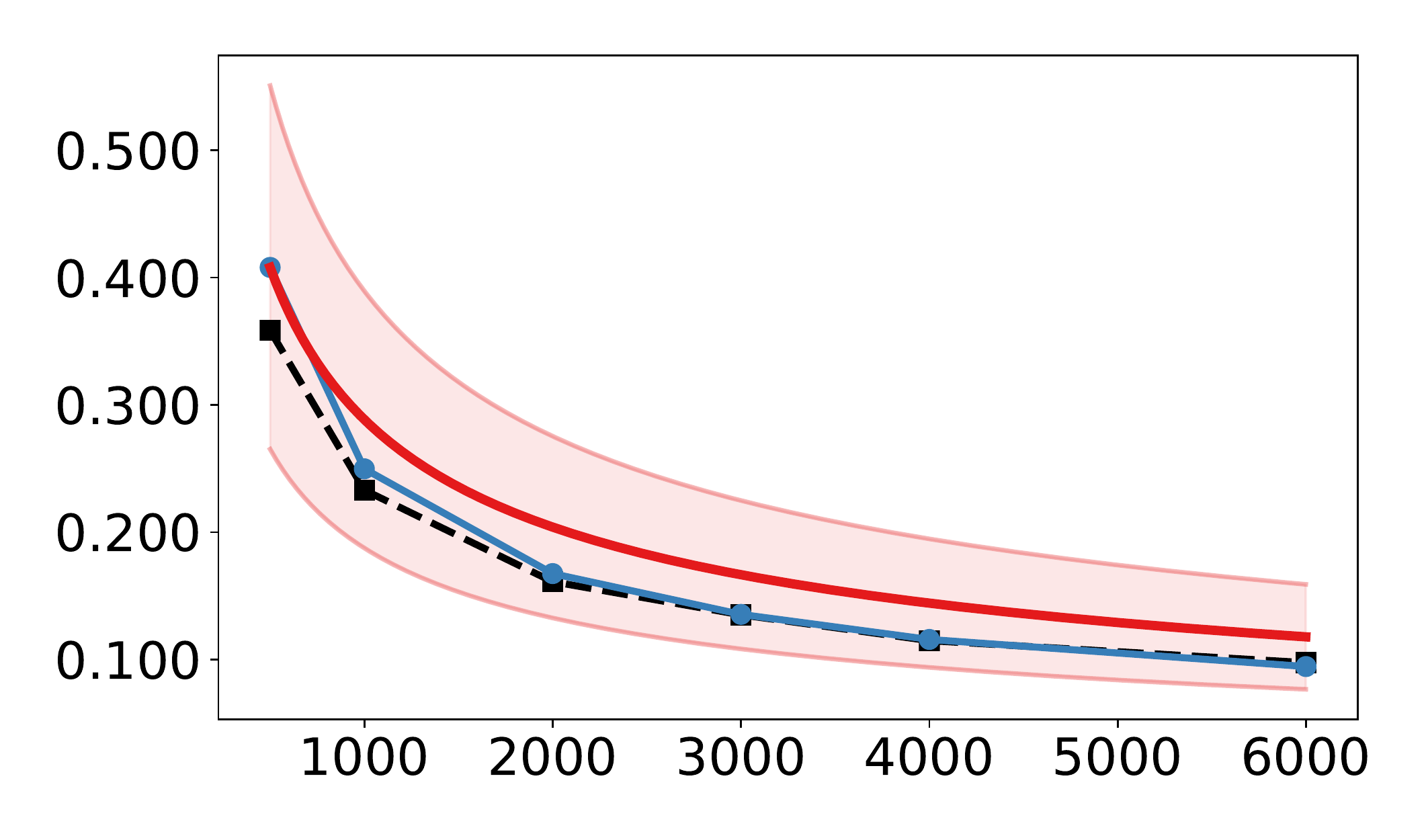} 
			\put(42,-2){\color{black}{\footnotesize sketch size $t$}}   
			\put(-6,24){\rotatebox{90}{\footnotesize $\tilde\e_{_{V}}(t)$}}
		\end{overpic}\hspace*{-0.2cm}
		~
		\begin{overpic}[width=0.31\textwidth]{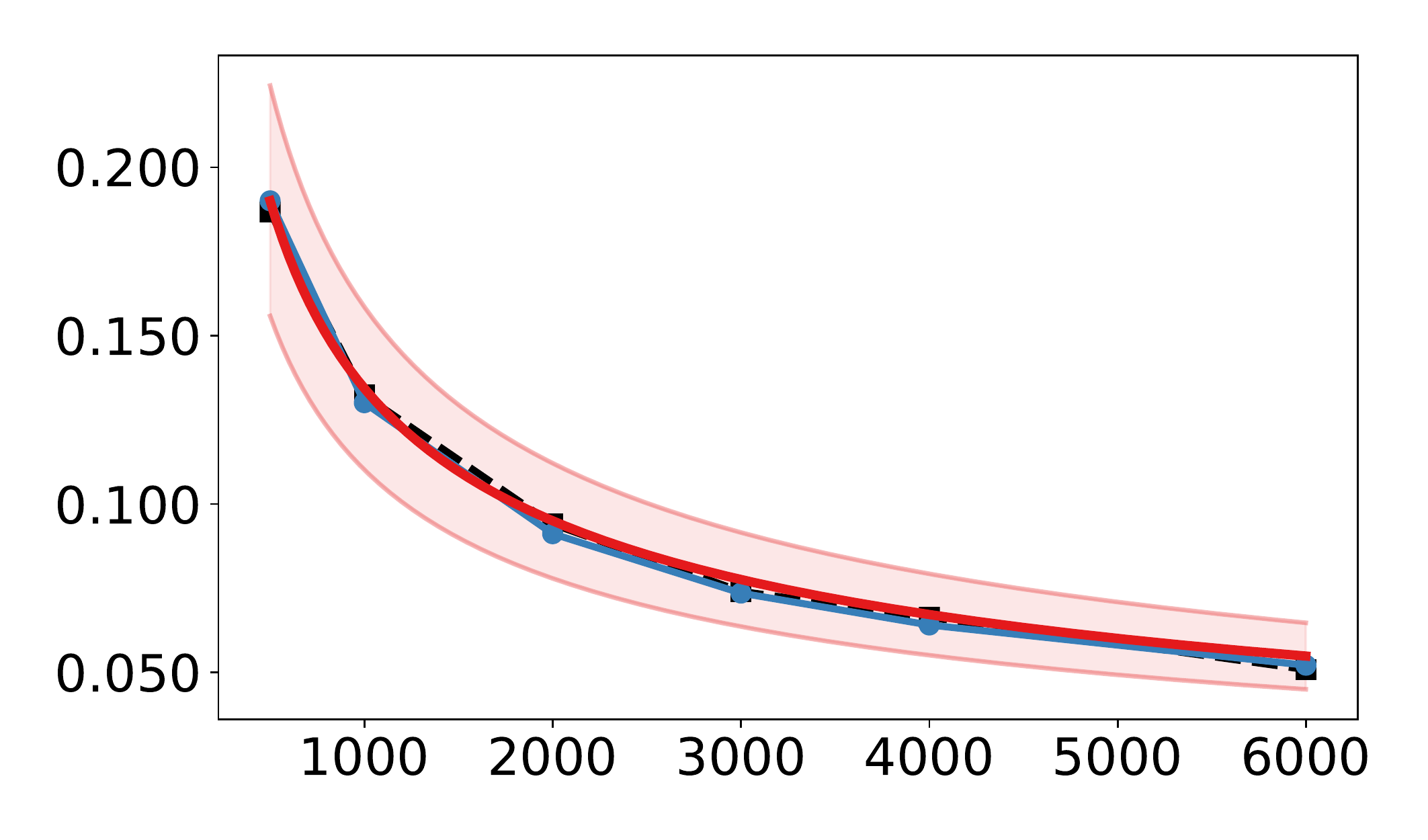} 
			\put(42,-2){\color{black}{\footnotesize sketch size $t$}}   
		\end{overpic}\hspace*{-0.2cm}
		~
		\begin{overpic}[width=0.31\textwidth]{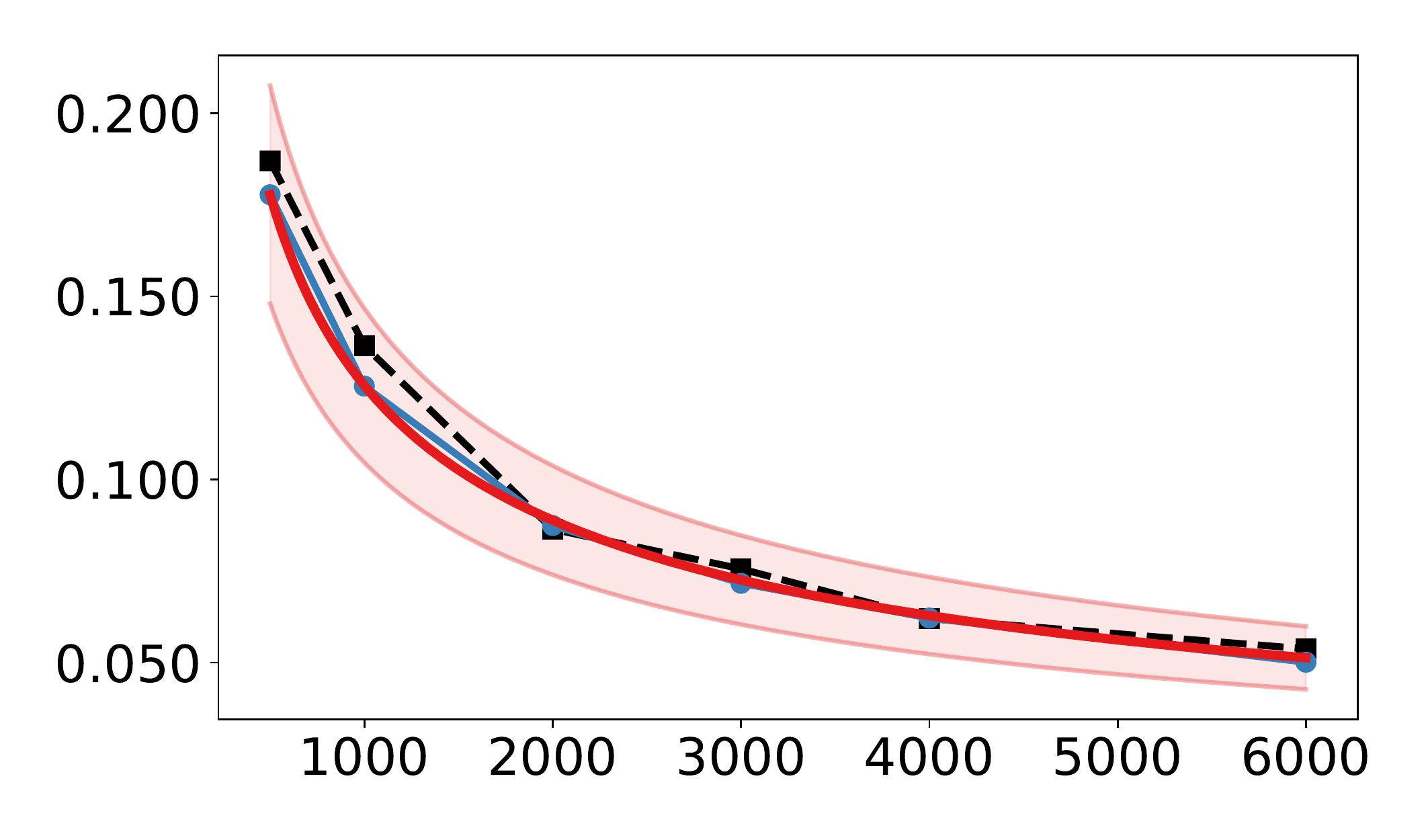} 
			\put(42,-2){\color{black}{\footnotesize sketch size $t$}}   
			\put(100,5){\rotatebox{90}{\scriptsize (left singular vectors)}}			
		\end{overpic}
	\end{subfigure}
	
	\vspace{+.5cm}	
	\caption{We consider artificial matrices of dimension $(n,d)=(3.5\times 10^4, 3\times 10^3)$ that have singular value decay profiles of the form $\sigma_j=j^{-\beta}$ for $j\in\{1,\dots,d\}$ with parameter values $\beta\in\{0.5,1,2\}$. The error variables correspond to the index set $\mathcal{J}=\{1,2,3\}$, and the simulations involve $500$ trials and $30$ bootstraps per trial. The rows correspond to the error quantiles for the singular values (top), right singular vectors (middle), and left singular vectors (bottom).}
	\label{fig:results_svd_sketching_poly_k}
\end{figure*}

\end{document}